%% file: main.tex
\documentclass{article}

\usepackage[numbers]{natbib}

\usepackage[preprint]{neurips_2024}

\usepackage[utf8]{inputenc} 
\usepackage[T1]{fontenc}    
\usepackage{hyperref}       
\usepackage{url}            
\usepackage{booktabs}       
\usepackage{amsfonts}       
\usepackage{nicefrac}       
\usepackage{microtype}      
\usepackage{xcolor}         
\usepackage{amssymb}
\usepackage{breqn}
\usepackage{amsthm}
\usepackage{amsmath}
\usepackage{comment}
\usepackage{multirow} 
\usepackage[makeroom]{cancel}
\usepackage{algorithm}
\usepackage{algpseudocode}
\usepackage{tabularx}
\usepackage{cases}
\usepackage{subfig}
\usepackage{graphicx}
\usepackage{comment}
\usepackage{amsmath} 
\usepackage{mathtools}
\usepackage{cases}
\usepackage{makecell}
\usepackage{pdfpages}
\theoremstyle{definition}
\newtheorem{definition}{Definition}

\newtheorem{theorem}{Theorem}[section]
\newtheorem{proposition}{Proposition}[section]
\newtheorem{assumption}{Assumption}[section]
\newtheorem{corollary}{Corollary}[proposition]
\newtheorem{lemma}[theorem]{Lemma}
\newcommand{\RNum}[1]{\uppercase\expandafter{\romannumeral #1\relax}}

\def\one{\mbox{1\hspace{-4.25pt}\fontsize{12}{14.4}\selectfont\textrm{1}}} 

\newcommand{\IndentState}[1][1]{\hspace{#1\algorithmicindent}}

\title{BN-SCAFFOLD: controlling the drift of Batch Normalization statistics in Federated Learning}

\author{%
Gonzalo Iñaki Quintana$^{1,2}$ \And Laurence Vancamberg$^2$ \And Vincent Jugnon$^2$ \And Mathilde Mougeot$^{1,3}$ \And Agnès Desolneux$^{1}$ \AND \\
$^1$ Centre Borelli, CNRS \& ENS Paris-Saclay, F-91190 Gif-sur-Yvette, France \\
$^2$ GE HealthCare, 283 Rue de la Minière, 78530 Buc, France \\
$^3$ ENSIIE, 91000 Evry, France \\
\texttt{\{gonzalo.quintana, mmougeot, agnes.desolneux\}@ens-paris-saclay.fr}\\
\texttt{\{gonzalinaki.quintana, laurence.vancamberg, vincent.jugnon\}@gehealthcare.com}
}

\begin{document}

\maketitle

\begin{abstract}
Federated Learning (FL) is gaining traction as a learning paradigm for training Machine Learning (ML) models in a decentralized way. Batch Normalization (BN) is ubiquitous in Deep Neural Networks (DNN), as it improves convergence and generalization. However, BN has been reported to hinder performance of DNNs in heterogeneous FL. Recently, the FedTAN algorithm has been proposed to mitigate the effect of heterogeneity on BN, by aggregating BN statistics and gradients from all the clients. However, it has a high communication cost, that increases linearly with the depth of the DNN. SCAFFOLD is a variance reduction algorithm, that estimates and corrects the client drift in a communication-efficient manner. Despite its promising results in heterogeneous FL settings, it has been reported to underperform for models with BN. In this work, we seek to revive SCAFFOLD, and more generally variance reduction, as an efficient way of training DNN with BN in heterogeneous FL. We introduce a unified theoretical framework for analyzing the convergence of variance reduction algorithms in the BN-DNN setting, inspired of by the work of Wang et al. 2023, and show that SCAFFOLD is unable to remove the bias introduced by BN. We thus propose the BN-SCAFFOLD algorithm, which extends the client drift correction of SCAFFOLD to BN statistics. We prove convergence using the aforementioned framework and validate the theoretical results with experiments on MNIST and CIFAR-10. BN-SCAFFOLD equals the performance of FedTAN, without its high communication cost, outperforming Federated Averaging (FedAvg), SCAFFOLD, and other FL algorithms designed to mitigate BN heterogeneity.
\end{abstract}

\section{Introduction}
\label{sec:introduction}

Federated Learning (FL) is a learning paradigm in which a collection of clients (which can be mobile devices, hospitals, or organizations) seek to collaboratively train a global model, without sharing the data. This seeks to increase the amount of training data, while mitigating data privacy violations and data collection costs. In its most common setting, the training process is orchestrated by a server. Each global round starts with the server sending the global model weights to all the clients. The clients then perform a pre-determined number of iterations with its local data and send their model versions to the server, which aggregates them to obtain the next global model. This process is repeated iteratively until convergence. Two FL settings are distinguished: cross-device and cross-silo. In cross-device FL, a large number of clients is considered (up to $10^{10}$), which are typically smartphones or IoT devices \cite{kairouz_FL_review}. Clients are unreliable, contain small datasets, and likely participate only once in the training. On the other hand, cross-silo FL features less clients (from 2 to 100 \cite{kairouz_FL_review}), which are more reliable, contain large datasets, and participate regularly in the training. When the data are distributed homogeneously across the clients, also referred to as Independent and Identically Distributed (IID) setting, the Federated Averaging (FedAvg) algorithm has been proven to converge as fast as centralized Stochastic Gradient Descent (SGD). However, in most practical applications data is non-IID and FedAvg performance significantly drops \cite{FedAvgPaperMcMahan, kairouz_FL_review}.  

Batch Normalization (BN) \cite{BatchNormalization} is a widely-used feature normalization strategy, which is present in most of the standard Deep Neural Network (DNN) architectures, in particular in Convolutional Neural Network(CNN)-based vision models, such as ResNet \cite{ResNet}, DenseNet \cite{DenseNet}, and EfficientNet \cite{EfficientNet}. BN reduces internal covariate shift \cite{BatchNormalization}, increases robustness to hyperparameters and initialization \cite{BatchNormalization, Bjorck2018BatchNorm, Yang2019BatchNorm}, provides a smoother optimization landscape \cite{Santurkar2019BatchNorm}, and regularizes the model \cite{BatchNormalization}. However, as this kind of normalization depends on the instances of the batch, it increases the vulnerability of the learning process to data heterogeneity in FL. The drop in performance of DNNs with BN is well described in the literature \cite{FedTAN, FixBN}, and standard FL algorithms that tackle data heterogeneity, such as FedProx \cite{FedProx} and SCAFFOLD \cite{SCAFFOLD}, have been reported to fail in this case \cite{FedTAN}.

In this work, we study the impact of data heterogeneity in models with BN, and focus on the cross-silo FL setting. All clients are assumed to participate in all global rounds, and are assumed stateful, i.e., they keep an internal state. In particular, we seek to revive SCAFFOLD, one of the state-of-the-art FL algorithms for non-IID settings, and extend it to mitigate the impact of heterogeneity on BN.

\subsection{Related work}

\paragraph{Federated Learning.}Several methods have been proposed to mitigate the effects of heterogeneity in FL. Some works propose adding a regularization term that prevents the local models from diverging too much during training, as in FedProx \cite{FedProx}, FedDANE \cite{FedDANE}, and FedDyn \cite{FedDyn}. Other approaches modify the server aggregation step, by matching the weights before averaging using a Bayesian approach \cite{yurochkin2019bayesianMatchedAveraging, wang2020federatedMatchedAveraging} or Optimal Transport \cite{fed_ler_optimal_transport}; by distilling clients models in the server \cite{FedBE, fed_ler_ensemble_distillation_model_fusion}, which requires a server dataset and involves performing additional training steps in the server; or by using adaptive algorithms like Adam \cite{kingma2014adam} or AdaGrad \cite{AdaGrad} to aggregate weights in the server, which has failed to tackle highly heterogeneous settings \cite{AdaptiveFederatedOptimization}. These kinds of approaches are orthogonal to ours, and can be combined with our strategy. Variance reduction algorithms, like SCAFFOLD \cite{SCAFFOLD}, tackle arbitrary levels of client heterogeneity by estimating and controlling client drift using control variates, which define a client state. While SCAFFOLD relies on stateful clients and is only useful in a cross-silo setting, MIME \cite{MIME} extends variance reduction to a cross-device setting with stateless clients. Finally, other works seek to obtain models specialized on each client instead of a single global model, by using Multi-task Learning \cite{MOCHA, NetworledFedMTL}, Meta learning \cite{PerFedAvg, FL-MAML, ARUBA}, or fine-tuning \cite{per_fl_fine_tunning_1, per_fl_fine_tunning_2}. In this work, we focus on standard FL and seek to obtain a single global model. Personalized FL is thus left out of the scope of this research.

\paragraph{Batch Normalization in Federated Learning.}The first works that studied the drop in performance in BN-DNN in Federated Learning proposed to replace BN by other batch-independent normalization techniques \cite{BN_quagmire}, such as Group Normalization (GN) \cite{group_norm}, and, since then, this has been the standard practice in FL \cite{FedTAN}. However, despite GN having been reported to outperform BN for small batches \cite{group_norm}, it is still not clear whether GN can replace BN in all scenarios. Lubana et al. \cite{LubanaNormLayers} compared theoretically and empirically different types of normalization strategies, and concluded that each normalization strategy has its strengths and weaknesses. In particular, they show that deep models with BN tend to have higher feature diversity than with GN and thus converge faster, which was also reported in \cite{FedBE}. Increasing the group size in GN alleviates this problem, but increases training instability. Additionally, BN has been reported to outperform GN for medium to large batches \cite{group_norm, BatchGroupNormalization}. In FL, GN has been shown to only outperform BN in strongly heterogeneous settings \cite{FixBN, mohamad2022fedos}, and is outperformed by BN in many other settings, especially in infrequent communication scenarios \cite{FixBN}. Other normalization techniques, such as Instance Normalization \cite{InstanceNorm} or Layer Normalization \cite{LayerNorm}, are particular cases of GN and thus suffer of the same drawbacks. The comparison between BN and other normalization techniques is left out of the scope of this research.

Other works have explored the local BN approach, which consists in keeping some parameters of BN layers locally, inspired on \cite{AdaBN}, where domain-specific BN layers are considered for Domain Adaptation. SiloBN \cite{SiloBN} keeps the running statistics locally, and exchanges the trainable weight and bias, while FedBN \cite{FedBN} keeps all BN parameters (both running statistics and trainable weight and bias) locally. This might be relevant for learning client-specific models, but not if a single global model is desired, or if each client has a different label distribution \cite{FedTAN}. Besides, when testing on unseen data from a potentially unknown domain, the running statistics have to be estimated from the new data \cite{FedBN}, which might result in inaccurately estimated statistics for small datasets, and impractical for single-input inference.

More recent works have investigated ways to learn a global model with shared BN parameters. Wang et al. \cite{FedTAN} present the first theoretical analysis of the impact of BN layers on FedAvg convergence, showing that it is affected both by the mismatch in the BN statistics, and also in the gradients with respect to those statistics. They propose the FedTAN algorithm, that consists in aggregating the BN statistics from all the clients, as well as the gradients in the backward pass, layer after layer. Although FedTAN has shown promising results in MNIST \cite{lecun2010mnist} and CIFAR-10 \cite{krizhevsky2009learning}, equalling the performance of the centralized model, its number of communication rounds per global step increases linearly with the number of layers of the DNN. This makes it prohibitively costly for deep models in real FL settings, where communication costs are high \cite{FedAvgPaperMcMahan, kairouz_FL_review}. Zhong et al. \cite{FixBN} conduct an extensive empirical analysis and found that the positive impacts of BN occur mostly in the first stages of training, and propose the FixBN algorithm. In FixBN, the training is divided into two stages. In the first stage, the mini-batch statistics are used and the running statistics are updated. After $T^*$ iterations, the running statistics are frozen and are used for normalization, as during inference. However, despite promising results in CIFAR-10, ImageNet \cite{deng2009imagenet}, and Cityscapes \cite{Cordts2016Cityscapes}, this approach lacks theoretical guarantees, and introduces an additional hyper-parameter $T^*$ that can be hard to fine-tune.

\subsection{Contributions}
\label{sec:contributions}

Our main contributions are three-fold:

\begin{itemize}
    \item We introduce a unified theoretical framework (Theorem \ref{theorem:variance_reduc_algos}) for analyzing the convergence of variance reduction algorithms when applied to DNNs with BN, in a general non-convex setting and assuming stateful clients. Theorem \ref{theorem:variance_reduc_algos} extends the work of Wang et al. \cite{FedTAN}, which presented the first known convergence theorem for FedAvg in DNNs with BN.
    
    \item We propose the BN-SCAFFOLD algorithm, an extension of SCAFFOLD that uses control variates to correct client drift in BN statistics. We provide theoretical guarantees of convergence, and show it outperforms FedAvg, SCAFFOLD, and other methods that tackle the non-IID issue in BN, on MNIST and CIFAR-10. BN-SCAFFOLD equals the performance of FedTAN, while keeping the same communication cost as FedAvg.

    \item Using Theorem \ref{theorem:variance_reduc_algos}, we compare several FL algorithms in Table \ref{tab:fl_algs_comparison} based on four figures of merit: convergence rate, communication rounds and overhead, and gradients computed. BN-SCAFFOLD appears to be the best trade-off given those metrics.

\end{itemize}

\section{Batch Normalization in Federated Learning}

\subsection{Batch Normalization}

Batch Normalization normalizes features with an empirical mean and variance, obtained from other instances in the batch. The BN layer is usually located between two subsequent feature extraction layers (usually composed of a convolutional layer, a non-linear activation, a pooling layer, etc.), and assures that the input to the next layer is normalized. Let $ y^l $ be the input to the $l$-th BN layer, and $ x^l $ its output, and thus the input to the $l+1$-th feature extraction layer. During training, the BN applies the following transformation to $ y^l $:

\begin{equation}
    \label{eq:bn}
     x_l = \alpha_l \frac{y_l - \mu_{\mathcal{B}, l} }{\sqrt{ \sigma_{\mathcal{B}, l}^2 +\epsilon }} + \beta_l,
\end{equation}

where $ \mu_{\mathcal{B}, l} $ and $ \sigma_{\mathcal{B},l}^2 $ are the empirical mini-batch statistics (mean and variance) of the $l$-th layer for batch $ \mathcal{B} $, $ \alpha_l $ and $ \beta_l $ are trainable weight and bias, which are included in the model parameters $w$ and are updated with an optimization algorithm like SGD, and $\epsilon $ is a small constant added for numerical stability. Additionally, the mini-batch statistics are accumulated with an Exponential Moving Average (EMA) of momentum $ \rho $

\begin{equation}
    \hat{s}^{t} = \rho \ \hat{s}^{t-1} + (1 - \rho ) s_{ \mathcal{B} }^t,
\end{equation}

where $ s_{ \mathcal{B} }^t \in \{ \mu_{ \mathcal{B} }^t , (\sigma^2_{ \mathcal{B} })^t \} $ and $ \hat{s}^t \in \{ \hat{\mu}^t , \hat{(\sigma^2)}^t \} $ represent, respectively, the mini-batch and running statistics of all the layers, and $t$ is the training iteration. It should be noticed that $ s_{ \mathcal{B} }^t $ depends both on the composition of the batch $ \mathcal{B} $, and on the model parameters $ w^t $. During evaluation, $ \mu_{\mathcal{B}} $ and $ \sigma_{\mathcal{B}}^2 $ are replaced in Eq. \eqref{eq:bn} by their running counterparts, and thus the model is batch-independent and can be used in different contexts. To simplify the notation, in the rest of this article the explicit dependence of the mini-batch statistics on the batch is removed, i.e., $s^t \coloneq s_{ \mathcal{B} }^t$, $\mu^t \coloneq \mu_{ \mathcal{B} }^t$, and $(\sigma^2)^t \coloneq (\sigma^2_{ \mathcal{B} })^t$.

\subsection{Federated Learning setting}

Consider a learning problem with data distributed across $N$ clients and a model parameterized by $ w \in \mathbb{R}^d $. The Global Empirical Risk Minimization (GERM) problem is defined as:

\begin{equation}
\label{eq:GERM}
    \min_{w \in \mathbb{R}^d} F(w) = \min_w \sum_{i=1}^N{ P_i F_i(w)}, \quad \text{where} \quad F_i(w) = \frac{1}{|\mathcal{D}_i|} \sum_{j=1}^{|\mathcal{D}_i|} \mathcal{L}(w;z_j)
\end{equation}

is the Local Empirical Risk of client $i$, $\mathcal{L}$ is a loss function that depends on the model parameters $w$ and on the data-points $z \in \mathcal{Z}$. $\mathcal{D}_i$ is the dataset of client $i$, and $P_i$ indicates an \textit{a priori} client probability, which is typically set to $P_i = |\mathcal{D}_i| / \sum_{j=1}^{N} |\mathcal{D}_j|$. 

If the model is a Neural Network with Batch Normalization layers, then the local loss function and gradients also depend on the BN statistics. In the standard FL setting, each client uses its own local BN statistics, which yields

\begin{equation}
    \nabla_w F_i(w, S_i) = \frac{1}{|\mathcal{D}_i|} \sum_{j=1}^{|\mathcal{D}_i|} \nabla_w \mathcal{L}(w, S_i; z_j) \ \forall i \in [1, N],
\end{equation}

where $S_i$ denotes the layer statistics of client $i$, calculated on the entire local dataset, i.e., $ \mathcal{B} = {\mathcal{D}_i} $.
In Federated Learning, we seek to obtain a set of model parameters $w^*$ that solves the GERM problem of Eq. \eqref{eq:GERM} while keeping the data in the clients. Consider a server-orchestrated FL setting, with $E$ local SGD steps per global step $r$. We denote by $ w^{r,t}_j $, $s^{r,t}_j$, and by $\hat{s}^{r,t}_j$ the local model weights, mini-batch statistics, and running statistics at client $j$, respectively, for global step $r$ and local step $t$. Each global step $r$, starts with the server sending the last global weights, $\bar{w}_{r-1}$, and global running statistics, $\bar{s}_{r-1}$, to every client. After $E$ local steps, each client sends $w^{r,E}_j$ and $\hat{s}^{r,E}_j$ back to the server, which aggregates them for obtaining the global weights and running statistics, typically using a weighted sum with weights $P_i$. This process is repeated for $R$ global steps.

\subsection{The SCAFFOLD algorithm}
\label{sec_2_subsec_scaffold}

SCAFFOLD extends variance reduction algorithms to FL. It uses control variates to estimate and correct the client drift at each client, defined as the deviation between the local and global model update directions. Each client tracks its local model update direction using a control variate, $c_i$. The local control variates are then aggregated to obtain a global control variate, $ c \coloneq \sum_{i=1}^N P_i c_i $, and each client then estimates its drift as the difference between the local and global control variates. This reduces the difference between local models at the time of aggregation, improving global convergence. SCAFFOLD is proven to converge in convex and non-convex settings, with an arbitrary level of client heterogeneity \cite{SCAFFOLD}. The local model update, for client $i$ and at global step $r$, is defined as

\begin{equation}
    w^{r, t}_i = w^{r, t-1}_i - \gamma \left[  g_i(w^{r, t-1}_i; s^{r, t-1}_i) + c^{r-1} - c^{r-1}_i \right],
\end{equation}

where $\gamma$ is the learning rate, $g_i$ the stochastic gradient, $ c^{r-1} $ and $c^{r-1}_i$ the global and local control variates, and $ c^{r-1}_i - c^{r-1} $ estimates the client drift. For calculating the local control variates, two options are provided. In option \RNum{1}, the client control variate is set to the full gradient of the global model calculated with local data, that is, $ c^r_i \coloneq \nabla_w F_i (\bar{w}_{r-1}; S^{r, 0}_i) $. In option \RNum{2}, for $E$ local steps, we have $ c^r_i \coloneq \frac{1}{E} \sum_{t=1}^E g_i (w^{r, 0}_i; s^{r, t}_i)$, which can be obtained using the following recursion, $ c^r_i = c^{r-1}_i - c^{r-1} + \frac{1}{E \gamma} ( \bar{w}_{r-1} - w^{r, E}_i ) $, with no additional calculation (for the proof, refer to Appendix \ref{app:practical_formulas_scaffold_and_bn_scaffold}). As option \RNum{1} can be prohibitively complex when handling large datasets or in frequent communication settings, option \RNum{2} is usually preferred in most practical applications \cite{SCAFFOLD}.
 
\section{Convergence of DNNs with BN in Federated Learning}
\label{sec:convergence_analysis}

\subsection{Assumptions for convergence analysis}
\label{sec:assumptions_conv_analysis}

We make the usual assumptions for analyzing convergence in a general non-convex setting. The global loss function $F$ is assumed lower-bounded by $\underbar{F}$, that is, $ F(w ; S) \geq \ \underbar{F} \ > \ - \infty $. The local loss functions $F_i$ are assumed differentiable and their gradients to be $L$-Lipschitz continuous, that is, $ \| \nabla_w F_i (w' ; S'_i) - \nabla_w F_i (w ; S_i) \| \leq L \| w' - w\| \quad \forall \quad i \in [1, N]$. We consider a stochastic gradient $ g_i (w ; s_i) $ with bounded variance that decreases with the batch size $ | \mathcal{B} | $, i.e., $ \mathbb{E} \left[ \| g_i (w ; s_i) - \nabla_w F_i ( w ; S_i ) \|^2 \right] \leq \sigma^2 = \sigma^2_0 / | \mathcal{B} | $, with $ \mathbb{E} \left[  g_i (w ; s_i) \right] = \nabla_w F_i ( w ; S_i ) $. As proposed by Wang et al. \cite{FedTAN} the bounded gradient dissimilarity assumption, typically used for proving convergence of FL algorithms, is split to isolate the effect of Batch Normalization.

\begin{assumption}[Bounded gradient dissimilarity]
\label{assumption:bounded_local_grad_dev}

There exist constants $B \geq 0$ and $ V \geq 0$ such that

\begin{equation}
    \label{eq:bounded_local_grad_dev}
    \begin{cases}
        \sum_{i=1}^N P_i \ \| \nabla_w F_i (w ; S_i) - \nabla_w F_i (w ; S) \|^2  \leq B^2 \\
        \sum_{i=1}^N P_i \ \| \nabla_w F_i (w ; S) - \nabla_w F (w ; S) \|^2 \leq V^2.
    \end{cases}
\end{equation}

\end{assumption}

In Eq. \eqref{eq:bounded_local_grad_dev}, $B^2$ represents the effect of heterogeneity when using local statistics for BN, while $V^2$ represents the effect of heterogeneity of the data, while using the global statistics $S$ for normalization. They can be combined to obtain a bound on the total gradient dissimilarity, $ \sum_{i=1}^N P_i \ \| \nabla_w F_i (w ; S_i) - \nabla_w F (w ; S) \|^2  \leq 2 \left( B^2 + V^2 \right) $, which corresponds to Assumption (A1) in the SCAFFOLD article \cite{SCAFFOLD}, with $G^2 =  2 \left( B^2 + V^2 \right) $, $ B^2 = 0 $, and $ P_j = 1 / N \quad \forall j \in [1, N] $. 

\paragraph{Additional assumptions for BN-SCAFFOLD.}The mini-batch statistics are assumed to have bounded variance that decreases with the batch size, i.e., $ \mathbb{E} \left[ \| s_i^{r, t} - S^{r,t}_i \|^2 \right] \leq \sigma^2_s = \sigma^2_{s,0} / | \mathcal{B}| $, with $ \mathbb{E} \left[ s_i^{r, t} \right] = S^{r,t}_i $. The Lipschitz continuity of the gradients is extended to the BN statistics.


\begin{assumption}[Lipschitz continuity for the statistics]
\label{assumption:l_lipschitz_continuity_statistics}

The gradients of the local functions $ F_i $ are assumed to be Lipschitz continuous with respect to the statistics with Lipschitz constant $J$. In addition, the difference between two statistics is assumed to be bounded by the product of a constant $M \geq 0$ and the difference between the model parameters from which they were obtained. That is,

\begin{equation}
\begin{cases}
    \| \nabla_w F_i (w ; S'_i) - \nabla_w F_i (w ; S''_i) \| \leq J \| S'_i - S''_i \| \\
    \| S_i' - S_i'' \| \leq  M \| w' - w'' \|,
\end{cases}    
\end{equation}

where $S'_i$ and $S''_i$ are the BN statistics obtained from the local data at client $i$, with model parameters $w'$ and $w''$, respectively. For BN-SCAFFOLD option \RNum{2}, it is also required that $ M^2 J^2 < L^2 / 3 $, which means that the local functions are smoother relative to the BN statistics than to the model weights.

\end{assumption}

\subsection{Convergence theorem}

In this work, we focus on variance reduction algorithms, which we formally define in Definition \ref{def:variance_reduc_algos}.

\begin{definition}
    
\label{def:variance_reduc_algos}

The variance reduction family of algorithms is defined by the following local updates:

\begin{numcases}
    \\
    \tilde{s}^{r, t}_i = \phi (s^{r, t}_i, k^{r-1}_i, k^{r-1}) \label{eq:var_reduc_algos_def_stats_correction} \\
    w^{r, t}_i = w^{r, t-1}_i - \gamma \left[  g_i(w^{r, t-1}_i; \tilde{s}^{r, t}_i) + c^{r-1} - c^{r-1}_i \right] \label{eq:var_reduc_algos_def_gradient_descent} \\
    \hat{s}^{r,t}_i = (1 - \rho) \hat{s}^{r,t-1}_i + \rho \ \tilde{s}^{r,t}_i \label{eq:var_reduc_algos_def_running_stats_corr} \\
    c^r_i = \Psi(\{ g_i(w^{r,t}_i; \tilde{s}^{r, t}_i ) \}_{t=1}^E, c^{r-1}_i, c^{r-1} ) \label{eq:var_reduc_algos_def_c_update} \\
    k^r_i = \Phi(\{\tilde{s}^{r, t}_i \}_{t=1}^E, k^{r-1}_i, k^{r-1}) \label{eq:var_reduc_algos_def_k_update},
\end{numcases}

where $\gamma$ is the learning rate, $\rho \in [0,1]$ the BN momentum, $c^r$ and $c^r_i$ are the global and local control variates for the gradient (introduced in SCAFFOLD \cite{SCAFFOLD}), $k^r$ and $k^r_i$ are the global and local control variates for the statistics, $ \phi $ defines the statistics correction, and $\Psi $ and $ \Phi $ define the gradient and statistics control variates updates, respectively.

\end{definition}

Definition \ref{def:variance_reduc_algos} defines variance reduction algorithms in a generic way, with the aim of having a general convergence theorem that can encompass a wide range of algorithms. Additionally, Definition \ref{def:variance_reduc_algos} also contains non-variance reduction algorithms, such as FedAvg and FedTAN. In Eq. \eqref{eq:var_reduc_algos_def_stats_correction} the corrected mini-batch statistics $\tilde{s}^{r,t}_i$ are obtained from the original mini-batch statistics $s^{r,t}_i$ and the statistics control variates, $k^{r-1}$ and $k^{r-1}_j$, by the application of a correction function $\phi(., k_i, k): \mathcal{S} \rightarrow  \mathcal{S}$. Eq. \eqref{eq:var_reduc_algos_def_gradient_descent} is the local model update of SCAFFOLD, where the local gradient $g_i$ is obtained with the corrected mini-batch statistics, and Equation \ref{eq:var_reduc_algos_def_running_stats_corr} 
 represents the update of the local running statistics with the corrected mini-batch statistics. Finally, the local gradient and statistics control variates are updated in Equations \eqref{eq:var_reduc_algos_def_c_update} and \eqref{eq:var_reduc_algos_def_k_update}, by using the generic functions $ \Psi $ and $ \Phi $. The global gradient and statistics control variates are the weighted sum of their local counterparts with weights $P_i$, as in SCAFFOLD.
 Definition \ref{def:variance_reduc_algos} enables to obtain different FL algorithms by specifying the generic functions $\phi$, $ \Psi $, and $ \Phi $. For obtaining FedAvg we set $c^r_i = 0 $, $k^r_i = 0 $, and $ \tilde{s}^{r,t}_i = s^{r,t}_i $. Setting $c^r_i = 0 $, $k^r_i = 0 $, and $ \tilde{s}^{r,t}_i = s^{r,t} \one_{\{t=0\}} + s^{r,t}_i \one_{\{t \geq 1 \}} $, with $ \one_{\{ . \} }$ the indicator function, gives the FedTAN algorithm, in which only the statistics of the first local steps are adapted. For SCAFFOLD, we have $k^r_i = 0 $, $ \tilde{s}^{r,t}_i = s^{r,t}_i $, and $ c^{r}_i $ obtained as explained in Section \ref{sec_2_subsec_scaffold} with options \RNum{1} and \RNum{2}. 
We now introduce the Theorem that enables to calculate the convergence rate of algorithms that can be obtained from Definition \ref{def:variance_reduc_algos}.

\begin{theorem}
\label{theorem:variance_reduc_algos}

The convergence rate of the family of algorithms defined in Definition \ref{def:variance_reduc_algos} for a general non-convex objective function $F$, and under the assumptions discussed in Section \ref{sec:assumptions_conv_analysis}, is given by

\begin{dmath}
\label{eq:convergence_var_reduction_algorithms_async}
\min_r \mathbb{E} \left[ \| \nabla_w F ( \bar{w}_{r-1} ; S^{r,0}) \|^2 \right]
\leq \frac{1}{R} \sum_{r=1}^R \mathbb{E} \left[ \| \nabla_w F_i ( \bar{w}_{r-1}; S^{r,0}) \|^2 \right]
\leq \frac{ \mathcal{T} }{1 - 2 \delta_{E, \gamma, L}},
\end{dmath}

where $ \mathcal{T} = \mathcal{T}_1 + \mathcal{T}_2 + \mathcal{T}_3 + \mathcal{T}_4 +  \mathcal{T}_5 - \mathcal{T}_6 $ with

\begin{dmath}
\begin{cases}
\mathcal{T}_1 = \frac{2}{\gamma R E} \left[ F( \bar{w}_0 ; S^{1,0} ) - \underbar{F} \right] \\[2pt]
\mathcal{T}_2 = \frac{8}{R} \delta_{E, \gamma, L} \sum_{j=1}^N P_j \| \nabla_w F_j ( \bar{w}_{0} ; \tilde{S}^{1, 0}_j )- c^{0}_j \|^2 \\[2pt]
\mathcal{T}_3 = \frac{2}{R} ( 1 + 2 \delta_{E, \gamma, L} ) \sum_{r=1}^R \sum_{j=1}^N P_j \mathbb{E} \left[ \| \nabla_w F_j ( \bar{w}_{r-1} ; S^{r,0}) - \nabla_w F_j ( \bar{w}_{r-1} ; \tilde{S}^{r,0}_j ) \|^2 \right] \\[2pt]
\mathcal{T}_4 = \frac{16}{R} \delta_{E, \gamma, L} \sum_{r=1}^R \sum_{j=1}^N P_j \mathbb{E} \left[ \| \nabla_w F_j ( \bar{w}_{r -1} ; \tilde{S}^{r, 0}_j ) - c^r_j \|^2 \right] \\[2pt]
\mathcal{T}_5 = ( 1 + 2 \delta_{E, \gamma, L} ) \sigma^2 \\[2pt]
\mathcal{T}_6 = \frac{1}{R E} (1 - \gamma L E - 16 L^2 E^2 \gamma^2 \delta_{E, \gamma, L} ) \sum_{r=1}^R \sum_{t=1}^E \mathbb{E} \left[ \| \sum_{j=1}^N P_j \ g_j( w_j^{r, t-1} ; \tilde{s}_j^{r, t-1}) \|^2 \right],
\end{cases}
\end{dmath}

and with $ \delta_{E, \gamma, L} \coloneq \frac{4 E^2 \gamma^2 L^2}{1 - 8 E^2 \gamma^2 L^2} $ and $ \gamma < \frac{1}{\sqrt{12} L E}$. Theorem \ref{theorem:variance_reduc_algos} is proven in Appendix \ref{app:proof_conv_theorem}, and the fundamental Lemmas used in its proof are presented in Appendix \ref{app:fundamental_lemmas}.

\end{theorem}

Theorem \ref{theorem:variance_reduc_algos} is applied to specific algorithms, like FedAvg or BN-SCAFFOLD, by setting the generic functions $\phi$, $\Phi$, and $\Psi$ and calculating terms $ \mathcal{T}_1 $ to $ \mathcal{T}_6 $. $\mathcal{T}_1$ depends on the weight initialization; $\mathcal{T}_2$ on the mismatch between the clients initial control variates and the initial full gradient; $ \mathcal{T}_3 $ on the mismatch between the local gradients with the global Batch Norm statistics, and the local gradients with the corrected statistics; $\mathcal{T}_4 $ on the mismatch between the local gradients and the local control variates; and $ \mathcal{T}_5 $ on the stochastic gradient variance. If the last term $ \mathcal{T}_6 $ is positive, it can be discarded from the upper bound, which can be accomplished by setting an upper bound on the learning rate $\gamma$. In most cases, $\mathcal{T}_6$ is usually combined with other terms before being discarded, giving a tighter bound. For instance, setting $ c^r_j = 0 $ and $ \tilde{S}^{r,0}_j = S^{r,0}_j $ gives the bound for FedAvg.
Theorem \ref{theorem:variance_reduc_algos} makes explicit the bias introduced by the gradient dissimilarity associated to the Batch Norm statistics. If no statistic correction is applied, that is $ \tilde{S}^{r,0}_j = S^{r,0}_j $, then  under Assumption \ref{assumption:bounded_local_grad_dev} we have $ \mathcal{T}_3  = 2 (1 + 2 \delta_{E, \gamma, L} ) B^2 $, a term that does not decrease with increasing $R$.

\section{The BN-SCAFFOLD algorithm}
 
The BN-SCAFFOLD algorithm is proposed to tackle the convergence issues introduced by the Batch Norm statistics in Federated Learning, made explicit in term $ \mathcal{T}_3 $ of Theorem \ref{theorem:variance_reduc_algos}, in a communication-efficient way. The concept of control variates used for the gradient in SCAFFOLD is extended to the BN statistics by the introduction of the global and local statistics control variates, denoted as $k^r$ and $k^r_i $, which are used to linearly correct the local statistics. That is, $\tilde{s}^{r,t}_i = s^{r,t}_i - k^{r-1}_i + k^{r-1}$. As in SCAFFOLD, two options are provided for the computation of the local statistics control variates. In option \RNum{1} we have $ k^r_i \coloneq S^{r, 0}_i$, which involves computing the statistics with the entire data of each client. In option \RNum{2}, the local control variates are obtained from the running statistics, that are accumulated during the $E$ local steps with no additional computations, using the following relationship:

 \begin{equation}
 \label{eq:bn_scaffold_option_2_pratical_update}
    k^r_i \coloneq \frac{1 - \rho}{1 - \rho^E} \sum_{t=0}^{E-1} \rho^{E-1-t} s^{r,t}_i = k^{r-1}_i - k^{r-1} + \frac{1}{1 - \rho^E} (\hat{s}^{r,E-1}_i - \rho^E \hat{s}^{r,0}_i ),
\end{equation}

where, in the second equality, $ k^r_i $ is obtained from the initial and the final running statistics, $ \hat{s}^{r,0}_i $ and $ \hat{s}^{r,E-1}_i $, respectively. A detailed description of the BN-SCAFFOLD algorithm is depicted in Algorithm \ref{alg:algorithm_bn_scaffold}. In Appendix \ref{app:practical_formulas_scaffold_and_bn_scaffold}, the proof of the second equality is given .

\begin{algorithm}[H]
\caption{BN-SCAFFOLD}
\label{alg:algorithm_bn_scaffold}
\begin{algorithmic}[1]
\State \textbf{Server inputs:} number of global steps $R$, number of clients $N$, initial model weights $w^0$, initial running statistics $\hat{s}_0$, initial control variates for the gradient, $c^0$, and for the statistics $k^0$.
\State \textbf{Client $i$ inputs:} initial control variates for the gradient, $c_i^0$, and the statistics, $k_i^0$, number of local steps $E$, local learning rate $\gamma$.
\For{$r = 1, ..., R$}
\State \textbf{communicate($\bar{w}_{r-1}$, $\bar{s}_{r-1}$, $c^{r-1}$, $k^{r-1}$) to all clients}
    \For{each client $i$ in parallel}
    \State $w^{r, E}_i$, $\hat{s}^{r, E}_i$, $ c^r_i \gets$, $ k^r+_i \gets$ \textbf{ClientUpdate}($\bar{w}_{r-1}$, $\bar{s}_{r-1}$, $c^{r-1}$, $k^{r-1}$)
    \EndFor
\State   $ \{ \bar{w}_r ; \bar{s}_r ; c^r ; k^r \} \gets \sum_{i=1}^N P_i \{ w^{r, E}_i ;\hat{s}^{r,E}_i; c^r_i; k^r_i \} $
\EndFor
\State
\State \textbf{ClientUpdate($\bar{w}_{r-1}$, $\bar{s}_{r-1}$, $c^{r-1}$, $k^{r-1}$}\textbf{):}
\IndentState
    \State $w^{r,0}_i \gets \bar{w}_{r-1}$, $\hat{s}^{r, 0}_i \gets \bar{s}_{r-1}$
    \For{local step $t = 0, \cdots, E-1$}
    \State $w^{r,t+1}_i \gets w - \gamma \left[ g_i(w^{r,t}_i; s^{r, t}_i - k_i^{r-1} + k^{r-1}) - c_i^{r-1} + c^{r-1} \right] $ 
    \State $\hat{s}^{r,t+1}_i \gets (1 - \rho ) \hat{s}^{r, t}_i + \rho ( s^{r, t}_i  - k^{r-1}_i + k^{r-1} ) $
    \EndFor
    \State $c_i^r \gets \nabla_w F_i (w^{r,0}_i; S^{r,0}_i) $ (option \RNum{1}), $c_i^r \gets c_i^{r-1} - c^{r-1} + \frac{1}{E \gamma } (w^{r,0}_i - w^{r,E}_i)$ (option \RNum{2})
    \State $k_i^r \gets S^{r,0}_i $ (option \RNum{1}), $k_i^r \gets k_i^{r-1} - k^{r-1} + \frac{1}{1 - \rho^E} (\hat{s}^{r,E}_i - \rho^E \bar{s}^{r,E}_i)$ (option \RNum{2})
    \State \textbf{return} ($w^{r,E}_i$, $\hat{s}^{r,E}_i$, $c_i^r$, $k_i^r$) to server
    \end{algorithmic}
\end{algorithm}

\section{Comparison of FL algorithms}
\label{sec:conv_rate_of_algos}

In this Section, we use Theorem \ref{theorem:variance_reduc_algos} to obtain the convergence rate of FedAvg, SCAFFOLD (options \RNum{1} and \RNum{2}), BN-SCAFFOLD (options \RNum{1} and \RNum{2}), and FedTAN. The details of the application of Theorem \ref{theorem:variance_reduc_algos} to each algorithm are provided in Appendices \ref{app:fed_avg_convergence_proof}, \ref{app:scaffold_convergence_proof}, \ref{app:bn_scaffold_convergence_proof}, and \ref{app:fed_tan_convergence_proof}.

Table \ref{tab:fl_algs_comparison} compares the considered FL algorithms on four figures of merit: convergence rate, communication rounds per local step, communication overhead per global step (measured in terms of the cardinality of the model), and gradients computed for each local step. These characteristics summarize the advantages, disadvantages, and trade-offs made in each FL algorithm. For FedAvg and SCAFFOLD, the presence of BN results in a constant term in the convergence rate, which is proportional to the gradient dissimilarity, and that does not decrease with the number of global steps $R$. FedTAN removes this term at the expense of introducing a number of communication rounds that grows linearly with the model depth $W_D$. In contrast, BN-SCAFFOLD removes it while keeping the same number of communication rounds as FedAvg and SCAFFOLD, and only adding a marginal increase in the communication overhead, proportional to the statistics parameter vector size $ | S | $. However, in many DL architectures, $ | S | $ is a relatively small fraction of the model parameter vector size $ | W |$. For instance, in ResNet-18 $| W | > 11.18 \times 10^{6} $ and $ | S | = 9.6 \times 10^{3} $, and thus BN-SCAFFOLD represents an increase of 4.3\% in communication overhead with respect to SCAFFOLD. We observe that options \RNum{1} of SCAFFOLD and BN-SCAFFOLD have better convergence rates than options \RNum{2}, albeit with a number of gradients computed that increases linearly with the dataset size $ | \mathcal{D} | $. Finally, we remark that in SCAFFOLD and BN-SCAFFOLD with option \RNum{2}, the stochastic gradient variance term is improved for an increasing number of local steps $E$, as the variance of the estimated gradient and BN statistics update direction is reduced. We believe the BN-SCAFFOLD algorithm represents the best trade-off in the four considered figures of merit.

\begin{table}[htbp]
\scriptsize 
\caption{Comparison of the different algorithms with $N$ clients, $E$ local updates, $R$ global steps, $W_D$ the model depth, and trained with a learning rate $ \gamma = \frac{\gamma_0}{L E} $ for BN-SCAFFOLD option \RNum{2}, and $ \gamma = \frac{\gamma_0}{L E \sqrt{R}} $ for the rest of the algorithms. The learning rates were set this way so as to obtain the best convergence rate for each algorithm. Communication rounds and gradients computed are indicated in a per local step basis. Communication overhead is indicated per global step. $F_0$ is the loss function at the initial model $w_0$, and $\nabla_w F_0^2$, $\Delta^2 c^0 $, $\Delta^2 k^0 $ the squared norm of its gradient, and control variates. $\Omega \coloneq \frac{L^2 - M^2 J^2}{L^2 + M^2 J^2} > 0 $ is the relative difference between the gradients and statistics Lipschitz constants. }
\label{tab:fl_algs_comparison}
\centering
\renewcommand{\arraystretch}{1.5} 
\begin{tabular}{p{1.7cm}p{5.9cm}p{1.5cm}p{1.4cm}p{1.4cm}}
\toprule
FL algorithm & Convergence rate & Communication rounds & Communication overhead &  Gradients computed \\
\midrule
FedAvg \cite{FedAvgPaperMcMahan} & $ \kern-1em \mathcal{O} ( \frac{L}{\sqrt{R}} \left[ F_0 - \underbar{F} \right] + \frac{\nabla F_0^2}{R^2} + B^2 + \frac{V^2}{R} + \frac{\sigma_0^2}{| \mathcal{B}|}  )$ & $ 2 N / E$  & $ | W | + | S | $ &  $ N | \mathcal{B} | $  \\
\midrule
SCAFFOLD \cite{SCAFFOLD} & & \\
\quad option \RNum{1} & $ \kern-1em \mathcal{O} \left( \frac{L}{\sqrt{R}} \left[ F_0 - \underbar{F} \right] + B^2 + \frac{\Delta^2c^0}{R^2} + \frac{\sigma_0^2}{| \mathcal{B}|} \right) $ & $ 2 N / E$  & $ 2 | W | + | S | $ &  $ N | \mathcal{B} | + | \mathcal{D} | / E $  \\
\quad option \RNum{2} & $\kern-1em  \mathcal{O} \left( \frac{L}{\sqrt{R}} \left[ F_0 - \underbar{F} \right] + B^2 + \frac{\Delta^2c^0}{R^2} + (1 +  \frac{1}{R} + \frac{1}{RE}) \frac{\sigma_0^2}{| \mathcal{B}|} \right) $ & $ 2 N / E$ & $ 2 | W | + | S | $ &  $ N | \mathcal{B} | $ \\
\midrule
BN-SCAFFOLD & & \\
\quad option \RNum{1} & $ \kern-1em \mathcal{O} \left( \frac{L}{\sqrt{R}} \left[ F_0 - \underbar{F} \right] + \frac{\Delta^2c^0}{R^2} + \frac{\sigma_0^2}{| \mathcal{B}|} \right) $ & $ 2 N / E$ & $ 2 | W | + 2 | S | $ &  $ N | \mathcal{B} | + | \mathcal{D} | / E $ \\
\quad option \RNum{2} & \makecell[l]{ $ \kern-1em \mathcal{O} \Bigl( \frac{L \Omega}{R} \left[ F_0 - \underbar{F} \right]+ \frac{J^2 + \Omega}{R} \Delta^2k^0 + \frac{1 + \Omega}{R} \Delta^2c^0 $ \\ $ \qquad + (\frac{1}{E} +\Omega) \frac{\sigma^2_0}{| \mathcal{B} |} + J^2 \Omega \frac{\sigma^2_s,0}{| \mathcal{B} |} \Bigr) $} & $ 2 N / E$ & $ 2 | W | + 2 | S | $ & $ N | \mathcal{B} | $ \\
\midrule
FedTAN \cite{FedTAN} & $ \kern-1em \mathcal{O} ( \frac{L}{\sqrt{R}} \left[ F_0 - \underbar{F} \right] + \frac{\nabla F_0^2}{R^2} + \frac{V^2}{R} + \frac{\sigma_0^2}{| \mathcal{B}|}  ) $ & $ (2 + 6 W_D) \frac{N}{E} $ &  $ 2 | W | + 4 | S | $ & $ N | \mathcal{B} | $ \\
\bottomrule
\end{tabular}
\end{table}

\section{Numerical experiments}
\label{sec:experiments}

We conduct experiments for classification on two distinct types of images: natural images (MNIST and CIFAR-10) and mammography images. For mammography images, two patch datasets were considered: a dataset of extracted patches from clinical mammography images, and a dataset synthetic patches. The patch extraction process and clinical mammography dataset were described in previous publications \cite{Quintana2023, Quintana2024}. The synthetic patches were generated using Gaussian textures and simple lesion (mass and micro-calcification) simulation. The details of Gaussian texture generation can be found in \cite{galerne2010random}.

\subsection{Natural images}

We run experiments on MNIST and CIFAR-10 using ResNet-18, in a cross-silo setting with 2 and 5 clients. For constructing the training and validation sets, 5-fold cross validation was used on the official train subsets. A skewed distribution of data labels was generated in the clients, by splitting the training and validation subsets into the clients depending on their labels. Reported results are the mean on the test set of the models trained on each of the folds, with the 95\% Confidence Interval (CI) and p-value obtained with the two tailed Welch's t-test, where normally distributed errors were assumed. All p-values are calculated with respect to BN-SCAFFOLD. The training of all FL models, with the exception of FedTAN, was conducted in a realistic FL setting, with each client in a physically separated computing station, using the open source FL framework Nvidia Flare \cite{nvflare}. In all experiments, SCAFFOLD and BN-SCAFFOLD use option \RNum{2}.

\begin{figure}
     \centering
     \subfloat[][Impact of client heterogeneity, with $E=10$.]{\label{fig:bn_scaffold_vs_baseline_heterogeneity} \includegraphics[width=0.5\textwidth]{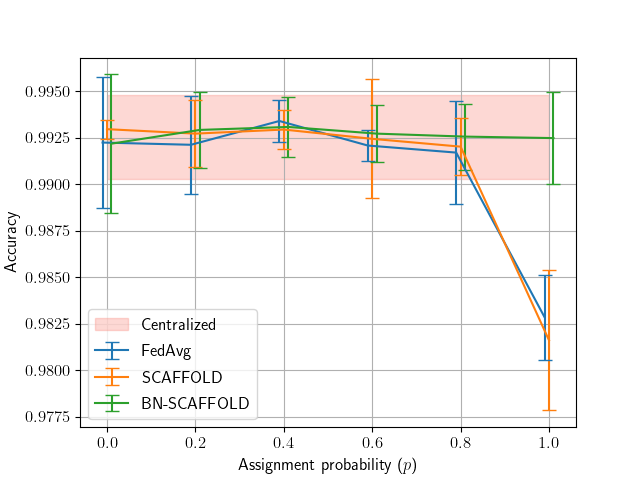}\label{<figure1>}}
     \subfloat[][Impact of local steps per round, with $p=1$.]{\label{fig:bn_scaffold_vs_baseline_local_steps} \includegraphics[width=0.5\textwidth]{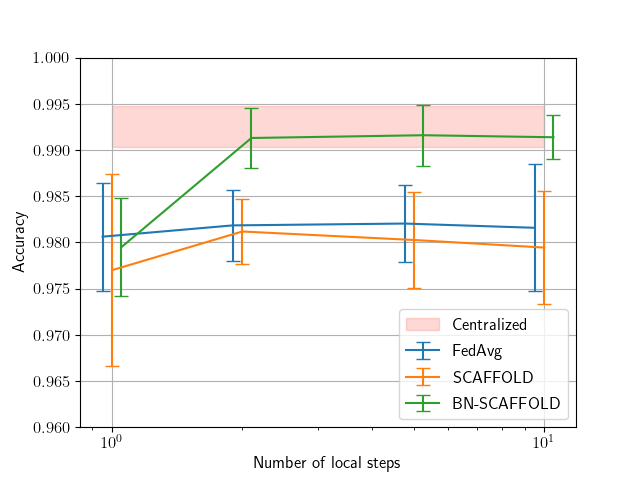}\label{<figure2>}}
     \caption{Comparison of FedAvg, SCAFFOLD, and BN-SCAFFOLD in MNIST with $N$=2 clients.}
     \label{fig:bn_scaffold_vs_baseline}
\end{figure}

Figure \ref{fig:bn_scaffold_vs_baseline} compares the performance of BN-SCAFFOLD with the two baselines, FedAvg and SCAFFOLD, in MNIST with $N=2$ clients, and in a high frequency communication setting (local steps per round $E \leq 10$). The models were trained using SGD with a learning rate of $\gamma = 0.5$, and fixed budget of 3500 iterations, or $R=3500/E$ global steps. To obtain different levels of heterogeneity, an assignment probability $p \in [0.5, 1] $ is used to assign images with label $y \in \{0, 1, 2, 3, 4 \}$ to client 1, and images with label $y \in \{5, 6, 7, 8, 9 \}$ to client 2. When $p=0.5$, client data are homogeneous or IID, and when $p=1$, data are strongly heterogeneous, with no overlap between the two label distribution. Figure \ref{fig:bn_scaffold_vs_baseline_heterogeneity} shows the impact of different levels of heterogeneity on model Accuracy, with $E=10$. The performance of FedAvg and SCAFFOLD is relatively stable in low and moderate heterogeneity settings, but drops significantly when heterogeneity is strong ($p > 0.9$). In contrast, the Accuracy of BN-SCAFFOLD is unaffected by heterogeneity, and matches the performance of the centralized setting. Figure \ref{fig:bn_scaffold_vs_baseline_local_steps} shows the impact of the number of local steps $E \in \{ 1, 2, 5, 10 \}$, in the strongly heterogeneous setting ($p = 1$). It can be noticed that, while BN-SCAFFOLD outperforms FedAvg and SCAFFOLD when $E \in \{ 2, 5, 10 \}$ and matches the Accuracy of the centralized setting, its performance drops when $E=1$. This can be explained by considering the $\sigma^2_0$ term of the BN-SCAFFOLD convergence rate in Table \ref{tab:fl_algs_comparison}, that shows decreased convergence with smaller $E$.

\begin{table}
\footnotesize
\caption{Test accuracy in MNIST and CIFAR-10. Reported results are the mean of the models trained on 5 folds, with the 95\% CI and the Welch's t-test p value indicated. n.s.: not significant ($p > 0.05$).}
\label{tab:fl_algos_results}
\begin{tabular}{llll}
\toprule
            &                       MNIST (2 clients) & MNIST (5 clients) & CIFAR-10 (2 clients) \\
\midrule
    Centralized &       0.993 $\pm$ 0.002 (n.s.) & 0.987 $\pm$ 0.003 ($<$ 0.01) &  0.820 $\pm$ 0.012 ($<$ 0.001) \\
\midrule
        FedAvg \cite{FedAvgPaperMcMahan} &  0.983 $\pm$ 0.002 ($<$ 0.001) &  0.931 $\pm$ 0.011 ($<$ 0.001) & 0.724 $\pm$ 0.015 ($<$ 0.001) \\
       FedBN \cite{FedBN} &  0.985 $\pm$ 0.003 ($<$ 0.001) & 0.936 $\pm$ 0.022 ($<$ 0.001) &  0.719 $\pm$ 0.021 ($<$ 0.001) \\
      SiloBN \cite{SiloBN} &  0.984 $\pm$ 0.002 ($<$ 0.001) & 0.935 $\pm$ 0.019 ($<$ 0.001) &  0.716 $\pm$ 0.006 ($<$ 0.001) \\
      FixBN \cite{FixBN} & 0.980 $\pm$ 0.003 ($<$ 0.001) & \textbf{0.982 $\pm$ 0.003 (n.s.)} &   0.754 $\pm$ 0.017 ($<$ 0.01) \\
      FedTAN \cite{FedTAN} &       \textbf{0.992 $\pm$ 0.003 (n.s.)} & \textbf{0.984 $\pm$ 0.003 (n.s.)} &       \textbf{0.778 $\pm$ 0.010 (n.s.)} \\
\midrule
  SCAFFOLD \cite{SCAFFOLD} &  0.982 $\pm$ 0.004 ($<$ 0.001) & 0.934 $\pm$ 0.007 ($<$ 0.001) &  0.685 $\pm$ 0.027 ($<$ 0.001) \\
  FedBN+SCAFFOLD &  0.983 $\pm$ 0.003 ($<$ 0.001) & 0.942 $\pm$ 0.011 ($<$ 0.001) &  0.697 $\pm$ 0.009 ($<$ 0.001) \\
 SiloBN+SCAFFOLD &  0.983 $\pm$ 0.004 ($<$ 0.001) & 0.940 $\pm$ 0.011 ($<$ 0.001) &  0.693 $\pm$ 0.027 ($<$ 0.001) \\
FixBN+SCAFFOLD & 0.979 $\pm$ 0.007 ($<$ 0.001) & 0.934 $\pm$ 0.018 ($<$ 0.001) & 0.718 $\pm$ 0.043 ($<$ 0.001) \\
\midrule
BN-SCAFFOLD (ours) &       \textbf{0.992 $\pm$ 0.002 (n.a.)} & \textbf{0.983 $\pm$ 0.004 (n.a.)} &       \textbf{0.776 $\pm$ 0.021 (n.a.)} \\
\bottomrule
\end{tabular}
\end{table}

Table \ref{tab:fl_algos_results} compares BN-SCAFFOLD in MNIST and CIFAR-10 with FedAvg, SCAFFOLD, FedTAN, and other state-of-the-art algorithms that tackle heterogeneity in BN-DNNs: FedBN, SiloBN, FixBN, and FedTAN. In addition, FedBN, SiloBN, and FixBN are combined with SCAFFOLD. FedTAN and SCAFFOLD are not combined, as there is not a direct way to integrate them. In MNIST, models were trained during 3500 ($N=2$) and 7000 ($N=5$) iterations with $E=10$. In CIFAR-10, they were trained during $4 \times 10^4$ iterations with $E=100$. The detailed list of hyper-parameter is given Appendix \ref{app:experimental_settings_details}. It can be seen that BN-SCAFFOLD and FedTAN outperform all the other algorithms, with the exception of FixBN in MNIST with 5 clients, and match the centralized performance in MNIST with 2 clients. However, FedTAN requires a higher number of communication steps than BN-SCAFFOLD. We remark that combining FedBN, SiloBN, or FixBN with SCAFFOLD does not increase performance with respect to plain SCAFFOLD.

\subsection{Clinical mammography patches}

For both clinical and synthetic mammography patches, a FL setting of two clients with image style heterogeneity due to different post-processing algorithms was considered. Specifically, the LUT transform \cite{Quintana2024} was used to introduce this type of heterogeneity: some images had the LUT applied, while others did not. For the two datasets, the DenseNet-121 developed in \cite{Quintana2023, Quintana2024} was used as classification model. The classification performance is given in terms of three metrics: the accuracy, the mean one-vs-one AUC, and the mean one-vs-rest AUC.

\begin{figure}[h]
  \centering
  \subfloat[\centering \small Accuracy. \label{fig:2-acc_patch_classifier_ddsm_fl}]{\includegraphics[width=0.32\textwidth, keepaspectratio]{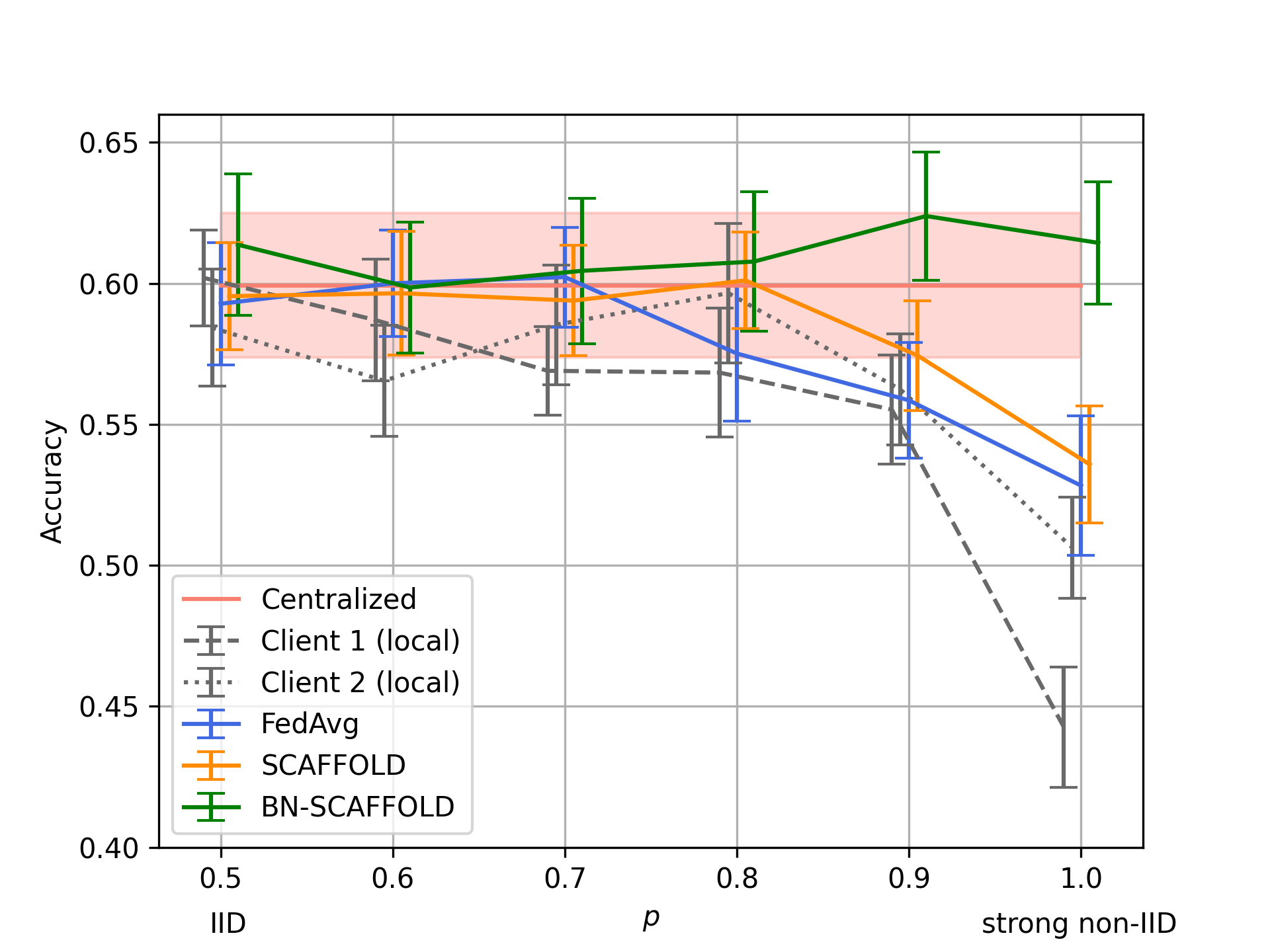}}
  \subfloat[\centering  \small Mean one-vs-one AUC. \label{fig:2-auc_one_vs_one_patch_classifier_ddsm_fl}]{\includegraphics[width=0.32\textwidth, keepaspectratio]{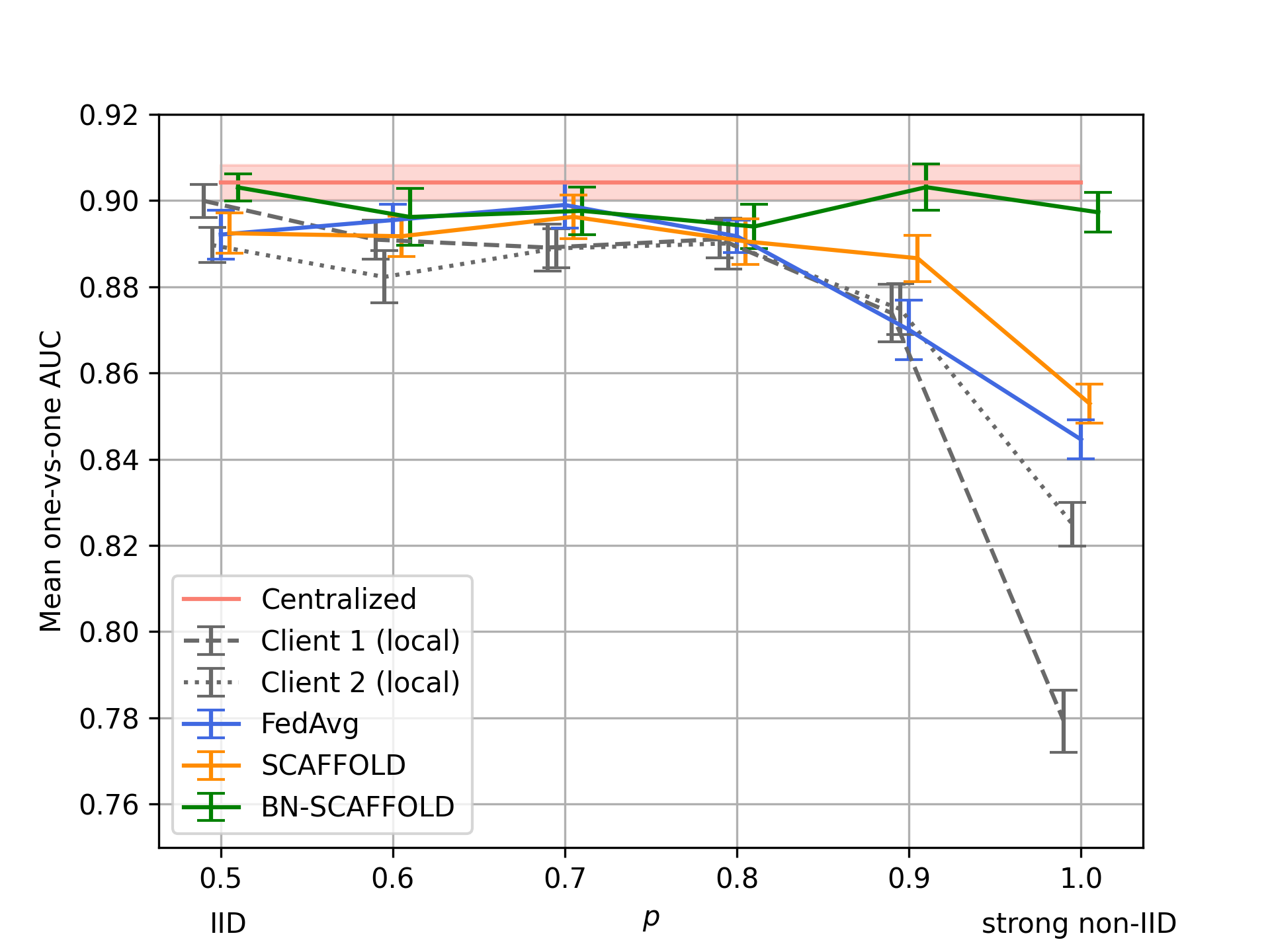}}
  \subfloat[\centering \small Mean one-vs-rest AUC. \label{fig:2-auc_one_vs_rest_patch_classifier_ddsm_fl}]{\includegraphics[width=0.32\textwidth, keepaspectratio]{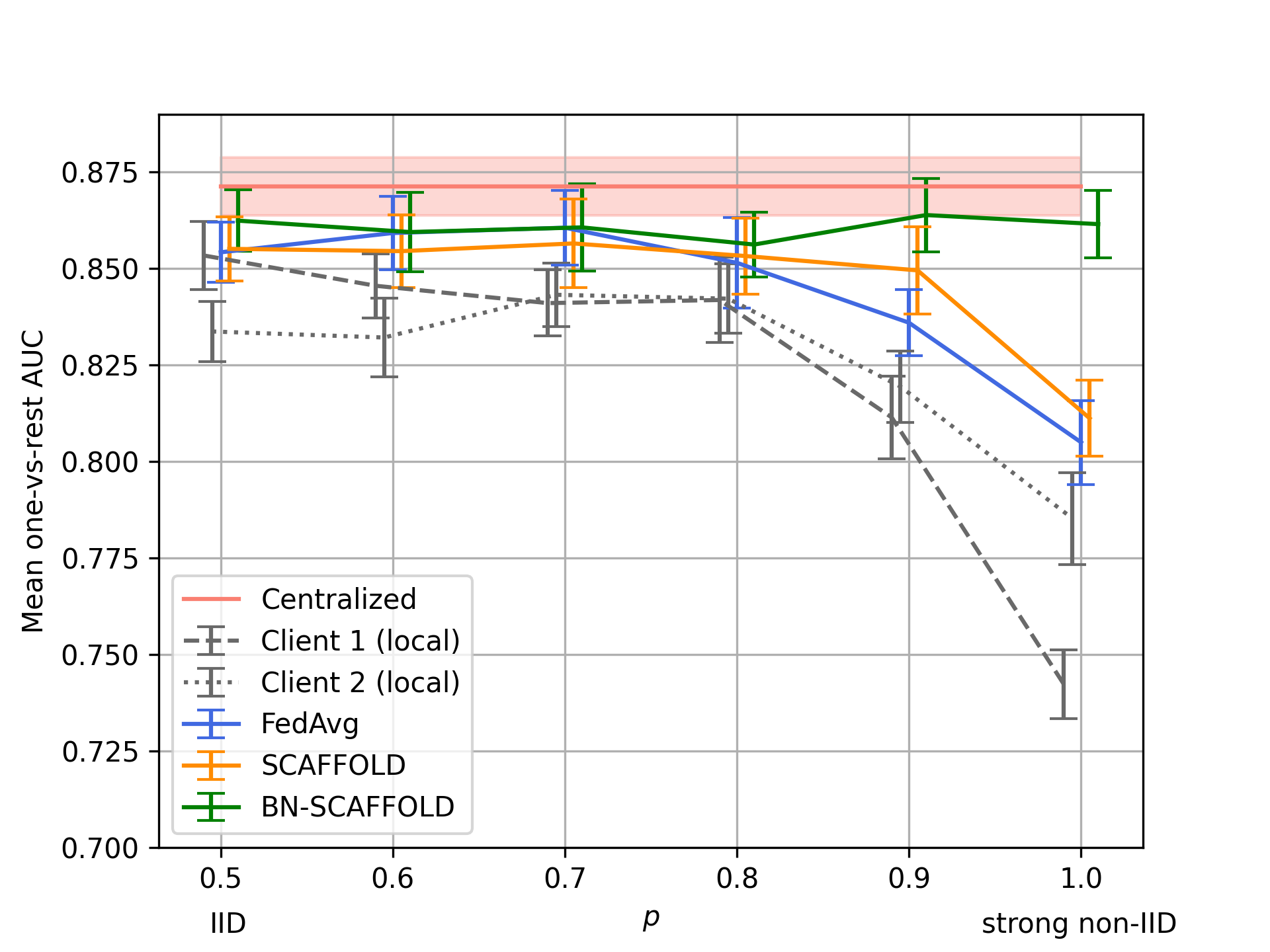}}
\caption{\small Classification performance with 95\% CI for different degrees of heterogeneity, using original x-ray image.} 
\label{fig:2-classif_perf_patch_classifier_ddsm_fl}
\end{figure}

In Figure \ref{fig:2-classif_perf_patch_classifier_ddsm_fl}, FedAvg, SCAFFOLD, and the proposed BN-SCAFFOLD are compared for 5-class classification on the test set. The performance of a centralized training, and of local training in each of the clients are also included. All the models were trained in one split, and Bootstrapping was used for calculating the 95\% Confidence Interval. We can see that, FedAvg and SCAFFOLD performance strongly degrades for $p > 0.8$. In terms of the Accuracy, BN-SCAFFOLD manages to maintain centralized learning performance for all degrees of heterogeneity. However, this is not verified for one-vs-one and one-vs-rest AUC. Local training performance is generally lower for each of the degrees of heterogeneity, and decreases more abruptly in the strongly heterogeneous setting.

Tables \ref{tab:2-patch_classifier_clinical_fl_acc}, \ref{tab:2-patch_classifier_clinical_fl_auc_one_vs_one}, and \ref{tab:2-patch_classifier_clinical_fl_auc_one_vs_rest} show the classification performance of the strong non-IID setting ($p=1$). In all cases, BN-SCAFFOLD significantly outperforms FedAvg, SCAFFOLD, and the models trained locally, but is outperformed by the centralized setting, with the exception of the first training setting, Client 1 w/ LUT (split 1), only when considering the Accuracy. We argue that this is probably an effect of the dataset split, although more experiments should be conducted to validate this hypothesis. The results of Figure \ref{fig:2-classif_perf_patch_classifier_ddsm_fl} correspond to the first column of Tables \ref{tab:2-patch_classifier_clinical_fl_acc}, \ref{tab:2-patch_classifier_clinical_fl_auc_one_vs_one}, and \ref{tab:2-patch_classifier_clinical_fl_auc_one_vs_rest}.

\begin{table}[H]
\footnotesize
\centering
\begin{tabular}{lllll}
\toprule
    & split 1 & split 2 & split 3 & split 4 \\
    \midrule
    Centralized &  0.601 $\pm$ 0.020 ($<$ 0.01) &  0.642 $\pm$ 0.019 ($<$ 0.01) &  0.650 $\pm$ 0.022 ($<$ 0.01) &  0.650 $\pm$ 0.022 ($<$ 0.01) \\
    \midrule
    Client 1 (local) &  0.446 $\pm$ 0.019 ($<$ 0.01) &  0.507 $\pm$ 0.021 ($<$ 0.01) &  0.450 $\pm$ 0.023 ($<$ 0.01) &  0.537 $\pm$ 0.024 ($<$ 0.01) \\
    Client 2 (local) &  0.508 $\pm$ 0.022 ($<$ 0.001) &  0.401 $\pm$ 0.022 ($<$ 0.01) &  0.509 $\pm$ 0.021 ($<$ 0.01) &  0.401 $\pm$ 0.022 ($<$ 0.01) \\
    \midrule
    FedAvg &  0.532 $\pm$ 0.021 ($<$ 0.01) &  0.501 $\pm$ 0.023 ($<$ 0.01) &  0.507 $\pm$ 0.020 ($<$ 0.01) &  0.505 $\pm$ 0.023 ($<$ 0.01) \\
    SCAFFOLD &  0.538 $\pm$ 0.021 ($<$ 0.01) &  0.502 $\pm$ 0.025 ($<$ 0.01) &  0.513 $\pm$ 0.021 ($<$ 0.01) &  0.494 $\pm$ 0.024 ($<$ 0.01) \\
    BN-SCAFFOLD & \textbf{0.614 $\pm$ 0.022 (n.a.)} &       \textbf{0.609 $\pm$ 0.023 (n.a.)} &       \textbf{0.615 $\pm$ 0.025 (n.a.)} &       \textbf{0.610 $\pm$ 0.022 (n.a.)} \\
\bottomrule
\end{tabular}
\caption{\small Accuracy, 95\% CI, and p-values in the strongly heterogeneous setting $p=1$, using original x-ray image.}
\label{tab:2-patch_classifier_clinical_fl_acc}
\end{table}

\begin{table}[H]
\footnotesize
\centering
\begin{tabular}{lllll}
\toprule
    & split 1 & split 2 & split 3 & split 4 \\
    \midrule
    centralized &  0.904 $\pm$ 0.005 ($<$ 0.01) &  0.922 $\pm$ 0.004 ($<$ 0.01) &  0.925 $\pm$ 0.004 ($<$ 0.01) &  0.919 $\pm$ 0.004 ($<$ 0.01) \\
    \midrule
    Client 1 (local) &  0.781 $\pm$ 0.006 ($<$ 0.01) &  0.858 $\pm$ 0.006 ($<$ 0.01) &  0.808 $\pm$ 0.007 ($<$ 0.01) &  0.873 $\pm$ 0.005 ($<$ 0.01) \\
    Client 2 (local) &  0.825 $\pm$ 0.007 ($<$ 0.01) &  0.771 $\pm$ 0.007 ($<$ 0.01) &  0.851 $\pm$ 0.006 ($<$ 0.01) &  0.770 $\pm$ 0.007 ($<$ 0.01) \\
    \midrule
    FedAvg &  0.844 $\pm$ 0.006 ($<$ 0.01) &  0.832 $\pm$ 0.007 ($<$ 0.01) &  0.843 $\pm$ 0.005 ($<$ 0.01) &  0.841 $\pm$ 0.006 ($<$ 0.01) \\
    SCAFFOLD &  0.851 $\pm$ 0.006 ($<$ 0.01) &  0.832 $\pm$ 0.007 ($<$ 0.01) &  0.849 $\pm$ 0.005 ($<$ 0.01) &  0.838 $\pm$ 0.006 ($<$ 0.01) \\   
    BN-SCAFFOLD &       \textbf{0.897 $\pm$ 0.005 (n.a.)} &       \textbf{0.887 $\pm$ 0.006 (n.a.)} &       \textbf{0.889 $\pm$ 0.005 (n.a.)} &       \textbf{0.884 $\pm$ 0.005 (n.a.)} \\
\bottomrule
\end{tabular}
\caption{\small Mean one-vs-one AUC, 95\% CI, and p-values in the strongly heterogeneous setting $p=1$, using original x-ray image.}
\label{tab:2-patch_classifier_clinical_fl_auc_one_vs_one}
\end{table}

\begin{table}[H]
\footnotesize
\centering
\begin{tabular}{lllll}
\toprule
        & split 1 & split 2 & split 3 & split 4 \\
        \midrule
        Centralized &  0.871 $\pm$ 0.009 ($<$ 0.01) &  0.891 $\pm$ 0.008 ($<$ 0.01) &  0.891 $\pm$ 0.009 ($<$ 0.01) &  0.883 $\pm$ 0.009 ($<$ 0.01) \\
        \midrule
        Client 1 (local) &  0.743 $\pm$ 0.011 ($<$ 0.01) &  0.801 $\pm$ 0.011 ($<$ 0.01) &  0.739 $\pm$ 0.011 ($<$ 0.01) &  0.827 $\pm$ 0.010 ($<$ 0.01) \\
        Client 2 (local) &  0.784 $\pm$ 0.011 ($<$ 0.01) &  0.647 $\pm$ 0.018 ($<$ 0.01) &  0.797 $\pm$ 0.011 ($<$ 0.01) &  0.634 $\pm$ 0.015 ($<$ 0.01) \\
        \midrule
        FedAvg &  0.807 $\pm$ 0.009 ($<$ 0.01) &  0.794 $\pm$ 0.011 ($<$ 0.01) &  0.800 $\pm$ 0.011 ($<$ 0.01) &  0.801 $\pm$ 0.011 ($<$ 0.01) \\
        SCAFFOLD &  0.812 $\pm$ 0.010 ($<$ 0.01) &  0.787 $\pm$ 0.012 ($<$ 0.01) &  0.806 $\pm$ 0.010 ($<$ 0.01) &  0.794 $\pm$ 0.011 ($<$ 0.01) \\
        BN-SCAFFOLD & \textbf{0.862 $\pm$ 0.009 (n.a.)} & \textbf{0.851 $\pm$ 0.009 (n.a.)} & \textbf{0.848 $\pm$ 0.010 (n.a.)} & \textbf{0.854 $\pm$ 0.009 (n.a.)} \\
\bottomrule
\end{tabular}
\caption{\small Mean one-vs-rest AUC, 95\% CI, and p-values in the strongly heterogeneous setting $p=1$, using original x-ray image.}
\label{tab:2-patch_classifier_clinical_fl_auc_one_vs_rest}
\end{table}

\subsection{Synthetic mammography patches}

\begin{figure}[H]
  \centering
  \subfloat[\centering \small Accuracy. \label{fig:2-acc_patch_classifier_synthetic_fl}]{\includegraphics[width=0.32\textwidth, keepaspectratio]{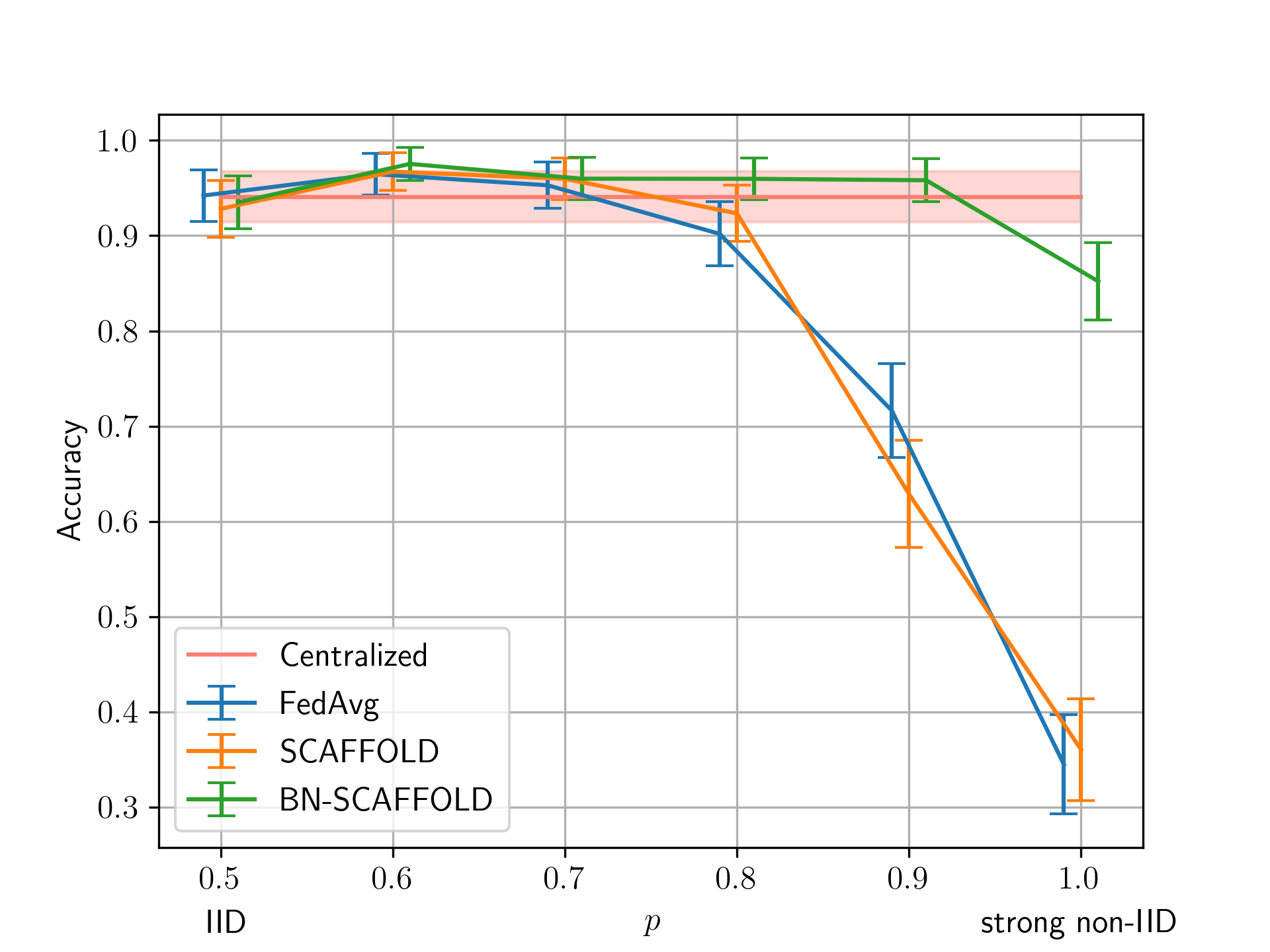}}
  \subfloat[\centering  \small Mean one-vs-one AUC. \label{fig:2-auc_one_vs_one_patch_classifier_synthetic_fl}]{\includegraphics[width=0.32\textwidth, keepaspectratio]{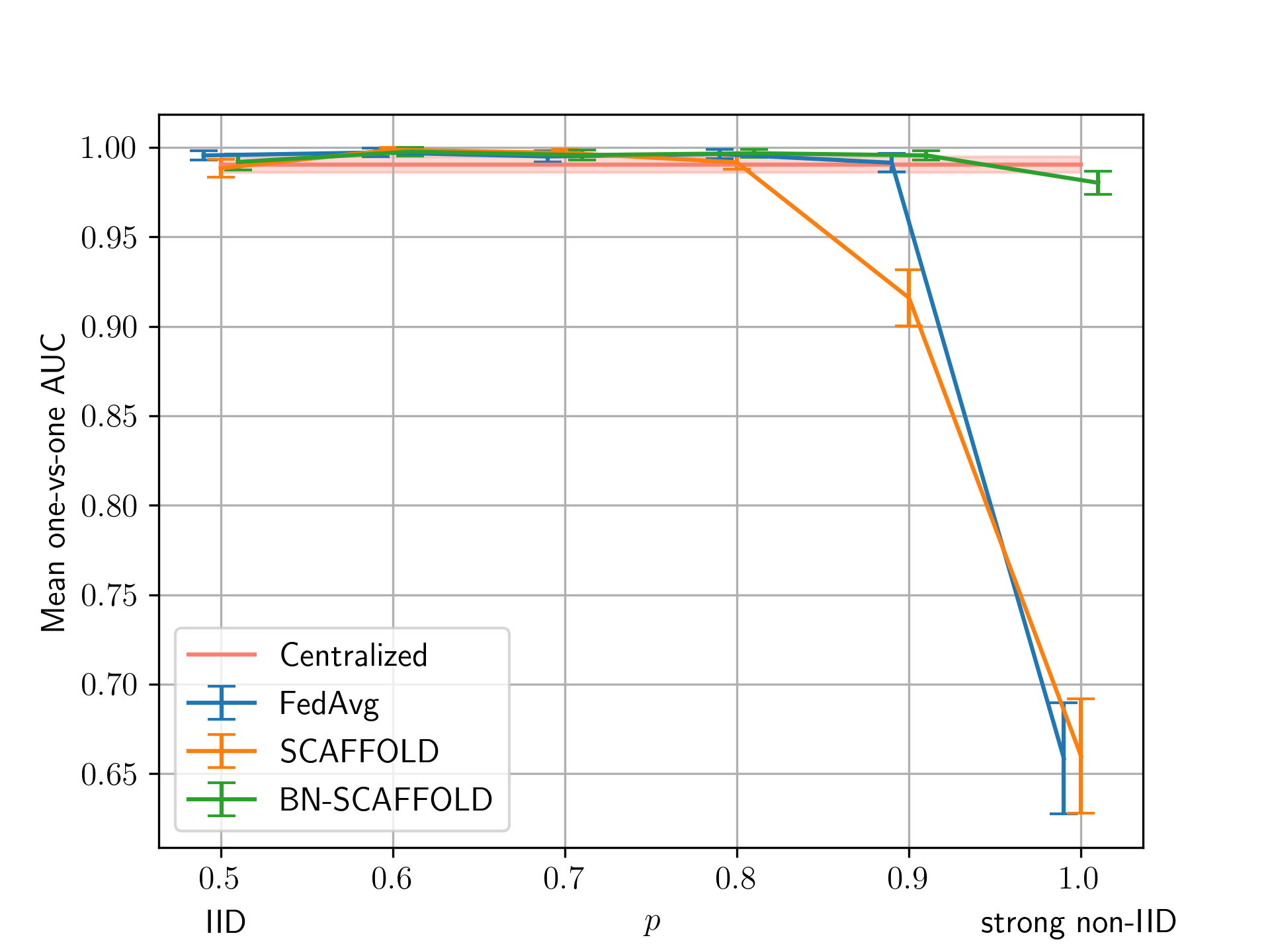}}
  \subfloat[\centering \small Mean one-vs-rest AUC. \label{fig:2-auc_one_vs_rest_patch_classifier_synthetic_fl}]{\includegraphics[width=0.32\textwidth, keepaspectratio]{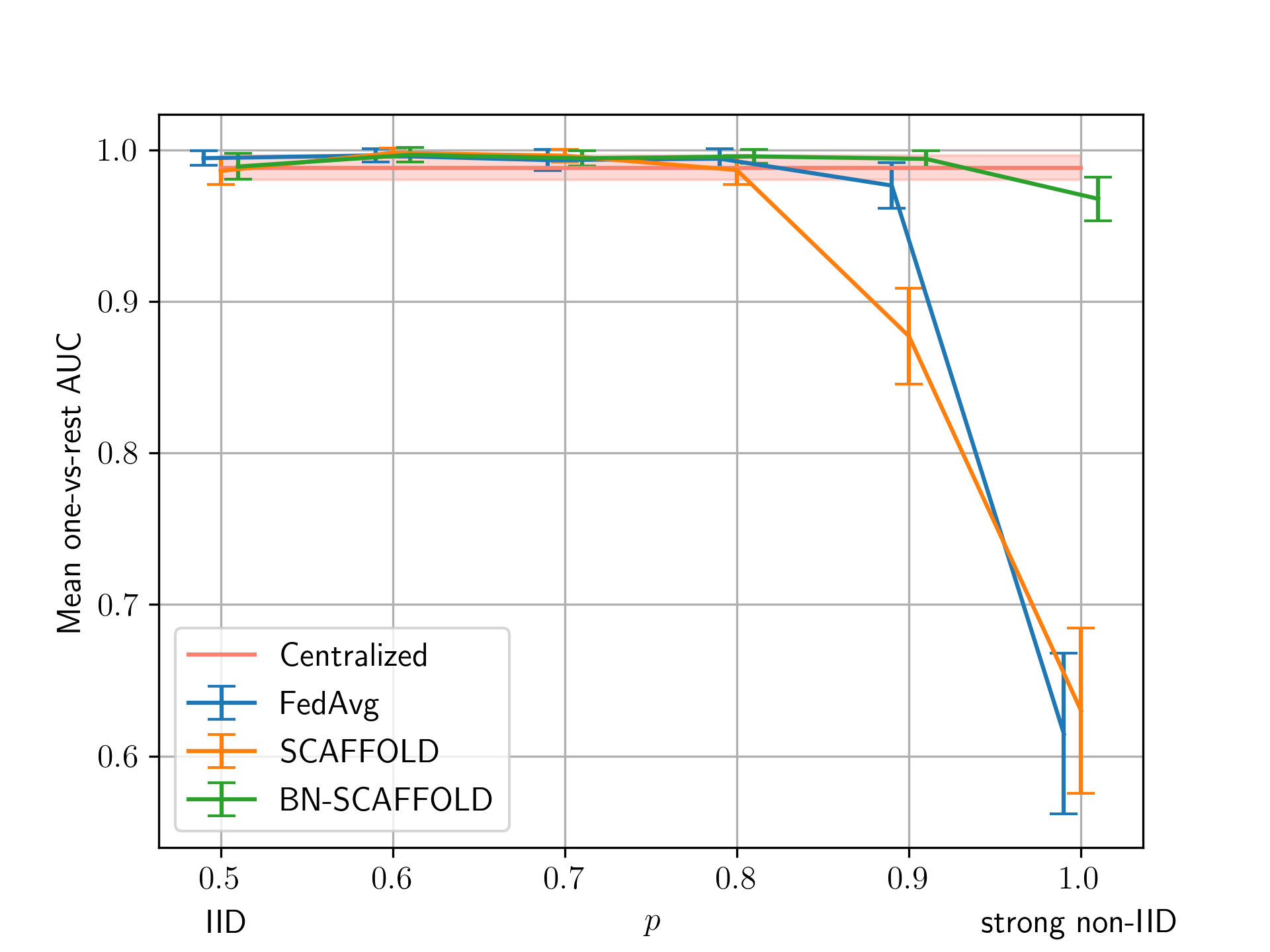}}
\caption{\small Classification performance with 95\% CI for different degrees of heterogeneity, using synthetic images.} 
\label{fig:2-classif_perf_patch_classifier_synthetic_fl}
\end{figure}

Figure \ref{fig:2-classif_perf_patch_classifier_synthetic_fl} compares the test set performance for different degrees of heterogeneity (given by $p$) of models trained with FedAvg, SCAFFOLD, and BN-SCAFFOLD for synthetic x-ray image classification, and compared to a centralized training. A single train-validation split was used for training, Confidence Intervals were obtained using Bootstrapping. We can see that performance using FedAvg and SCAFFOLD is relatively constant for moderate to low heterogeneity, but drops significantly when $p > 0.8$. In contrast, the performance of BN-SCAFFOLD is less affected by heterogeneity, and it manages to maintain performance in par with that of centralized training for $p \leq 0.9$. When $p=1$, it performance drops with respect to centralized training, but it is still significantly higher than that of the FL baselines.

\section{Conclusions and discussion}
\label{sec:conclusions}

In this work, the BN-SCAFFOLD algorithm is proposed to mitigate the impact of data heterogeneity by correcting the BN statistics client drift, in a cross-silo FL setting with stateful clients. A framework for analyzing the convergence of variance reduction algorithms in FL was introduced, which enabled to obtain convergence guarantees for BN-SCAFFOLD, and to compare its rate with other FL algorithms. We believe BN-SCAFFOLD represents the best trade-off among the considered algorithms, as it improves the convergence rate in a communication-efficient way. The theoretical results were validated with experiments on MNIST and CIFAR-10. BN-SCAFFOLD can also be applied to other normalization strategies that are impacted by batch heterogeneity, such as Batch Group Normalization \cite{BatchGroupNormalization}. The combination of BN-SCAFFOLD with Differential Privacy techniques should be investigated, as has been done for SCAFFOLD \cite{noble2022differentially} and for Batch Normalization \cite{davody2020effect, ponomareva2023dp}.

\paragraph{Limitations.} We consider that the main limitation of BN-SCAFFOLD is its reliance on stateful clients, which limits its applicability to cross-silo FL settings. However, we argue that the principle of BN statistics drift control can be extended to stateless algorithms, such as MIME \cite{MIME}. Secondly, it was considered that all clients were involved in all the global rounds, an assumption that is reasonable in cross-silo FL, but not in cross-device FL. Considering that only a fraction of the clients are involved in each global step adds a lag in the control variates \cite{SCAFFOLD}, which should translate into an additional source of variability in Theorem \ref{theorem:variance_reduc_algos}. Thirdly, checking that Assumption \ref{assumption:l_lipschitz_continuity_statistics} is verified can be complicated in practice, as it involves computing Lipschitz constants, which are known to be hard to estimate. Besides, the numerical results presented in this work were obtained for a single type of client heterogeneity on MNIST and CIFAR-10. In future works, the generalization of BN-SCAFFOLD to other datasets and types of data heterogeneity should be assessed. Finally, we focused on the convergence of SGD, but we consider that the theoretical study should be extended to other optimization algorithms, such as SGD with momentum, Adam \cite{kingma2014adam} or AdaGrad \cite{AdaGrad}. 

\section{Acknowledgements}

The authors wish to acknowledge the French ANRT (Association Nationale de la Recherche et de la Technologie) for partially funding this research, under CIFRE grant number 2021/0422. The authors also wish to acknowledge the creators and contributors of Nvidia Flare \cite{nvflare}.

\newpage

\medskip

\bibliography{neurips_bib}{}
\bibliographystyle{plain}

{
\small
}


\newpage
\appendix
\begin{center}
    \LARGE \textbf{Appendix} 
\end{center}
\input{appendix}


\end{document}

%% file: appendix.tex
The appendix is organized as follows. Appendix \ref{app:experimental_settings_details} details the experimental setting and hyperparameters of Section \ref{sec:experiments}. Appendix \ref{app:experiment_compute_resources} gives the compute resources used for each experiment. Appendix \ref{app:notation_summary} gives a summary of the notation used in the article. Appendix \ref{app:detailed_list_asumptions} gives a detailed list of the assumptions. Appendix \ref{app:fundamental_lemmas} details the fundamental Lemmas used in all the proofs, which seeks to give an intuition on them. In Appendix \ref{app:practical_formulas_scaffold_and_bn_scaffold}, the proof of the practical formulas for the calculation of the control variates in SCAFFOLD and BN-SCAFFOLD with option \RNum{2} are provided. Appendix \ref{app:proof_conv_theorem} contains the proof of Theorem \ref{theorem:variance_reduc_algos}. Appendices \ref{app:fed_avg_convergence_proof}, \ref{app:scaffold_convergence_proof}, \ref{app:bn_scaffold_convergence_proof}, \ref{app:fed_tan_convergence_proof} contain the application of Theorem \ref{theorem:variance_reduc_algos} to obtain the convergence bounds of FedAvg, SCAFFOLD, BN-SCAFFOLD, and FedTAN, respectively. Finally, Appendix \ref{app:useful_lemmas} contains several Lemmas used in the proof of Theorem \ref{theorem:variance_reduc_algos}.

\section{Experimental settings details}
\label{app:experimental_settings_details}

In this Section, we detail the hyperparameters used for the experiments of Section \ref{sec:experiments}. Two learning rate schedulers were used: $ StepScheduler$ and $ MultiStepScheduler$, which decrease a base learning rate $\gamma_0$ in some given step iterations $\tau$ by using a multiplicative factor $\alpha$. For instance, if $ MultiStepScheduler(\gamma_0=0.1, \alpha=0.5, \tau = \{ 100, 200 \})$ is used, the initial $\gamma=0.1$ is decreased to $\gamma=0.05$ after iteration 100, and to $\gamma=0.025$ after iteration 200. Additionally, for SCAFFOLD-based algorithms, a learning rate warmup scheduler was used to stabilize the control variates, as recommended in \cite{SCAFFOLD}. During the first $\tau_{warmup}$ iterations, the learning rate $\gamma $ is linearly increased from 0 to $\gamma_0$.

In BN-SCAFFOLD, the corrected variance was prevented from taking too small values, which can make the features diverge after normalization, by clipping it to a pre-determined minimum threshold. That is, for client $i$ at global step $r$ and local step $t$,

\begin{equation}
[(\tilde{\sigma}^2)^{r,t}_i]_j = \max \left\{ \sigma^2_{ths} ;  [(\tilde{\sigma}^2)^{r,t}_i]_j \right\}
\end{equation}

where $[(\tilde{\sigma}^2)^{r,t}_i]_j$ denotes the j-th component of the corrected variance vector, and $\sigma^2_{ths}$ the variance threshold.

All the models in each experiment were trained during the same number of iterations, which was set by making sure that all the algorithms had converged. In Figures \ref{fig:bn_scaffold_vs_baseline_heterogeneity} and Table \ref{tab:fl_algos_results}, the number of local steps was set to $E=10$ for MNIST and $E=100$ for CIFAR-10 in order to have moderate training time, and be able to training multiple models in 5-fold cross validation. Appendix \ref{app:experiment_compute_resources} details the training time of each experiment.

\subsection{MNIST}

The following hyperparameters are common to all the FL algorithms:

\begin{itemize}
    \item Model: ResNet-18 with projection shortcut (see \cite{ResNet}).
    \item Learning rate: fixed learning rate $\gamma = 0.5$ ($N=2$), and $ MultiStepScheduler$ with base learning rate $ \gamma_0 = 0.05 $ and parameters given in Table \ref{tab:mnist_lr_detail} for $N=5$. The base learning rate used for $N=5$ is smaller than for $N=2$, as an increased training instability was observed during training of the FL models with $N=5$.
    \item Batch-size: 128 at each client, and 128$N$ in the centralized setting, following standard practice on FL.
    \item Training iterations: $ 3.5 \times 10^3 $ ($N=2$) and $ 7 \times 10^3 $ ($N=2$).
    \item Local steps: $E=10$.
    \item Global steps: $R=350$ ($N=2$) and $R=700$ ($N=5$).
    \item Optimizer: SGD without momentum.
    \item Data Augmentation: no.
    \item Normalization: mean - std normalization.
    \item Loss function: Cross Entropy.
    \item Weight initialization: random with uniform distribution (default Pytorch).
    \item Control variate initialization: zero initialization with learning rate warm-up for stabilizing control variates.
    \item For FixBN, $T^* = 1.75 \times 10^3 $ ($N=2$), and $T^* = 2.5 \times 10^3 $ ($N=5$), or half the number of iterations.
    \item For BN-SCAFFOLD, the corrected variance threshold is set to $ \sigma^2_{ths} = 10^{-2} $.
\end{itemize}

\begin{table}[H]
\centering
\caption{Parameters of the $ MultiStepScheduler$ used for the learning rate $\gamma$ in each FL algorithm in MNIST experiments WITH $N=5$, where $\gamma_0$ is the initial learning rate, $\alpha$ the learning rate multiplicative factor, $\tau$ are the step iterations, and $\tau_{warmup}$ is the warm up iteration used for SCAFFOLD-based algorithms. n.a.: not applicable.}
\label{tab:mnist_lr_detail}
\begin{tabular}{c|c|c|c|c}
\toprule
 &  $\gamma_0$ & $\alpha$ & $\tau$ & $\tau_{warmup}$ \\
\midrule
 Centralized & 0.05 & 0.5 & $\{ 2 \times 10^3$, $3 \times 10^3 \}$ & n.a. \\
 FedAvg & 0.05 & 0.5 & $\{ 2 \times 10^3$, $3 \times 10^3 \}$ & n.a. \\
 FedBN & 0.05 & 0.5 & $\{ 2 \times 10^3$, $3 \times 10^3 \}$ & n.a. \\
 SiloBN & 0.05 & 0.5 & $\{ 2 \times 10^3$, $3 \times 10^3 \}$ & n.a. \\
 FixBN & 0.05 & 0.5 & $\{ 2 \times 10^3$, $3 \times 10^3 \}$ & n.a. \\
 FedTAN & 0.05 & 0.5 & $\{ 2 \times 10^3$, $3 \times 10^3 \}$ & n.a. \\
 SCAFFOLD & 0.05 & 0.5 & $\{ 2.5 \times 10^3$, $3.5 \times 10^3 \}$ & 500 \\
 FedBN+SCAFFOLD & 0.05 & 0.5 & $\{ 2.5 \times 10^3$, $3.5 \times 10^3 \}$ & 500 \\
 SiloBN+SCAFFOLD & 0.05 & 0.5 & $\{ 2.5 \times 10^3$, $3.5 \times 10^3 \}$ & 500 \\
 FixBN+SCAFFOLD & 0.05 & 0.5 & $\{ 2.5 \times 10^3$, $3.5 \times 10^3 \}$ & 500 \\
 BN-SCAFFOLD & 0.05 & 0.5 & $\{ 2.5 \times 10^3$, $3.5 \times 10^3 \}$ & 500 \\
\bottomrule
\end{tabular}
\end{table}

\subsection{CIFAR-10}

The following hyperparameters are common to all the FL algorithms:

\begin{itemize}
    \item Model: ResNet-18 with projection shortcut (see \cite{ResNet})
    \item Learning rate: $ StepScheduler$ with base learning rate $ \gamma_0 = 0.5 $ and parameters given in Table \ref{tab:cifar10_lr_detail}. For FixBN, FixBN+SCAFFOLD, and BN-SCAFFOLD, the step iteration $\tau$ was set to $2 \times 10^4$, instead of $ 3 \times 10^4 $ as for the other algorithms, to increase training stability.
    \item Batch-size: 128 at each client, and 128$N$ in the centralized setting..
    \item Optimizer: SGD without momentum.
    \item Training iterations: $ 4 \times 10^4 $.
    \item Local steps: $E=100$.
    \item Global steps: $R=400$.
    \item Data Augmentation: zero-padding with 2 pixels on each side, random horizontal flip with probability $p=0.5$, random crop to the initial size $32 \times 32$.
    \item Normalization: mean - std normalization with mean and std from ImageNet: mean = [125.3, 123.0, 113.9], std = [63.0, 62.1, 66.7].
    \item Loss function: Cross Entropy.
    \item Weight initialization: random with uniform distribution (default Pytorch).
    \item Control variate initialization: zero initialization with learning rate warm-up for stabilizing control variates.
    \item For FixBN, $T^* = 2 \times 10^4 $, or half the number of iterations.
    \item For BN-SCAFFOLD, the corrected variance threshold is set to $ \sigma^2_{ths} = 1 $.
\end{itemize}

\begin{table}[H]
\centering
\caption{Parameters of the $ StepScheduler$ used for the learning rate $\gamma$ in each FL algorithm in CIFAR-10 experiments, where $\gamma_0$ is the base or initial learning rate, $\alpha$ the learning rate multiplicative factor, $\tau$ is the step iteration, and $\tau_{warmup}$ is the warm up iteration used for SCAFFOLD-based algorithms. n.a.: not applicable.}
\label{tab:cifar10_lr_detail}
\begin{tabular}{c|c|c|c|c}
\toprule
 & $\gamma_0$ & $\alpha$ & $\tau$ & $\tau_{warmup}$ \\
 \midrule
 Centralized & 0.5 & 0.1 & $ 3 \times 10^4 $ & n.a. \\
 FedAvg & 0.5 & 0.1 & $ 3 \times 10^4 $ & n.a. \\
 FedBN & 0.5 & 0.1 & $ 3 \times 10^4 $ & n.a. \\
 SiloBN & 0.5 & 0.1 & $ 3 \times 10^4 $ & n.a. \\
 FixBN & 0.5 & 0.1 & $ 2 \times 10^4 $ & n.a. \\
 FedTAN & 0.5 & 0.1 & $ 3 \times 10^4 $ & n.a. \\
 SCAFFOLD & 0.5 & 0.1 & $ 3 \times 10^4 $ & n.a. \\
 FedBN+SCAFFOLD & 0.5 & 0.1 & $ 3 \times 10^4 $ & 400 \\
 SiloBN+SCAFFOLD & 0.5 & 0.1 & $ 3 \times 10^4 $ & 400 \\
 FixBN+SCAFFOLD & 0.5 & 0.1 & $ 2 \times 10^4 $ & 400 \\
 BN-SCAFFOLD & 0.5 & 0.1 & $ 2 \times 10^4 $ & 400 \\
\bottomrule
\end{tabular}
\end{table}

\section{Experiment compute resources and implementation}
\label{app:experiment_compute_resources}

The experiments were conducted in five workstations, each equipped with one 24 GB Nvidia Quadro RTX 6000 Graphical Processing Unit (GPU). All the FL algorithms, with the exception of FedTAN, were implemented and executed in the open-source FL framework Nvidia Flare with version 2.3.8 \cite{nvflare}, which is distributed under the Apache 2.0 license (\url{github.com/NVIDIA/NVFlare/blob/main/LICENSE}) . Each client was hosted in one separate workstation, and the server was hosted in one of these workstations. We remark that this constitutes a realistic FL environment. The training of all these algorithms took approximately the same amount of time. The centralized training model was directly implemented on PyTorch \cite{paszke2017automatic} with version 1.13.1 \url{github.com/pytorch/pytorch/blob/main/LICENSE}. For FedTAN, the implementation released by the authors in the original paper was used \cite{FedTAN}. In the FedTAN implementation, all the clients are hosted in the same workstation, which dramatically decreases communication cost with respect to the implementation of the other FL algorithms. In this Section, we provide the running time of all the trained algorithms, and the total amount of time required to compute each experiment if the models are trained sequentially. However, total computing time can be dramatically decreased if several models are trained in parallel, which can be achieved if GPUs with relatively large RAM are available, as it was the case in this work. The training times provided can also be decreased if the workstations are solely dedicated to running the algorithms, which was not the case in this work.

The training times of the models on MNIST with $N=2$ and for the different values of $E$ featured in Figure \ref{fig:bn_scaffold_vs_baseline_local_steps} are depicted in Table \ref{tab:mnist_n_2_training_times}. Thus, training all the models to obtain Figure \ref{fig:bn_scaffold_vs_baseline_heterogeneity} requires 1192.5 hours of computing time, and 2055 hours to obtain Figure \ref{fig:bn_scaffold_vs_baseline_local_steps}, if the models are executed sequentially. However, we remark that there are some models that are shared between Figures \ref{fig:bn_scaffold_vs_baseline_heterogeneity} and \ref{fig:bn_scaffold_vs_baseline_local_steps}, that do not need to be re-computed (centralized; FedAvg, SCAFFOLD, and BN-SCAFFOLD with $p=1$ in Figure \ref{fig:bn_scaffold_vs_baseline_heterogeneity} or $E=10$ in Figure \ref{fig:bn_scaffold_vs_baseline_local_steps}). Thus, generating both figures requires 3030 hours of sequential computing time. Generating the "MNIST (2 clients)" column in Table \ref{tab:fl_algos_results} requires 662.5 sequential computing hours, as there are 9 "other FL algorithms" (FedAvg, FedBN, SiloBN, FixBN, SCAFFOLD, FedBN+SCAFFOLD, SiloBN+SCAFFOLD, FixBN-SCAFFOLD, BN-SCAFFOLD). Again, this can be reduced to 445 hours if Figures \ref{fig:bn_scaffold_vs_baseline_heterogeneity} or \ref{fig:bn_scaffold_vs_baseline_local_steps} have already been generated.

\begin{table}[h]
\centering
\caption{Training times in MNIST with $N=2$ and 3500 iterations.}
\label{tab:mnist_n_2_training_times}
\begin{tabular}{llll}
\toprule
 & $E$ & running time (hours) & running time 5 runs (hours) \\
 \midrule
 Centralized & n.a. & 4.5 & 22.5 \\
 FedTAN & 10 & 11 & 55 \\
 Other FL algorithms & 1 & 68 & 340 \\
 Other FL algorithms & 2 & 34 & 170 \\
 Other FL algorithms & 5 & 16 & 80 \\
 Other FL algorithms & 10 & 13 & 65 \\
\bottomrule
\end{tabular}
\end{table}

The training times of MNIST with $N=5$ and 7000 iterations are depicted in Table \ref{tab:mnist_n_5_training_times}. Generating the "MNIST (5 clients)" column in Table \ref{tab:fl_algos_results} requires 790 hours of sequential GPU computing.

\begin{table}[h]
\centering
\caption{Training times in MNIST with $N=5$, local step $E=10$, and 7000 iterations.}
\label{tab:mnist_n_5_training_times}
\begin{tabular}{lll}
\toprule
  & running time (hours) & running time 5 runs (hours) \\
 \midrule
 Centralized. & 11 & 55 \\
 FedTAN  & 21 & 105 \\
 Other FL algorithms & 14 & 70 \\
\bottomrule
\end{tabular}
\end{table}

The training times on CIFAR-1O are depicted in Table \ref{tab:cifar10_training_times}. Obtaining the results of Table \ref{tab:fl_algos_results} requires a total of 765 sequential computing hours, as there are 9 "other FL algorithms" Thus, generating the entire Table \ref{tab:fl_algos_results} requires 2217.5 hours of sequential computing, and 2000 hours if Figure \ref{fig:bn_scaffold_vs_baseline_heterogeneity} or \ref{fig:bn_scaffold_vs_baseline_local_steps} has already been generated. Finally, the total amount of time to generate all the experiments of Section \ref{sec:experiments}, considering sequential training, is 5030 hours. However, additional trainings were necessary for hyperparameter selection and debugging.

\begin{table}[H]
\centering
\caption{Training times in CIFAR-10 with $N=2$, local step $E=100$, and $4 \times 10^4$ iterations.}
\label{tab:cifar10_training_times}
\begin{tabular}{lll}
\toprule
  & running time (hours) & running time 5 runs (hours) \\
 \midrule
 Centralized. & 7 & 35 \\
 FedTAN  & 20 & 100   \\
 Other FL algorithms & 14 & 70 \\
\bottomrule
\end{tabular}
\end{table}

\section{Notation summary}
\label{app:notation_summary}

Table \ref{tab:notations} summarizes the notations used in the main article, and in all the proofs in this appendix.

\begin{table}
    \begin{tabular}{ll}
    \textbf{Symbol} & \textbf{Description} \\
    \midrule
      $\gamma$   & learning rate \\
      $P_i$   & client $i$ \textit{a priori} probability \\
      $\mathcal{D}_i$ & client $i$ dataset \\
      $\mathcal{D}$   & global dataset \\
      $\mathcal{B}$   & mini-batch \\
      $R$      & number of global steps \\
      $F_i$    & loss function at client $i$ \\
      $F$     & global loss function  \\
      $\bar{w}_r$     & global model parameters at global step $r$ \\
      $\bar{w}^{r, t}_i$     & local model parameters at global step $r$, local step $t$, and client $i$ \\
      $ \nabla_w F_i (w^{r,t}_i ; S^{r,t}) $ & full local gradient at global step $r$, local step $t$, and client $i$ \\
      $ \nabla_w F (\bar{w}_r ; S^{r,0}) $ & full global gradient at global step $r$ \\
      $ g_i (w^{r,t}_i ; s^{r,t}) $ & local stochastic gradient at global step $r$, local step $t$, and client $i$ \\
      \underline{Problem constants: } \\
      $N$ & number of clients \\
      $E$ & number of local steps \\
      $V$ & bound on gradient dissimilarity when using global BN statistics \\
      $B$ & bound on gradient dissimilarity when using local and global BN statistics \\
      $L$ & Lipschitz constant gradient \\
      $J$ & Lipschitz constant statistics \\
      $M$ & difference between statistics bound \\
      \underline{Batch Normalization parameters: } \\
      $S^{r,t}$   & full global statistics at global step $r$ and local step $t$ \\
      $s^{r,t}_i$   & mini-batch global statistics at global step $r$ and local step $t$ \\
      $S^{r,t}_i$   & full statistics at global step $r$ and local step $t$, at client $i$ \\
      $s^{r,t}_i$   & mini-batch statistics at global step $r$ and local step $t$, at client $i$ \\
      $\tilde{S}^{r,t}_i$   & full corrected statistics at global step $r$ and local step $t$, at client $i$ \\
      $\tilde{s}^{r,t}_i$   & mini-batch corrected statistics at global step $r$ and local step $t$, at client $i$ \\
      $\hat{s}^{r,t}_i$   & running statistics at global step $r$ and local step $t$, at client $i$ \\
      $\bar{s}^{r,t}_i$   & global running statistics at global step $r$ and local step $t$ \\
      \underline{Control variates: } \\
      $c^{r}_i$ & local gradient control variate at global step $r$ and client $i$ \\
      $c^{r}$   & global gradient control variate at global step $r$ \\
      $k^{r}_i$ & local statistics control variate at global step $r$ and client $i$ \\
      $k^{r}$   & global statistics control variate at global step $r$ \\
      \bottomrule
     \end{tabular}
    \caption{Notation summary.}
    \label{tab:notations}
     
\end{table}

\section{Detailed list of assumptions}
\label{app:detailed_list_asumptions}

The assumptions use in the proofs are detailed below.

\begin{assumption}[Bounded stochastic gradient variance]
\label{assumption_app:bounded_stochastic_grad_var}

The mini-batch stochastic gradient $g_i(w, s_i)$ is unbiased, and its variance is bounded by $ \sigma^2 $, which decreases with the mini-batch size $ | \mathcal{B} | $:

\begin{equation}
    \begin{cases}
        \mathbb{E} \left[  g_i (w ; s_i) \right] = \nabla_w F_i ( w ; S_i ) \\
        \mathbb{E} \left[ \| g_i (w ; s_i) - \nabla_w F_i ( w ; S_i ) \|^2 \right] \leq \sigma^2 = \sigma^2_0 / | \mathcal{B} |.
    \end{cases}
\end{equation}

\end{assumption}

\begin{assumption}[Lipschitz continuity]
\label{assumption_app:l_lipschitz_continuity}

The local function $F_i$ are differentiable and their gradients are $L$-Lipschitz continuous, that is,

\begin{equation}
    \| \nabla_w F_i (w' ; S'_{D_i}) - \nabla_w F_i (w ; S_i) \| \leq L \| w' - w\| \quad \forall \quad i \in [1, N].
\end{equation}

\end{assumption}

\begin{assumption}[Lower-bounded global loss function]
\label{assumption_app:global_loss_function_lower_bound}

The global loss function is assumed non-convex and lower-bounded

\begin{equation}
    F(w ; S) \geq \ \underbar{F} \ > \ - \infty.
\end{equation}

\end{assumption}

\begin{assumption}[Bounded gradient dissimilarity]
\label{assumption_app:bounded_local_grad_dev}

There exist constants $B \geq 0$ and $ V \geq 0$ such that

\begin{equation}
    \begin{cases}
        \sum_{i=1}^N P_i \ \| \nabla_w F_i (w ; S_i) - \nabla_w F_i (w ; S) \|^2  \leq B^2 \\
        \sum_{i=1}^N P_i \ \| \nabla_w F_i (w ; S) - \nabla_w F (w ; S) \|^2 \leq V^2.
    \end{cases}
\end{equation}

\end{assumption}

\begin{assumption}[Bounded stochastic statistics]
\label{assumption_app:bounded_batch_statistics_variance}

The stochastic mini-batch statistics are unbiased, and their variance is bounded by $\sigma^2_s$

\begin{equation}
\begin{cases}
    \mathbb{E} \left[ s_i^{r, t} \right] = S^{r,t}_i \\ 
    \mathbb{E} \left[ \| s_i^{r, t} - S^{r,t}_i \|^2 \right] \leq \sigma^2_s = \sigma^2_{s,0} / | \mathcal{B}|.
\end{cases}
\end{equation}
    
\end{assumption}

\begin{assumption}[Lipschitz continuity for the statistics]
\label{assumption_app:l_lipschitz_continuity_statistics}

The gradients of the local functions $ F_i $ are Lipschitz continuous with respect to the statistics with constant $J$:

\begin{equation}
    \| \nabla_w F_i (w ; S'_i) - \nabla_w F_i (w ; S''_i) \| \leq J \| S'_i - S''_i \| \\
\end{equation}

In addition, the difference between two statistics is bounded by the difference between the model parameters from which they were obtained. That is,

\begin{equation}
    \| S'_i - S''_i \| \leq  M \| w' - w'' \| \quad \forall \quad i \in [1, N],
\end{equation}

where $S'_i$ and $S''_i$ are the BN statistics obtained from the local data at client $i$, with model parameters $w'$ and $w''$, respectively.

\end{assumption}

\section{Fundamental Lemmas used in the proofs}
\label{app:fundamental_lemmas}

The main fundamental Lemmas used in the proofs are presented below. Some of them are given without proof.

\begin{lemma}[Jensen's inequality]
\label{lemma:jensen_inequality}

Let $ \{ v_n \}_{n=1}^N $ be a sequence of vectors in $ \mathbb{R}^d $ and $ P_n \in [ 0, 1 ] $ such that $ \sum_{n=1}^N P_n = 1 $ and $ \phi : \mathbb{R}^d \rightarrow \mathbb{R} $ a convex function. Then,

\begin{equation*}
    \phi \left( \sum_{n=1}^N P_n v_n \right) \leq \sum_{n=1}^N P_n \phi \left( v_n \right).
\end{equation*}

\end{lemma}

\begin{corollary}
    If $ \phi ( . ) = \| . \|^2 $, we have

\begin{equation*}
    \| \sum_{n=1}^N P_n v_n \|^2 \leq \sum_{n=1}^N P_n \| v_n \|^2.
\end{equation*}

\begin{proof}
    The corollary is trivially proven by noting that $ \phi ( . ) = \| . \|^2 $ is a convex function, and using Lemma \ref{lemma:jensen_inequality}.
\end{proof}
    
\end{corollary}

\begin{lemma}[Relaxed triangle inequality]
\label{lemma:relaxed_triangle_inequality}

Let $ \{ v_n \}_{n=1}^N $ be a sequence of vectors in $ \mathbb{R}^d $. Then, the following are true:

\begin{itemize}
    \item[1.] $\| v_i + v_j \|^2 \leq (1+a) \| v_i \|^2 + (1 + \frac{1}{a}) \| v_j \|^2 $ for any $a > 0$, and
    \item[2.] $ \| \sum_{n=1}^N v_n \|^2 \leq N \sum_{n=1}^N \| v_n \|^2 $.
\end{itemize}
    
\end{lemma}

\begin{proof}
    For the proof of 1., see \cite{SCAFFOLD}. The proof of 2. follows from Jensen's inequality and the convexity of $\| . \|^2 $.
\end{proof}

\begin{lemma}
\label{lemma:identity_v1_cdot_v2}

Consider two vectors $v_1, v_2 \in \mathbb{R}^d $. We have that,

\begin{equation}
\langle v_1, v_2 \rangle = \frac{1}{2} ( \| v_1 \|^2 + \| v_2 \|^2 - \| v_1 - v_2 \|^2 ).
\end{equation}

\begin{proof}
    Trivially proven by developing $ \| v_1 - v_2 \|^2 $ and solving for $ \langle v_1, v_2 \rangle $.
\end{proof}

\begin{lemma}[Separating mean and variance]
\label{lemma:separating_mean_and_var}

Let $X \in \mathbb{R}^d$ be a random vector with $ \mathbb{E} [ X ] = \mu $ and $ \mathbb{E} \left[ \| X - \mu \|^2 \right] \leq \sigma^2 $, and $ a \in \mathbb{R}^d $. We have that,

\begin{equation*}
    \mathbb{E} \left[ \| X - a \|^2 \right] \leq \| \mu - a \|^2 + \sigma^2.
\end{equation*}

\begin{proof}

\begin{dmath}
    \mathbb{E} \left[ \| X - a  \|^2 \right] = \mathbb{E} \left[ \| X - \mu + \mu - a \|^2 \right] \\
    = \mathbb{E} \langle X - \mu + \mu - a, X - \mu + \mu - a \rangle \\
    = \underbrace{\mathbb{E} \left[ \| X - \mu \|^2 \right]}_{ \leq \sigma^2 } + \mathbb{E} \left[ \| \mu - a \|^2 \right] + 
\langle \underbrace{\mathbb{E} \left[ X - \mu \right]}_{=0}, \mu - a \rangle
\end{dmath}

\end{proof}

\end{lemma}

\end{lemma}

\begin{lemma}[$L$-Lipschitz continuity]
\label{lemma:l_lipschitz}

If the gradients of a function $f$ are $L$-Lipschitz continuous, then we have
\begin{equation}
    f (x) \leq f (y) + \langle \nabla f (y), x - y \rangle + \frac{L}{2} \| x - y \|^2.
\end{equation}
\end{lemma}

\begin{lemma}[Recursion]
\label{lemma:recursion}

Consider the following recursion equation

\begin{equation*}
    u_n \leq a u_{n-1} + b_{n-1},
\end{equation*}

with $ a < 1 $.

The following are true:

\begin{itemize}
    \item[1.] $ u_n \leq a^{n-1} u_1 + \sum_{l=1}^{n-1} a^{n-1-l} b_l $ for $ n \geq 2 $, and
    \item[2.] $ \sum_{n=1}^N u_n \leq \frac{1-a^N}{1-a} \left( u_1 + \sum_{n=1}^{N-1} b_n \right) \leq \frac{1}{1-a} \left( u_1 + \sum_{n=1}^{N-1} b_n \right) $ 
\end{itemize}
    
\end{lemma}

\begin{proof}
The proof of 1. follows from unrolling the recursion.

For 2. using 1. yields,

\begin{dmath*}
    \sum_{n=1}^N u_n \leq \frac{1-a^N}{1-a} u_1 + \sum_{n=2}^N \sum_{l=1}^{n-1} a^{n-1-l} b_l \\
    \leq \frac{1-a^N}{1-a} u_1 + b_1 (1+a+ \cdots +a^{N-2}) + b_2 (1 + a + \cdots + a^{N-3}) + \cdots + b_{N-1} (1) \\
    \leq \frac{1-a^N}{1-a} u_1 + \sum_{n=1}^{N-1} b_n \sum_{l=0}^{N-1-n} a^l \leq \frac{1-a^N}{1-a} u_1 + \sum_{n=1}^{N-1} \frac{1 - a^{N-n}}{1-a} b_n
    \leq \frac{1-a^N}{1-a} \left( u_1 + \sum_{n=1}^{N-1} b_n \right).
\end{dmath*}
    
\end{proof}

\section{SCAFFOLD and BN-SCAFFOLD option \RNum{2} practical formulas for control variates}
\label{app:practical_formulas_scaffold_and_bn_scaffold}

In this Section, we provide the proofs of the practical formulas used for calculating the gradient control variates $c^r_i$ in SCAFFOLD and BN-SCAFFOLD, and the statistics control variates $k^r_i$ in BN-SCAFFOLD. Despite the formula for the gradient control variates having been introduced in the SCAFFOLD original paper \cite{SCAFFOLD}, we proove it here for the sake of completeness.

\begin{lemma}[SCAFFOLD and BN-SCAFFOLD \RNum{2} gradient control variates update]
\label{lemma:scaffold_op_2_gradient_control_variates_proof}

If the local control variate at step $r$ and client $i$ is defined as 
    
\begin{equation}
    c^r_i \coloneq \frac{1}{E} \sum_{t=0}^{E-1} g_i (w^{r,t}_i ; \tilde{s}^{r,t}_i ),
\end{equation}

then it can be obtained by

\begin{equation}
    c^r_i = \frac{1}{\gamma E} \left( w^{r,0}_i - w^{r,E}_i \right) + c^{r-1}_i - c^{r-1}.
\end{equation}

\end{lemma}

\begin{proof}

The recursive model update is given by

\begin{equation}
    w^{r,t}_i = w^{r,t-1}_i - \gamma \left[ g_i (w^{r,t}_i ; \tilde{s}^{r,t}_i ) + c^{r-1} - c^{r-1}_i \right],
\end{equation}

and by unrolling it over $E$ steps

\begin{dmath}
    w^{r,E}_i = w^{r,0}_i - \gamma \sum_{t=0}^{E-1} \left[ g_i (w^{r,t}_i ; \tilde{s}^{r,t}_i ) + c^{r-1} - c^{r-1}_i \right] \\
    = w^{r,0}_i - \gamma \sum_{t=0}^{E-1} g_i (w^{r,t}_i ; \tilde{s}^{r,t}_i ) + \gamma E \left( c^{r-1} - c^{r-1}_i \right).
\end{dmath}

Solving for $ \sum_{t=0}^{E-1} g_i (w^{r,t}_i ; \tilde{s}^{r,t}_i ) $ and dividing by $E$ proves the relationship.
    
\end{proof}

\begin{lemma}[BN-SCAFFOLD option \RNum{2} statistics control variates update]
\label{lemma:bn_scaffold_op_2_stats_control_variates_proof}

    If the local control variate at step $r$ and client $i$ is defined as 

    \begin{equation}
        k^r_i \coloneq \frac{1 - \rho}{1 - \rho^E} \sum_{t=0}^{E-1} \rho^{E-1-t} s^{r,t}_i,
    \end{equation}

then it can be obtained from the running statistics by

\begin{equation}
    k^r_i = k^{r-1}_i - k^{r-1} + \frac{1}{1 - \rho^E} (\hat{s}^{r,E-1}_i - \rho^E \hat{s}^{r,0}_i ).
\end{equation}

\end{lemma}

\begin{proof}

The recursive update equation of the running statistics in BN-SCAFFOLD is given by,

\begin{equation}
    \hat{s}^{r, t}_i = \rho \hat{s}^{t-1}_i + (1 - \rho) \left( s^{t-1}_i + k^{r-1} - k^{r-1}_i \right).
\end{equation}

Unrolling the recursion over $E$ steps yields

\begin{dmath}
    \hat{s}^{r, E-1}_i = \rho^E \hat{s}^0_i + (1 - \rho) \left[ \sum_{t=0}^{E-1} \rho^{E-1-t} s^t_i + ( k^{r-1} - k^{r-1}_i ) \sum_{t=0}^{E-1} \rho^t \right] \\
    = \rho^E \hat{s}^0_i + (1 - \rho) \sum_{t=0}^{E-1} \rho^{E-1-t} s^t_i + (1 - \rho^E) ( k^{r-1} - k^{r-1}_i ).
\end{dmath}

Solving for $ (1 - \rho) \sum_{t=0}^{E-1} \rho^{E-1-t} s^t $ and diving by $ (1 - \rho^E) $ proves the relationship.
\end{proof}

\section{Proof of Theorem \ref{theorem:variance_reduc_algos}}
\label{app:proof_conv_theorem}

\begin{proof}
    
Starting from Assumption \ref{assumption_app:l_lipschitz_continuity}, we have:

\begin{dmath}
F ( \bar{w}_r ; S^{r+1,0} ) \leq F ( \bar{w}_{r-1} ; S^{r,0} ) + \langle \nabla_w F ( \bar{w}_{r-1} ; S^{r,0} ), \bar{w}_r - \bar{w}_{r-1} \rangle + \frac{L}{2} \| \bar{w}_r - \bar{w}_{r-1} \|^2.
\end{dmath}

By taking the expectation and using Lemmas \ref{lemma1} and \ref{lemma2},

\begin{dmath}
\label{eq:convergence_demonstration_inter_eq_1}
\mathbb{E} \left[ F ( \bar{w}_r ; S^{r+1,0} ) \right] \\
\leq F ( \bar{w}_{r-1} ; S^{r,0} ) 
- \frac{ \gamma E}{2} \mathbb{E} \left[ \| \nabla_w F ( \bar{w}_{r-1} ; S^{r,0} )  \|^2 \right] - \frac{ \gamma}{2} \sum_{t=1}^E \mathbb{E} \left[ \Big\| \sum_{j=1}^N P_j \ g_j( w_j^{r, t-1} ; \tilde{s}_j^{r, t-1})  \Big\|^2 \right] 
+ \gamma E \sum_{j=1}^N  P_j \mathbb{E} \left[ \| \nabla_w F_j ( \bar{w}_{r-1} ; S^{r,0} ) - \nabla_w F_j ( \bar{w}_{r-1} ; \tilde{S}^{r,0}_j ) \|^2 \right] \\ + \gamma L^2 \sum_{j=1}^N  P_j \sum_{t=1}^E \mathbb{E} \left[ \| \bar{w}_{r-1} - w_j^{r, t-1} \|^2 \right] + \frac{\gamma}{2} E \sigma^2
+ \frac{\gamma^2 L E}{2} \sum_{t=1}^E \mathbb{E} \left[ \Big\| \sum_{i=1}^{N} P_i \ g_i( w_i^{r, t-1} ; \tilde{s}_i^{r, t-1}) \Big\|^2 \right] \\
\leq F ( \bar{w}_{r-1} ; S^{r,0} ) - \frac{\gamma E}{2} \mathbb{E} \left[ \| \nabla_w F_i ( \bar{w}_{r-1} ; S^{r,0} ) \|^2 \right] - \frac{\gamma}{2} (1 - \gamma L E) \sum_{t=1}^E \mathbb{E} \left[ \Big\| \sum_{j=1}^N P_j \ g_j( w_j^{r, t-1} ; \tilde{s}_j^{r, t-1}) \Big\|^2 \right] \\ + \gamma E \sum_{j=1}^N P_j \mathbb{E} \left[ \| \nabla_w F ( \bar{w}_{r-1} ; S^{r,0} ) - \nabla_w F_j ( \bar{w}_{r-1} ; \tilde{S}^{r,0}_j ) \|^2 \right] + \gamma L^2 \sum_{j=1}^N P_j \sum_{t=1}^E \mathbb{E} \left[ \| \bar{w}_{r-1} - w_j^{r, t-1} \|^2 \right] + \frac{1}{2} \gamma E \sigma^2.
\end{dmath}

By using Lemma \ref{lemma3} and imposing $ \gamma < \frac{1}{\sqrt{8} L E} $:

\begin{dmath}
\mathbb{E} \left[ F ( \bar{w}_r ; S^{r+1,0} ) \right] \\
\leq F ( \bar{w}_{r-1} ; S^{r,0} ) - \frac{\gamma E}{2} \mathbb{E} \left[ \| \nabla_w F ( \bar{w}_{r-1} ; S^{r,0} ) \|^2 \right] - \frac{\gamma}{2} (1 - \gamma L E) \sum_{t=1}^E \mathbb{E} \left[ \Big\| \sum_{j=1}^N P_j \ g_j( w_j^{r, t-1} ; \tilde{s}_j^{r, t-1}) \Big\|^2 \right]
+ \gamma E \sum_{j=1}^N P_j \mathbb{E} \left[ \| \nabla_w F_j ( \bar{w}_{r-1} ; S^{r,0} ) - \nabla_w F_j ( \bar{w}_{r-1} ; \tilde{S}^{r,0}_j ) \|^2 \right]
\\ + \gamma L^2 \sum_{j=1}^N P_j \frac{E}{L^2} \delta_{E, \gamma, L} \ \sigma^2 + \gamma L^2 \sum_{j=1}^N P_j \frac{2E}{L^2} \delta_{E, \gamma, L} \mathbb{E} \left[ \Big\| \nabla_w F_j ( \bar{w}_{r-1}; \tilde{S}^{r, 0}_j ) - c_j^{r-1} \Big\|^2 \right] \\
+  \gamma L^2 \sum_{j=1}^N P_j \frac{2E}{L^2} \delta_{E, \gamma, L} \mathbb{E} \left[ \Big\| \sum_{i=1}^N P_i \left[ \nabla_w F_i ( \bar{w}_{r -1} ; \tilde{S}^{r,0}_{D_i} ) - c_i^{r-1} \right] \Big\|^2 \right] \\
+ \gamma L^2 \sum_{j=1}^N P_j \frac{2E}{L^2} \delta_{E, \gamma, L} \sum_{i=1}^N P_i \mathbb{E} \left[ \| \nabla_w F_j ( \bar{w}_{r-1} ; \tilde{S}^{r,0}_j ) - \nabla_w F_j ( \bar{w}_{r-1} ; S^{r,0} ) \|^2 \right] \\ + \gamma L^2 \sum_{j=1}^N P_j \frac{E}{L^2} \delta_{E, \gamma, L} \mathbb{E} \left[ \| \nabla_w F ( \bar{w}_{r-1} ; S^{r,0} ) \|^2 \right] + \frac{1}{2} \gamma E \sigma^2.
\end{dmath}

And by grouping terms and using Jensen's inequality, 

\begin{dmath}
\label{eq:convergence_demonstration_async_inter_eq_1}
\mathbb{E} \left[ F ( \bar{w}_r ; S^{r+1,0} ) \right] \\
\leq F ( \bar{w}_{r-1} ; S^{r,0} ) - \gamma E \left(\frac{1}{2} - \delta_{E, \gamma, L} \right) \mathbb{E} \left[ \| \nabla_w F ( \bar{w}_{r-1} ; S^{r,0} ) \|^2 \right] - \frac{\gamma}{2} (1 - \gamma L E) \sum_{t=1}^E \mathbb{E} \left[ \Big\| \sum_{j=1}^N P_j \ g_j( w_j^{r, t-1} ; \tilde{s}_j^{r, t-1}) \Big\|^2 \right] \\
+ \gamma E ( 1 + 2 \delta_{E, \gamma, L}) \sum_{j=1}^N P_j \mathbb{E} \left[ \| \nabla_w F_j ( \bar{w}_{r-1} ; S^{r,0} ) - \nabla_w F_j ( \bar{w}_{r-1} ; \tilde{S}^{r,0}_j ) \|^2 \right] \\
+ 4 \gamma E \ \delta_{E, \gamma, L} \sum_{j=1}^N P_j \mathbb{E} \left[ \| \nabla_w F_j ( \bar{w}_{r-1}; \tilde{S}^{r, 0}_j ) - c_j^{r-1} \|^2 \right]
+ ( \frac{1}{2} + \delta_{E, \gamma, L} ) \gamma E \sigma^2.
\end{dmath}

Using Lemma \ref{lemma:grad_vs_control_var}, we can bound the 5-th term of the right side of Equation \eqref{eq:convergence_demonstration_async_inter_eq_1}:

\begin{dmath}
 \label{eq:convergence_demonstration_async_inter_eq_2}
\| \nabla_w F_j ( \bar{w}_{r -1} ; \tilde{S}^{r, 0}_j ) - c^{r-1}_j \|^2 \\
  \leq \| \nabla_w F_j ( \bar{w}_{0} ; \tilde{S}^{1, 0}_j )- c^{0}_j \|^2 \one_{\{r=1\}} \\ + \left[  2 L^2 \gamma^2 E \sum_{t=1}^E \Big\| \sum_{i=1}^N P_i \ g_i ( w^{r-1, t-1}_i ; \tilde{s}^{r-1,t-1}_i ) \Big\|^2  + 2 \| \nabla_w F_j ( \bar{w}_{r -2} ; \tilde{S}^{r-1, 0}_j ) - c^{r-1}_j \|^2 \right] \one_{\{r \textgreater 1\}},
\end{dmath}

where $ \one_{ \{ \} } $ denotes the indicator function. By plugging Equation \eqref{eq:convergence_demonstration_async_inter_eq_2} into Equation \eqref{eq:convergence_demonstration_async_inter_eq_1} we get

\begin{dmath}
\label{eq:convergence_demonstration_async_inter_eq_3}
\mathbb{E} \left[ F ( \bar{w}_r ; S^{r+1,0} ) \right] \\
\leq F ( \bar{w}_{r-1} ; S^{r,0} ) - \gamma E \left(\frac{1}{2} - \delta_{E, \gamma, L} \right) \mathbb{E} \left[ \| \nabla_w F ( \bar{w}_{r-1} ; S^{r,0} ) \|^2 \right] \\ - \frac{\gamma}{2} (1 - \gamma L E) \sum_{t=1}^E \mathbb{E} \left[ \Big\| \sum_{j=1}^N P_j \ g_j( w_j^{r, t-1} ; \tilde{s}_j^{r, t-1}) \Big\|^2 \right] \\
+ \gamma E ( 1 + 2 \delta_{E, \gamma, L}) \sum_{j=1}^N P_j \mathbb{E} \left[ \| \nabla_w F_j ( \bar{w}_{r-1} ; S^{r,0} ) - \nabla_w F_j ( \bar{w}_{r-1} ; \tilde{S}^{r,0}_j ) \|^2 \right] \\
+ 4 \gamma E \ \delta_{E, \gamma, L} \sum_{j=1}^N P_j \| \nabla_w F_j ( \bar{w}_{0} ; \tilde{S}^{1, 0}_j )- c^{0}_j \|^2 \one_{ \{ r =  1 \} } \\
+ 8 \gamma E \ \delta_{E, \gamma, L} \sum_{j=1}^N P_j \mathbb{E} \left[ \| \nabla_w F_j ( \bar{w}_{r -2} ; \tilde{S}^{r-1, 0}_j ) - c^{r-1}_j \|^2 \right] \one_{ \{ r >  1 \} } \\
+ 8 L^2 \gamma^3 E^2 \ \delta_{E, \gamma, L} \sum_{j=1}^N P_j \sum_{t=1}^E \mathbb{E} \left[ \Big\| \sum_{i=1}^N P_i \ g_i ( w^{r-1, t-1}_i ; \tilde{S}^{r-1,t-1}_i ) \Big\|^2 \right] \one_{ \{ r >  1 \} } + ( \frac{1}{2} + \delta_{E, \gamma, L} ) \gamma E \sigma^2.
\end{dmath}

By summing from $r=1$ to $R$ and dividing by $\gamma R E$, where $RE$ is the total number of steps:

\begin{dmath}
\frac{1}{\gamma R E} \sum_{r=1}^R \mathbb{E} \left[ F ( \bar{w}_r ; S^{r+1,0} ) \right] \\
\leq \frac{1}{\gamma R E} \sum_{r=1}^R \mathbb{E}  \left[ F ( \bar{w}_{r-1} ; S^{r,0} ) \right] - 
\frac{1}{R} \left(\frac{1}{2} - \delta_{E, \gamma, L} \right) \sum_{r=1}^R \mathbb{E} \left[ \| \nabla_w F ( \bar{w}_{r-1} ; S^{r,0} ) \|^2 \right] \\ 
- \frac{1}{2 R E} (1 - \gamma L E) \sum_{r=1}^R \sum_{t=1}^E \mathbb{E} \left[ \Big\| \sum_{j=1}^N P_j \ g_j( w_j^{r, t-1} ; \tilde{s}_j^{r, t-1}) \Big\|^2 \right] \\
+ \frac{1}{R} ( 1 + 2 \delta_{E, \gamma, L}) \sum_{r=1}^R \sum_{j=1}^N P_j \mathbb{E} \left[ \| \nabla_w F_j ( \bar{w}_{r-1} ; S^{r,0} ) - \nabla_w F_j ( \bar{w}_{r-1} ; \tilde{S}^{r,0}_j ) \|^2 \right] \\
+ 4 \delta_{E, \gamma, L} \frac{1}{R} \sum_{j=1}^N P_j \| \nabla_w F_j ( \bar{w}_{0} ; \tilde{S}^{1, 0}_j )- c^{0}_j \|^2 \\
+ 8 \delta_{E, \gamma, L} \frac{1}{R} \sum_{r=2}^R \sum_{j=1}^N P_j \mathbb{E} \left[ \| \nabla_w F_j ( \bar{w}_{r -2} ; \tilde{S}^{r-1, 0}_j ) - c^{r-1}_j \|^2 \right] \\
+ 8 L^2 \gamma^2 E \ \delta_{E, \gamma, L} \frac{1}{R} \sum_{r=2}^R \sum_{j=1}^N P_j \sum_{t=1}^E \mathbb{E} \left[ \Big\| \sum_{i=1}^N P_i \ g_i ( w^{r-1, t-1}_i ; \tilde{S}^{r-1,t-1}_i ) \Big\|^2 \right]
+ ( \frac{1}{2} + \delta_{E, \gamma, L} ) \sigma^2.
\end{dmath}

By re-arranging indices, we have

\begin{dmath}
\label{eq:convergence_demonstration_async_inter_eq_4}
\frac{1}{\gamma R E} \sum_{r=1}^R \mathbb{E} \left[ F ( \bar{w}_r ; S^{r+1,0} ) \right] \\
\leq \frac{1}{\gamma R E} \sum_{r=1}^R \mathbb{E}  \left[ F ( \bar{w}_{r-1} ; S^{r,0} ) \right] - 
\frac{1}{R} \left(\frac{1}{2} - \delta_{E, \gamma, L} \right) \sum_{r=1}^R \mathbb{E} \left[ \| \nabla_w F ( \bar{w}_{r-1} ; S^{r,0} ) \|^2 \right] \\ 
- \frac{1}{2 R E} (1 - \gamma L E - 16 L^2 \gamma^2 E^2 \delta_{E, \gamma, L} ) \sum_{r=1}^R \sum_{t=1}^E \mathbb{E} \left[ \Big\| \sum_{j=1}^N P_j \ g_j( w_j^{r, t-1} ; \tilde{s}_j^{r, t-1}) \Big\|^2 \right] \\
+ \frac{1}{R} ( 1 + 2 \delta_{E, \gamma, L}) \sum_{r=1}^R \sum_{j=1}^N P_j \mathbb{E} \left[ \| \nabla_w F_j ( \bar{w}_{r-1} ; S^{r,0} ) - \nabla_w F_j ( \bar{w}_{r-1} ; \tilde{S}^{r,0}_j ) \|^2 \right] \\
+ 4 \delta_{E, \gamma, L} \frac{1}{R} \sum_{j=1}^N P_j \| \nabla_w F_j ( \bar{w}_{0} ; \tilde{S}^{1, 0}_j )- c^{0}_j \|^2 \\
+ 8 \delta_{E, \gamma, L} \frac{1}{R} \sum_{r=1}^R \sum_{j=1}^N P_j \mathbb{E} \left[ \| \nabla_w F_j ( \bar{w}_{r -1} ; \tilde{S}^{r, 0}_j ) - c^r_j \|^2 \right]
+ ( \frac{1}{2} + \delta_{E, \gamma, L} ) \sigma^2.
\end{dmath}

Re-organizing:

\begin{dmath}
\label{eq:convergence_demonstration_inter_eq_5}
\frac{1}{R} \left( \frac{1}{2} - \delta_{E, \gamma, L} \right) \sum_{r=1}^R \mathbb{E} \left[ \| \nabla_w F ( \bar{w}_{r-1} ; S^{r,0} ) \|^2 \right]  \\
\leq \frac{1}{\gamma R E} \mathbb{E} \left[ \sum_{r=1}^R \left[ F ( \bar{w}_{r-1} ; S^{r,0} ) - F ( \bar{w}_r ; S^{r+1,0} ) \right] \right]
+ \frac{1}{R} ( 1 + 2 \delta_{E, \gamma, L}) \sum_{r=1}^R \sum_{j=1}^N P_j \mathbb{E} \left[ \| \nabla_w F_j ( \bar{w}_{r-1} ; S^{r,0} ) - \nabla_w F_j ( \bar{w}_{r-1} ; \tilde{S}^{r,0}_j ) \|^2 \right] \\
+ 4 \delta_{E, \gamma, L} \frac{1}{R} \sum_{j=1}^N P_j \| \nabla_w F_j ( \bar{w}_{0} ; \tilde{S}^{1, 0}_j )- c^{0}_j \|^2
+ 8 \delta_{E, \gamma, L} \frac{1}{R} \sum_{r=1}^R \sum_{j=1}^N P_j \mathbb{E} \left[ \| \nabla_w F_j ( \bar{w}_{r -1} ; \tilde{S}^{r, 0}_j ) - c^r_j \|^2 \right]
+ ( \frac{1}{2} + \delta_{E, \gamma, L} ) \sigma^2 \\
- \frac{1}{2 R E} (1 - \gamma L E - 16 L^2 E^2 \gamma^2 \delta_{E, \gamma, L} ) \sum_{r=1}^R \sum_{t=1}^E \mathbb{E} \left[ \Big\| \sum_{j=1}^N P_j \ g_j( w_j^{r, t-1} ; \tilde{s}_j^{r, t-1}) \Big\|^2 \right].
\end{dmath}

But, using that $ \sum_{r=1}^R \left( \alpha_{r-1} - \alpha_r \right) = \alpha_0 - \alpha_R $, we have

\begin{dmath}
\label{eq:convergence_demonstration_inter_eq_6}
    \mathbb{E} \left[ \sum_{r=1}^R \left[ F ( \bar{w}_{r-1} ; S^{r,0} ) - F( \bar{w}_r ; S^{r+1,0} ) \right] \right]
    = \mathbb{E} \left[ F( \bar{w}_0 ; S^{1,0}) - F( \bar{w}_r ; S^{r+1,0} ) \right]
    \leq F( \bar{w}_0 ; S^{1,0}) - \underbar{F},
\end{dmath}

as the global function F is assumed lower bounded (Assumption \ref{assumption_app:global_loss_function_lower_bound}).

By plugging Equation \eqref{eq:convergence_demonstration_inter_eq_6} into Equation \eqref{eq:convergence_demonstration_inter_eq_5}, we have

\begin{dmath}
\label{eq:convergence_demonstration_inter_eq_7}
\frac{1}{R} \left( \frac{1}{2} - \delta_{E, \gamma, L} \right) \sum_{r=1}^R \mathbb{E} \left[ \| \nabla_w F ( \bar{w}_{r-1} ; S^{r,0} ) \|^2 \right]  \\
\leq \frac{1}{\gamma R E} \left[ F( \bar{w}_0 ; S^{1,0}) - \underbar{F} \right]
+ \frac{1}{R} ( 1 + 2 \delta_{E, \gamma, L}) \sum_{r=1}^R \sum_{j=1}^N P_j \mathbb{E} \left[ \| \nabla_w F_j ( \bar{w}_{r-1} ; S^{r,0} ) - \nabla_w F_j ( \bar{w}_{r-1} ; \tilde{S}^{r,0}_j ) \|^2 \right] \\
+ 4 \delta_{E, \gamma, L} \frac{1}{R} \sum_{j=1}^N P_j \| \nabla_w F_j ( \bar{w}_{0} ; \tilde{S}^{1, 0}_j )- c^{0}_j \|^2 \\
+ 8 \delta_{E, \gamma, L} \frac{1}{R} \sum_{r=1}^R \sum_{j=1}^N P_j \mathbb{E} \left[ \| \nabla_w F_j ( \bar{w}_{r -1} ; \tilde{S}^{r, 0}_j ) - c^r_j \|^2 \right]
+ ( \frac{1}{2} + \delta_{E, \gamma, L} ) \sigma^2 \\
- \frac{1}{2 R E} (1 - \gamma L E - 16 L^2 E^2 \gamma^2 \delta_{E, \gamma, L} ) \sum_{r=1}^R \sum_{t=1}^E \mathbb{E} \left[ \Big\| \sum_{j=1}^N P_j \ g_j( w_j^{r, t-1} ; \tilde{s}_j^{r, t-1}) \Big\|^2 \right],
\end{dmath}

which proves the Theorem. For Equation \eqref{eq:convergence_demonstration_inter_eq_7} to be valid, we need to have

\begin{equation}
\frac{1}{2} - \delta_{E, \gamma, L} > 0 \leftrightarrow \gamma < \frac{1}{\sqrt{12} L E}.   
\end{equation}

The final condition for the learning rate is then

\begin{equation}
    \gamma < \min \left\{ \frac{1}{\sqrt{12}}; \frac{1}{\sqrt{8}} \right\} \frac{1}{L E} < \frac{1}{\sqrt{12} L E}.
\end{equation}

\end{proof}

\section{FedAvg convergence}
\label{app:fed_avg_convergence_proof}

\begin{lemma}
\label{lemma:fed_avg_convergence}

The convergence rate of Vanilla FedAvg is given by

\begin{dmath}
    \frac{1}{R} \sum_{r=1}^R \mathbb{E} \left[ \| \nabla_w F ( \bar{w}_{r-1} ; S^{r,0} ) \|^2 \right]  \\
    \leq \frac{1}{1 - 50 \delta_{E, \gamma, L}} \left[ \frac{2}{\gamma R E} \left[ F( \bar{w}_0 ; S^{1,0}) - \underbar{F} \right]
    + 2 ( 1 + 26 \delta_{E, \gamma, L} + \frac{8}{R} \delta_{E, \gamma, L}) B^2 \\
    + 2 (24 + \frac{8}{R} ) \delta_{E, \gamma, L} V^2
    + 24 \delta_{E, \gamma, L} \frac{1}{R} \nabla F_0^2
    + ( 1 + 2 \delta_{E, \gamma, L} ) \sigma^2 \right],
\end{dmath}

with $ \gamma < \frac{1}{\sqrt{208} L E} $.
    
\end{lemma}

\begin{proof}
By setting $ c_j^r = 0 $ and $\tilde{S}_j^{r, 0} = S_j^{r, 0} $, $ \mathcal{T}_2 $ yields

\begin{dmath}
\mathcal{T}_2 = \frac{8}{R} \delta_{E, \gamma, L} \mathbb{E} \left[ \sum_{j=1}^N P_j \| \nabla_w F_j ( \bar{w}_{0} ; S^{1, 0}_j ) \|^2 \right] \\
\leq \frac{24}{R} \delta_{E, \gamma, L} \mathbb{E} \left[ \sum_{j=1}^N P_j \| \nabla_w F_j ( \bar{w}_{0} ; S^{1, 0}_j) - \nabla_w F_j ( \bar{w}_{0} ; S^{1, 0} ) \|^2 + \sum_{j=1}^N P_j \| \nabla_w F_j ( \bar{w}_{0} ; S^{1, 0} ) - \nabla_w F ( \bar{w}_{0} ; S^{1, 0} ) \|^2  \right] \\ + \frac{24}{R} \delta_{E, \gamma, L} \| \nabla_w F ( \bar{w}_{0} ; S^{1, 0} ) \|^2 \\
\leq \frac{24}{R} \delta_{E, \gamma, L} \left[ B^2 + V^2 + \| \nabla_w F ( \bar{w}_{0} ; S^{1, 0} ) \|^2 \right],
\end{dmath}

and $ \mathcal{T}_3 $ yields,

\begin{dmath}
\mathcal{T}_3
= \frac{2}{R} (1 + 2 \delta_{E, \gamma, L}) \sum_{r=1}^R \mathbb{E} \left[ \sum_{j=1}^N P_j \| \nabla_w F_j ( \bar{w}_{r-1} ; S^{r,0} ) - \nabla_w F_j ( \bar{w}_{r-1} ; S^{r,0}_j ) \|^2 \right] \leq 2 (1 + 2 \delta_{E, \gamma, L})  B^2,
\end{dmath}

and $ \mathcal{T}_4 $ yields

\begin{dmath}
\label{eq:fed_avg_conv_eq_int_1}
    \mathcal{T}_4
    = \frac{16}{R} \delta_{E, \gamma, L} \sum_{r=1}^R \sum_{j=1}^N P_j \mathbb{E} \left[ \| \nabla_w F_j (\bar{w}_{r-1}; S^{r, 0}_j) \|^2 \right] \\
    = \frac{16}{R} \delta_{E, \gamma, L} \sum_{r=1}^R \sum_{j=1}^N P_j \| \nabla_w F_j (\bar{w}_{r-1}; S^{r, 0}_j) - \nabla_w F_j (\bar{w}_{r-1}; S^{r,0}) \\ + \nabla_w F_j (\bar{w}_{r-1}; S^{r,0}) - \nabla_w F (\bar{w}_{r-1}; S^{r,0}) + \nabla_w F (\bar{w}_{r-1}; S^{r,0})  \|^2 \\
    \leq \frac{48}{R} \delta_{E, \gamma, L} \left[ \sum_{r=1}^R \sum_{j=1}^N P_j  \| \nabla_w F_j (\bar{w}_{r-1}; S^{r, 0}_j) - \nabla_w F_j (\bar{w}_{r-1}; S^{r,0}) \|^2 + \\
    \sum_{r=1}^R \sum_{j=1}^N P_j \| \nabla_w F_j (\bar{w}_{r-1}; S^{r,0}) - \nabla_w F (\bar{w}_{r-1}; S^{r,0}) \|^2  \\ 
    + \sum_{r=1}^R \sum_{j=1}^N P_j \| \nabla_w F (\bar{w}_{r-1}; S^{r,0})  \|^2 \right] \\
    \leq 48 \delta_{E, \gamma, L} \left[ B^2 + V^2 + \frac{1}{R} \sum_{r=1}^R \| \nabla_w F (\bar{w}_{r-1}; S^{r,0})  \|^2 \right],
\end{dmath}

where Assumption \ref{assumption_app:bounded_local_grad_dev} was used. In addition, by setting $ \gamma $ small enough such that $ 1 - \gamma L E - 16 L^2 E^2 \gamma^2 \delta_{E, \gamma, L} \geq 0 $, then the last term is negative and can be removed from the bound. We then get,

\begin{dmath}
\frac{1}{R} \left( 1 - 2 \delta_{E, \gamma, L} \right) \sum_{r=1}^R \mathbb{E} \left[ \| \nabla_w F ( \bar{w}_{r-1} ; S^{r,0} ) \|^2 \right]  \\
\leq \frac{2}{\gamma R E} \left[ F( \bar{w}_0 ; S^{1,0}) - \underbar{F} \right]
+ 2 ( 1 + 2 \delta_{E, \gamma, L}) B^2 \\
+ 24 \delta_{E, \gamma, L} \frac{1}{R} B^2 + 24 \delta_{E, \gamma, L} \frac{1}{R} V^2 + 24 \delta_{E, \gamma, L} \frac{1}{R}  \| \nabla_w F ( \bar{w}_{0} ; S^{1, 0} ) \|^2 \\
+ 48 \delta_{E, \gamma, L} B^2 + 48 \delta_{E, \gamma, L} V^2 + 48 \delta_{E, \gamma, L} \frac{1}{R} \sum_{r=1}^R \| \nabla_w F (\bar{w}_{r-1}; S^{r,0})  \|^2
+ ( 1 + 2 \delta_{E, \gamma, L} ) \sigma^2.
\end{dmath}

And thus,

\begin{dmath}
\frac{1}{R} \left( 1 - 50 \delta_{E, \gamma, L} \right) \sum_{r=1}^R \mathbb{E} \left[ \| \nabla_w F ( \bar{w}_{r-1} ; S^{r,0} ) \|^2 \right]  \\
\leq \frac{2}{\gamma R E} \left[ F( \bar{w}_0 ; S^{1,0}) - \underbar{F} \right]
+ 2 ( 1 + 26 \delta_{E, \gamma, L} + \frac{12}{R} \delta_{E, \gamma, L}) B^2 \\
+ 2 (24 + \frac{12}{R} ) \delta_{E, \gamma, L} V^2 
+ 24 \delta_{E, \gamma, L} \frac{1}{R}  \| \nabla_w F ( \bar{w}_{0} ; S^{1, 0} ) \|^2 
+ ( 1 + 2 \delta_{E, \gamma, L} ) \sigma^2 \\
\leq \frac{2}{\gamma R E} \left[ F_0 - \underbar{F} \right]
+ 2 ( 1 + 26 \delta_{E, \gamma, L} + \frac{12}{R} \delta_{E, \gamma, L}) B^2 \\
+ 2 (24 + \frac{12}{R} ) \delta_{E, \gamma, L} V^2 
+ 24 \delta_{E, \gamma, L} \frac{1}{R} \nabla F_0^2
+ ( 1 + 2 \delta_{E, \gamma, L} ) \sigma^2,
\end{dmath}

where $ F( \bar{w}_0 ; S^{1,0}) = F_0 $ and $ \| \nabla_w F ( \bar{w}_{0} ; S^{1, 0} ) \|^2 \coloneq \nabla F_0^2 $.

The following conditions on the learning rate need to be verified:

\begin{equation}
    \begin{cases}
        1 - 50 \delta_{E, \gamma, L} > 0 \\
        1 - \gamma L E - 16 L^2 E^2 \gamma^2 \delta_{E, \gamma, L} \geq 0.
    \end{cases}
\end{equation} 

The first condition is verified if $\gamma < \frac{1}{\sqrt{208} LE} $.

For the second condition, to avoid solving for the roots of a 4-th order polynomial, we split the second condition in two sub-conditions:

\begin{equation}
    1 - \gamma L E - 16 L^2 E^2 \gamma^2 \delta_{E, \gamma, L} = 1/2 - \gamma L E + 1/2 - 16 L^2 E^2 \gamma^2 \delta_{E, \gamma, L} \geq 0,
\end{equation}

which is ensured to be met if $ 1/2 - \gamma L E  \geq 0 $ and $ 1/2 - 16 L^2 E^2 \gamma^2 \delta_{E, \gamma, L} \geq 0 $. This imposes two new conditions: $ \gamma \leq \frac{1}{2LE} $ and $ \gamma \leq \frac{1}{4 L E} $. We thus have,

\begin{equation}
    \gamma < \frac{1}{\sqrt{208} L E}.
\end{equation}

\end{proof}

\begin{corollary}
\label{corollary:fed_avg_final_conv}

By taking $ \gamma = \gamma_0  \frac{1}{L E \sqrt{R}} $, we obtain

\begin{dmath}
    \min_r \mathbb{E} \left[ \| \nabla_w F ( \bar{w}_{r-1} ; S^{r,0} ) \|^2 \right] \\
    \leq 
    \frac{1}{R} \sum_{r=1}^R \mathbb{E} \left[ \| \nabla_w F ( \bar{w}_{r-1} ; S^{r,0} ) \|^2 \right] \\
    \leq \mathcal{O} \left( \frac{L}{\sqrt{R}} \left[ F_0 - \underbar{F} \right] + (1 + \frac{1}{R} + \frac{1}{R^2} ) B^2 + (1 + \frac{1}{R} ) \frac{1}{R} V^2 + \frac{1}{R^2} \nabla F_0^2 + (1 + \frac{1}{R} ) \sigma^2 \right)
\end{dmath}
    
\end{corollary}

\begin{proof}

If $ \gamma = \gamma_0  \frac{1}{L E \sqrt{R}} $, we have that

\begin{equation}
 \delta_{E, \gamma, L} = \frac{4 E^2 \gamma^2 L^2}{1 - 8 E^2 \gamma^2 L^2} = \frac{4 \gamma_0^2 / R}{1 - 8 \gamma_0^2 / R} = \mathcal{O} \left( \frac{1}{R} \right),
\end{equation}

and, by recalling that $ \mathcal{O} \left( a + \frac{1}{R} \right) = a $, the convergence rate of Corollary \ref{corollary:fed_avg_final_conv} is obtained.

\end{proof}

\section{SCAFFOLD convergence}
\label{app:scaffold_convergence_proof}

\subsection{SCAFFOLD - option \RNum{1}}
    
\begin{lemma}

The convergence rate of SCAFFOLD with option \RNum{1} is given by,

\begin{dmath}
\frac{1}{R} \sum_{r=1}^R \mathbb{E} \left[ \| \nabla_w F ( \bar{w}_{r-1} ; S^{r,0} ) \|^2 \right]  \\
\leq \frac{1}{1 - 2 \delta_{E, \gamma, L}} \left[ \frac{2}{\gamma R E} \left[ F( \bar{w}_0 ; S^{1,0}) - \underbar{F} \right]
+ 2 ( 1 + 2 \delta_{E, \gamma, L}) B^2
+ 8 \delta_{E, \gamma, L} \frac{1}{R} \Delta^2 c^0
+ ( 1 + 2 \delta_{E, \gamma, L} ) \sigma^2 \right],
\end{dmath}

with $ \gamma < \frac{1}{4 L E} $.

\end{lemma}

\begin{proof}

$ \mathcal{T}_2 $ depends on the initialization of the control variates:

\begin{dmath}
    \label{eq:scaffold_op1_intermediary_3}
    \mathcal{T}_2 = \frac{8}{R} \delta_{E, \gamma, L} \sum_{j=1}^N P_j \| \nabla_w F_j ( \bar{w}_0; S^{1,0}_j ) - c^0_j \|^2 = \frac{8}{R} \delta_{E, \gamma, L} \Delta^2 c^0.
\end{dmath}

where $ \sum_{j=1}^N P_j \| \nabla_w F_j ( \bar{w}_0; S^{1,0}_j ) - c^0_j \|^2 \coloneq \Delta^2 c^0 $.

By setting $ c_j^r = \nabla_w F_j ( w_{r-1} ; S^{r, 0}_j ) $ and $\tilde{S}_j^{r, 0} = S_j^{r, 0} $, $ \mathcal{T}_3 $ yields

\begin{dmath}
    \label{eq:scaffold_op1_intermediary_1}
    \mathcal{T}_3
    = \frac{2}{R} (1 + \delta_{E, \gamma, L}) \sum_{r=1}^R \mathbb{E} \left[ \sum_{j=1}^N P_j \| \nabla_w F_j ( \bar{w}_{r-1} ; S^{r,0} ) - \nabla_w F_j ( \bar{w}_{r-1} ; S^{r,0}_j ) \|^2 \right] \leq 2 (1 + \delta_{E, \gamma, L}) B^2,
\end{dmath}

and $ \mathcal{T}_4 $ yields,

\begin{equation}    
    \label{eq:scaffold_op1_intermediary_2}
    \mathcal{T}_4 = \frac{16}{R} \delta_{E, \gamma, L} \sum_{r=1}^R \mathbb{E} \left[ \sum_{j=1}^N P_j \| \nabla_w F_j ( \bar{w}_{r -1} ; S^{r, 0}_j ) - \nabla_w F_j ( \bar{w}_{r -1} ; S^{r, 0}_j ) \|^2 \right] =  0.
\end{equation}

By setting $ \gamma$ small enough such that $ 1 - \gamma L E - 16 L^2 E^2 \gamma^2 \delta_{E, \gamma, L} \geq 0 $, we get

\begin{dmath}
\frac{1}{R} \left( 1- 2 \delta_{E, \gamma, L} \right) \sum_{r=1}^R \mathbb{E} \left[ \| \nabla_w F ( \bar{w}_{r-1} ; S^{r,0} ) \|^2 \right]  \\
\leq \frac{2}{\gamma R E} \left[ F( \bar{w}_0 ; S^{1,0}) - \underbar{F} \right]
+ 2 ( 1 + 2 \delta_{E, \gamma, L}) B^2
+ 8 \delta_{E, \gamma, L} \frac{1}{R} \Delta^2 c^0
+ ( 1 + 2 \delta_{E, \gamma, L} ) \sigma^2.
\end{dmath}

The following conditions on the learning rate need to be verified:

\begin{equation}
    \begin{cases}
    1- 2 \delta_{E, \gamma, L} > 0 \\
    1 - \gamma L E - 16 L^2 E^2 \gamma^2 \delta_{E, \gamma, L} \geq 0.
    \end{cases}
\end{equation}

The first condition is met if and only if $ \gamma < \frac{1}{4 L E} $. For the second condition, we repeat the procedure used in the proof of Lemma \ref{lemma:fed_avg_convergence} to obtain $ \gamma \leq \frac{1}{2LE} $ and $ \gamma \leq \frac{1}{4 L E} $. Thus,

\begin{equation}
    \gamma < \frac{1}{4 L E}.
\end{equation}

\end{proof}

\begin{corollary}
    By taking $ \gamma = \gamma_0  \frac{1}{L E \sqrt{R}} $, we obtain
    
    \begin{dmath}
        \min_r \mathbb{E} \left[ \| \nabla_w F ( \bar{w}_{r-1} ; S^{r,0} ) \|^2 \right] \\
        \leq 
        \frac{1}{R} \sum_{r=1}^R \mathbb{E} \left[ \| \nabla_w F ( \bar{w}_{r-1} ; S^{r,0} ) \|^2 \right] \\
        \leq \mathcal{O} \left( \frac{L}{\sqrt{R}} \left[ F_0 - \underbar{F} \right] + (1 + \frac{1}{R} ) B^2 + \frac{1}{R^2} \Delta^2c^0 + (1 + \frac{1}{R} ) \sigma^2 \right).
    \end{dmath}
    
\end{corollary}

\begin{proof}

    Refer to the proof of Corollary \ref{corollary:fed_avg_final_conv}.
    
\end{proof}

\subsection{SCAFFOLD - option \RNum{2}}

\begin{lemma}

The convergence of SCAFFOLD with option \RNum{2} is given by

\begin{dmath}
\frac{1}{R} \sum_{r=1}^R \mathbb{E} \left[ \| \nabla_w F ( \bar{w}_{r-1} ; S^{r,0} ) \|^2 \right]  \\
\leq \frac{1}{1 - 50/21 \delta_{E, \gamma, L}} \left[ \frac{2}{\gamma R E} \left[ F_0 - \underbar{F} \right]
+ 2 \left( 1 + \frac{28}{21} \delta_{E, \gamma, L} \right) B^2
+ \frac{184}{7} \delta_{E, \gamma, L} \frac{1}{R} \Delta^2 c^0
+ \left( 1 + 2 \delta_{E, \gamma, L} ( 1 + \frac{193}{21} \frac{1}{E} ) \right) \sigma^2 \right],
\end{dmath}

with $ \gamma <  \sqrt{\frac{21}{368}} \frac{1}{L E} $.
    
\end{lemma}

\begin{proof}
The terms $ \mathcal{T}_2 $ and $ \mathcal{T}_3 $ are equal to those obtained from SCAFFOLD option \RNum{1}, in Equations \eqref{eq:scaffold_op1_intermediary_3} and \eqref{eq:scaffold_op1_intermediary_1}. For $ \mathcal{T}_4 $, by assuming $ \gamma \leq \frac{1}{24 L E} $, we have

\begin{dmath}
    \label{eq:scaffold_op2_intermediary_1}
    \sum_{j=1}^N P_j \mathbb{E} \left[ \| \nabla_w F_j ( \bar{w}_{r -1} ; S^{r, 0}_j ) - c^r_j \|^2 \right] \\
    = \sum_{j=1}^N P_j \mathbb{E} \left[ \| \nabla_w F_j ( \bar{w}_{r -1} ; S^{r, 0}_j ) - \frac{1}{E} \sum_{t=1}^E g_j (w_j^{r,t-1} ; s^{r,t-1}_j) \|^2 \right] \\
    \leq \frac{1}{E} \sum_{j=1}^N P_j \sum_{t=1}^E \mathbb{E} \left[ \| \nabla_w F_j ( \bar{w}_{r -1} ; S^{r, 0}_j ) -  \nabla F_j (w_j^{r,t-1} ; S^{r,t-1}_j) \|^2 \right] + \frac{\sigma^2}{E} \\
    \leq \frac{L^2}{E} \sum_{j=1}^N P_j \sum_{t=1}^E \mathbb{E} \left[ \| w_j^{r,t-1} - \bar{w}_{r -1} \|^2 \right] + \frac{\sigma^2}{E}.
\end{dmath}

where Lemma \ref{lemma:separating_mean_and_var} and the $L$-Lipschitzness were used. Using Lemma \ref{lemma:client_drift_bound} with $ \tilde{S}^{r,0}_j = S^{r,0}_j $ yields

\begin{dmath}
    \label{eq:scaffold_op2_intermediary_2}
    \sum_{j=1}^N P_j \mathbb{E} \left[ \| \nabla_w F_j ( \bar{w}_{r -1} ; S^{r, 0}_j ) - c^r_j \|^2 \right] \\
    \leq
    \frac{193}{192} \frac{\sigma^2}{E} 
    + \frac{1}{48} \mathbb{E} \left[ \| \nabla_w F ( \bar{w}_{r-1} ; S^{r,0}) - c^{r-1} \|^2 \right] \\
    + \frac{1}{48} \sum_{j=1}^N P_j \ \mathbb{E} \left[ \| \nabla_w F_j ( \bar{w}_{r-1} ; S^{r,0}_j) - c^{r-1}_j \|^2 \right]
    + \frac{1}{48} \mathbb{E} \left[ \| \nabla_w F ( \bar{w}_{r-1} ; S^{r,0}) \|^2 \right] \\
    \leq
    \frac{193}{192} \frac{\sigma^2}{E} 
    + \frac{1}{24} \sum_{j=1}^N P_j \mathbb{E} \left[ \| \nabla_w F_j ( \bar{w}_{r-1} ; S^{r,0}_j) - c^{r-1}_j \|^2 \right] + \frac{1}{24} \sum_{j=1}^N P_j \mathbb{E} \left[ \| \nabla_w F_j ( \bar{w}_{r-1} ; S^{r,0}) - \nabla_w F_j ( \bar{w}_{r-1} ; S^{r,0}_j)  \|^2 \right] \\
    + \frac{1}{48} \sum_{j=1}^N P_j \ \mathbb{E} \left[ \| \nabla_w F_j ( \bar{w}_{r-1} ; S^{r,0}_j) - c^{r-1}_j \|^2 \right]
    + \frac{1}{48} \ \mathbb{E} \left[ \| \nabla_w F ( \bar{w}_{r-1} ; S^{r,0}) \|^2 \right] \\\
    \leq
    \frac{193}{192} \frac{\sigma^2}{E} + \frac{1}{24} B^2 
    + \frac{3}{48} \sum_{j=1}^N P_j \ \mathbb{E} \left[ \| \nabla_w F_j ( \bar{w}_{r-1} ; S^{r,0}_j) - c^{r-1}_j \|^2 \right]
    + \frac{1}{48} \mathbb{E} \left[ \| \nabla_w F ( \bar{w}_{r-1} ; S^{r,0}) \|^2 \right]. 
\end{dmath}

By using Lemma \ref{lemma:grad_vs_control_var} with $ r > 1$, we have the following recursion

\begin{dmath}
\label{eq:scaffold_op2_intermediary_3}
    \underbrace{\sum_{j=1}^N P_j \mathbb{E} \left[ \| \nabla_w F_j ( \bar{w}_{r -1} ; S^{r, 0}_j ) - c^r_j \|^2 \right]}_{u_r} \\
    \leq
    \underbrace{\frac{193}{192} \frac{\sigma^2}{E} + \frac{1}{24} B^2 
    + \frac{3}{24} \gamma^2 L^2 E \sum_{t=1}^E \Big\| \sum_{i=1}^N P_i g_i ( w^{r-1,t-1}_i; s^{r-1,t-1}_i ) \Big\|^2  \\
    + \frac{1}{48} \mathbb{E} \left[ \| \nabla_w F ( \bar{w}_{r-1} ; S^{r,0}) \|^2 \right]}_{b_{r-1}}
    + \underbrace{\frac{3}{24}}_{a} \underbrace{\sum_{j=1}^N P_j \mathbb{E} \left[ \| \nabla_w F_j ( \bar{w}_{r-2} ; S^{r-1,0}_j) - c^{r-1}_j \|^2 \right]}_{u_{r-1}}. 
\end{dmath}

and we can use Lemma \ref{lemma:recursion}.

\begin{dmath}
    \frac{1}{R} \sum_{r=1}^R \sum_{j=1}^N P_j \mathbb{E} \left[ \| \nabla_w F_j ( \bar{w}_{r -1} ; S^{r, 0}_j ) - c^r_j \|^2 \right] \\
    \leq \frac{1}{R} \frac{1 - ( 3/24 )^R}{1 - 3 / 24} \left( \sum_{j=1}^N P_j \mathbb{E} \left[ \| \nabla_w F_j ( \bar{w}_{0} ; S^{1, 0}_j ) - c^1_j \|^2 \right] + \frac{193}{192} \frac{\sigma^2}{E} (R-1) + \frac{1}{24} (R-1) B^2 \\ + \frac{3}{24} \gamma^2 L^2 E \sum_{r=1}^{R-1} \sum_{t=1}^E \Big\| \sum_{i=1}^N P_i g_i ( w^{r,t-1}_i; s^{r,t-1}_{D_i} ) \Big\|^2
    + \frac{1}{48} \sum_{r=2}^R  \mathbb{E} \left[ \| \nabla_w F ( \bar{w}_{r-1} ; S^{r,0}) \|^2 \right]  \right).
\end{dmath}

And by using Equation \eqref{eq:scaffold_op2_intermediary_2} with $r=1$

\begin{dmath}
    \label{eq:scaffold_op2_intermediary_4}
    \frac{1}{R} \sum_{r=1}^R \sum_{j=1}^N P_j \mathbb{E} \left[ \| \nabla_w F_j ( \bar{w}_{r -1} ; S^{r, 0}_j ) - c^r_j \|^2 \right] \\
    \leq \frac{1 - ( 3/24 )^R}{1 - 3 / 24} \left( \frac{1}{R} \sum_{j=1}^N P_j \mathbb{E} \left[ \| \nabla_w F_j ( \bar{w}_{0} ; S^{1, 0}_j ) - c^0_j \|^2 \right] + \frac{193}{192} \frac{\sigma^2}{E} + \frac{1}{24} B^2 \\ + \frac{3}{24} \gamma^2 L^2 E \frac{1}{R} \sum_{r=1}^R \sum_{t=1}^E \Big\| \sum_{i=1}^N P_i g_i ( w^{r,t-1}_i; s^{r,t-1}_{D_i} ) \Big\|^2
    + \frac{1}{48} \frac{1}{R} \sum_{r=1}^R  \mathbb{E} \left[ \| \nabla_w F ( \bar{w}_{r-1} ; S^{r,0}) \|^2 \right]  \right) \\
    \leq \frac{1 - ( 3/24 )^R}{1 - 3 / 24} \left( \frac{1}{R} \Delta^2c^0 + \frac{193}{192} \frac{\sigma^2}{E} + \frac{1}{24} B^2 \\ + \frac{3}{24} \gamma^2 L^2 E \frac{1}{R} \sum_{r=1}^R \sum_{t=1}^E \Big\| \sum_{i=1}^N P_i g_i ( w^{r,t-1}_i; s^{r,t-1}_{D_i} ) \Big\|^2
    + \frac{1}{48} \frac{1}{R} \sum_{r=1}^R  \mathbb{E} \left[ \| \nabla_w F ( \bar{w}_{r-1} ; S^{r,0}) \|^2 \right]  \right) \\
    \leq  \frac{8}{7}\frac{1}{R} \Delta^2c^0 + \frac{193}{168} \frac{\sigma^2}{E} + \frac{1}{21} B^2 + \frac{1}{7} \gamma^2 L^2 E \frac{1}{R} \sum_{r=1}^R \sum_{t=1}^E \Big\| \sum_{i=1}^N P_i g_i ( w^{r,t-1}_i; s^{r,t-1}_{D_i} ) \Big\|^2
    + \frac{1}{42} \frac{1}{R} \sum_{r=1}^R  \mathbb{E} \left[ \| \nabla_w F ( \bar{w}_{r-1} ; S^{r,0}) \|^2 \right],
\end{dmath}

as $ \frac{1 - ( 3/24 )^R}{1 - 3 / 24} \leq \frac{1}{1 - 3 / 24} = \frac{8}{7} $. Thus, we have

\begin{dmath}
    \label{eq:scaffold_option_2_t4}
    \mathcal{T}_4 \leq 16 \delta_{E, \gamma, L} \left[ \frac{8}{7}\frac{1}{R} \Delta^2c^0 + \frac{193}{168} \frac{\sigma^2}{E} + \frac{1}{21} B^2 + \frac{1}{7} \gamma^2 L^2 E \frac{1}{R} \sum_{r=1}^R \sum_{t=1}^E \Big\| \sum_{i=1}^N P_i g_i ( w^{r,t-1}_i; s^{r,t-1}_{D_i} ) \Big\|^2
    + \frac{1}{42} \frac{1}{R} \sum_{r=1}^R  \mathbb{E} \left[ \| \nabla_w F ( \bar{w}_{r-1} ; S^{r,0}) \|^2 \right] \right].
\end{dmath}

Plugging $\mathcal{T}_2$, $\mathcal{T}_3$, and $\mathcal{T}_4$ into Equation \eqref{eq:convergence_var_reduction_algorithms_async} yields

\begin{dmath}
\frac{1}{R} \left( 1 - 50/21 \delta_{E, \gamma, L} \right) \sum_{r=1}^R \mathbb{E} \left[ \| \nabla_w F ( \bar{w}_{r-1} ; S^{r,0} ) \|^2 \right]  \\
\leq \frac{2}{\gamma R E} \left[ F( \bar{w}_0 ; S^{1,0}) - \underbar{F} \right]
+ 2 \left( 1 + \frac{28}{21} \delta_{E, \gamma, L} \right) B^2
+ \frac{184}{7} \delta_{E, \gamma, L} \frac{1}{R} \Delta^2 c^0
+ \left( 1 + 2 \delta_{E, \gamma, L} ( 1 + \frac{193}{21} \frac{1}{E} ) \right) \sigma^2 \\
- \frac{1}{R E} \left( 1 - \gamma L E - \frac{128}{7} L^2 E^2 \gamma^2 \delta_{E, \gamma, L} \right) \sum_{r=1}^R \sum_{t=1}^E \mathbb{E} \left[ \Big\| \sum_{j=1}^N P_j \ g_j( w_j^{r, t-1} ; s_{D_j}^{r, t-1}) \Big\|^2 \right].
\end{dmath}

And by setting $\gamma$ sufficiently small such that $ 1 - \gamma L E - \frac{128}{7} L^2 E^2 \gamma^2 \delta_{E, \gamma, L} \geq 0 $ we get,

\begin{dmath}
\frac{1}{R} \left( 1- 50/21 \delta_{E, \gamma, L} \right) \sum_{r=1}^R \mathbb{E} \left[ \| \nabla_w F ( \bar{w}_{r-1} ; S^{r,0} ) \|^2 \right]  \\
\leq \frac{2}{\gamma R E} \left[ F( \bar{w}_0 ; S^{1,0}) - \underbar{F} \right]
+ 2 \left( 1 + \frac{28}{21} \delta_{E, \gamma, L} \right) B^2
+ \frac{184}{7}\delta_{E, \gamma, L} \frac{1}{R} \Delta^2 c^0
+ \left( 1 + 2 \delta_{E, \gamma, L} ( 1 + \frac{193}{21} \frac{1}{E} ) \right) \sigma^2.
\end{dmath}

The following conditions on the learning rate need to be verified:

\begin{equation}
    \begin{cases}
    1 - 50/21 \delta_{E, \gamma, L} > 0 \\
    1 - \gamma L E - \frac{128}{7} L^2 E^2 \gamma^2 \delta_{E, \gamma, L} \geq 0.
    \end{cases}
\end{equation}

The first condition is verified if $ \gamma < \sqrt{\frac{21}{368}} \frac{1}{L E} $. Again, we seek to avoid calculating the roots of a forth order polynomial. The second condition is then split into two conditions: $ 1/2 - \gamma L E \geq 0 $, $ 1/2 - 128/7 L^2 E^2 \gamma^2 \delta_{E, \gamma, L} \geq 0 $. This imposes two additional conditions on the learning rate: $ \frac{1}{2 L E} $ and $ \gamma \leq \frac{\sqrt{-7 + \sqrt{497}}}{16} \frac{1}{L E}$. Taking the most restrictive condition, we have,

\begin{equation}
    \gamma <  \sqrt{\frac{21}{368}} \frac{1}{L E},
\end{equation}

as $ \sqrt{\frac{21}{368}} \approx 0.239 $ and $ \frac{\sqrt{-7 + \sqrt{497}}}{16} \frac{1}{L E} \approx 0.244 $.

\end{proof}

\begin{corollary}
    By taking $ \gamma = \gamma_0  \frac{1}{L E \sqrt{R}} $, we obtain
    
    \begin{dmath}
        \min_r \mathbb{E} \left[ \| \nabla_w F ( \bar{w}_{r-1} ; S^{r,0} ) \|^2 \right] \\
        \leq 
        \frac{1}{R} \sum_{r=1}^R \mathbb{E} \left[ \| \nabla_w F ( \bar{w}_{r-1} ; S^{r,0} ) \|^2 \right] \\
        \leq \mathcal{O} \left( \frac{L}{\sqrt{R}} \left[ F_0 - \underbar{F} \right] + (1 + \frac{1}{R} ) B^2 + \frac{1}{R^2} \Delta^2c^0 + \left( 1 + \frac{1}{R} (1 + \frac{1}{E}) \right) \sigma^2 \right).
    \end{dmath}
    
\end{corollary}

\begin{proof}
    Refer to the proof of Corollary \ref{corollary:fed_avg_final_conv}.
\end{proof}

\section{BN-SCAFFOLD convergence}
\label{app:bn_scaffold_convergence_proof}

\subsection{BN-SCAFFOLD - option \RNum{1}}

\begin{lemma}

The convergence of BN-SCAFFOLD with option \RNum{1} is given by 

\begin{dmath}
\frac{1}{R} \sum_{r=1}^R \mathbb{E} \left[ \| \nabla_w F ( \bar{w}_{r-1} ; S^{r,0} ) \|^2 \right]  \\
\leq \frac{1}{1 - 2 \delta_{E, \gamma, L}} \left[ \frac{2}{\gamma R E} \left[ F_0 - \underbar{F} \right]
+ 8 \delta_{E, \gamma, L} \frac{1}{R} \Delta^2 c^0 + (1 + 2 \delta_{E, \gamma, L} ) \sigma^2 \right],
\end{dmath}

with

\begin{equation}
    \begin{cases}
    \gamma < \frac{1}{4 L E} \\
    \gamma \leq \min \left\{ \sqrt{\frac{-8 L^2 + \sqrt{64 L^4 + 4 (L^2 + M^2 J^2) L^2}}{384 (L^2 + M^2 J^2)}} \frac{1}{L} ;\frac{1}{\sqrt{24} J M} \right\} \frac{1}{E}.
    \end{cases}
\end{equation}
    
\end{lemma}

\begin{proof}

As in SCAFFOLD - option \RNum{1}, setting $ c_j^r = \nabla_w F_j ( w_{r-1} ; \tilde{S}^{r, 0}_j ) $, the second and fourth terms of the right side of Equation \eqref{eq:convergence_var_reduction_algorithms_async} yields

\begin{dmath}   
    \label{eq:bn_scaffold_op1_intermediary_1}
    \mathcal{T}_4 = \frac{16}{R} \delta_{E, \gamma, L} \sum_{r=1}^R \mathbb{E} \left[ \sum_{j=1}^N P_j \| \nabla_w F_j ( \bar{w}_{r -1} ; \tilde{S}^{r, 0}_j ) - c^r_j \|^2 \right] \\ 
    =  \frac{16}{R} \delta_{E, \gamma, L} \sum_{r=1}^R \mathbb{E} \left[ \sum_{j=1}^N P_j \| \nabla_w F_j ( \bar{w}_{r -1} ; \tilde{S}^{r, 0}_j ) - \nabla_w F_j ( \bar{w}_{r -1} ; \tilde{S}^{r, 0}_j ) \|^2 \right] =  0,
\end{dmath}

and,

\begin{dmath}
    \label{eq:bn_scaffold_op1_intermediary_2}
    \mathcal{T}_2 = \frac{8}{R} \delta_{E, \gamma, L}  \sum_{j=1}^N P_j \| \nabla_w F_j ( \bar{w}_{0} ; \tilde{S}^{1, 0}_j )- c^{0}_j \|^2 
    = \frac{8}{R} \delta_{E, \gamma, L} \Delta^2 c^0.
\end{dmath}

By setting

\begin{equation}
\begin{cases}
    \tilde{S}_j^{r, 0} = S_j^{r, 0} + k^{r-1} - k^{r-1}_j \\
    k^r_j =  S_j^{r, 0} \\
    k^r = \sum_{j=1}^N P_j k^r_j =  \sum_{j=1}^N P_j S_j^{r, 0} = S^{r, 0},
\end{cases}
\end{equation}

$ \mathcal{T}_3 $ yields

\begin{dmath}
    \label{eq:bn_scaffold_op1_intermediary_3}
    \mathcal{T}_3 = \frac{2}{R} ( 1 + 2 \delta_{E, \gamma, L}) \sum_{r=1}^R  \mathbb{E} \left[ \sum_{j=1}^N P_j \| \nabla_w F_j ( \bar{w}_{r-1} ; S^{r,0} ) - \nabla_w F_j ( \bar{w}_{r-1} ; \tilde{S}^{r,0}_j ) \|^2 \right] \\
    = \frac{2}{R} ( 1 + 2 \delta_{E, \gamma, L}) \sum_{r=1}^R \mathbb{E} \left[ \sum_{j=1}^N P_j \| \nabla_w F_j ( \bar{w}_{r-1} ; S^{r,0} ) - \nabla_w F_j ( \bar{w}_{r-1} ; S^{r,0}_j - S^{r-1,0}_j + S^{r-1,0} ) \|^2 \right] \\ 
    \leq \frac{2}{R} J^2 ( 1 + 2 \delta_{E, \gamma, L}) \sum_{r=1}^R \mathbb{E} \left[ \sum_{j=1}^N P_j \| S^{r,0} - S^{r,0}_j + S^{r-1,0}_j - S^{r-1,0}  \|^2 \right] \\
    \leq \frac{4}{R} J^2 ( 1 + 2 \delta_{E, \gamma, L}) \sum_{r=1}^R \left( \mathbb{E} \left[ \| S^{r,0} - S^{r-1,0} \|^2 \right] + \mathbb{E} \left[ \sum_{j=1}^N P_j \| S^{r,0}_j - S^{r-1,0}_j  \|^2 \right] \right) \\
    \leq \frac{8}{R} J^2 ( 1 + 2 \delta_{E, \gamma, L}) \sum_{r=1}^R \mathbb{E} \left[ \sum_{j=1}^N P_j \| S^{r,0}_j - S^{r-1,0}_j  \|^2 \right] \\
    \leq \frac{8}{R} J^2 M^2 ( 1 + 2 \delta_{E, \gamma, L}) \sum_{r=1}^R \mathbb{E} \left[ \| \bar{w}_{r-1} - \bar{w}_{r-2} \|^2 \right] \\
    \leq \frac{8}{R} J^2 M^2 \gamma^2 E ( 1 + 2 \delta_{E, \gamma, L}) \sum_{r=1}^R \sum_{t=1}^E \mathbb{E} \left[ \Big\| \sum_{j=1}^N g_j (w^{r, t-1}_j ; \tilde{s}^{r, t-1}_{D_j} ) \Big\|^2 \right], 
\end{dmath}

where Assumptions \ref{assumption_app:l_lipschitz_continuity_statistics}, Lemma \ref{lemma2}, and Jensen's inequality were used. We thus obtain

\begin{dmath}
\frac{1}{R} \left( 1 - 2 \delta_{E, \gamma, L} \right) \sum_{r=1}^R \mathbb{E} \left[ \| \nabla_w F ( \bar{w}_{r-1} ; S^{r,0} ) \|^2 \right]  \\
\leq \frac{2}{\gamma R E} \left[ F( \bar{w}_0 ; S^{1,0}) - \underbar{F} \right]
+ 8 ( 1 + 2 \delta_{E, \gamma, L}) J^2 M^2 \gamma^2 E \frac{1}{R} \sum_{r=1}^R \sum_{t=1}^E \mathbb{E} \left[ \Big\| \sum_{j=1}^N g_j (w^{r, t-1}_j ; \tilde{s}^{r, t-1}_{D_j} ) \Big\|^2 \right] \\
+ 8 \delta_{E, \gamma, L} \frac{1}{R} \Delta^2 c^0
+ ( 1 + 2 \delta_{E, \gamma, L} ) \sigma^2 \\
- \frac{1}{R E} (1 - \gamma L E - 16 L^2 E^2 \gamma^2 \delta_{E, \gamma, L} ) \sum_{r=1}^R \sum_{t=1}^E \mathbb{E} \left[ \Big\| \sum_{j=1}^N P_j \ g_j( w_j^{r, t-1} ; \tilde{s}_j^{r, t-1}) \Big\|^2 \right] \\
\leq \frac{2}{\gamma R E} \left[ F( \bar{w}_0 ; S^{1,0}) - \underbar{F} \right]
+ 8 \delta_{E, \gamma, L} \frac{1}{R} \Delta^2 c^0
+ ( 1 + 2 \delta_{E, \gamma, L} ) \sigma^2 \\
- \frac{1}{R E} (1 - \gamma L E - 16 L^2 E^2 \gamma^2 \delta_{E, \gamma, L} - ( 1 + 2 \delta_{E, \gamma, L}) 8 J^2 M^2 \gamma^2 E^2 ) \sum_{r=1}^R \sum_{t=1}^E \mathbb{E} \left[ \Big\| \sum_{j=1}^N P_j \ g_j( w_j^{r, t-1} ; \tilde{s}_j^{r, t-1}) \Big\|^2 \right].
\end{dmath}

By setting $ \gamma $ such that $ 1 - \gamma L E - 16 L^2 E^2 \gamma^2 \delta_{E, \gamma, L} - ( 1 + 2 \delta_{E, \gamma, L}) 8 J^2 M^2 \gamma^2 E^2 \geq 0 $, the last term can be removed from the bound and we obtain,

\begin{dmath}
\frac{1}{R} \left( 1 - 2 \delta_{E, \gamma, L} \right) \sum_{r=1}^R \mathbb{E} \left[ \| \nabla_w F ( \bar{w}_{r-1} ; S^{r,0} ) \|^2 \right]  \\
\leq \frac{2}{\gamma R E} \left[ F( \bar{w}_0 ; S^{1,0}) - \underbar{F} \right]
+ 8 \delta_{E, \gamma, L} \frac{1}{R} \Delta^2 c^0 + ( 1 + 2 \delta_{E, \gamma, L} ) \sigma^2.
\end{dmath}

When have the following conditions on the learning rate

\begin{equation}
    \begin{cases}
        1 - 2 \delta_{E, \gamma, L} > 0 \\
        1 - \gamma L E - 16 L^2 E^2 \gamma^2 \delta_{E, \gamma, L} - ( 1 + 2 \delta_{E, \gamma, L}) 8 J^2 M^2 \gamma^2 E^2 \geq 0.
    \end{cases}
\end{equation}

The first condition imposes $ \gamma < \frac{1}{4 L E} $. As done for the other algorithms, we divide the second condition in three conditions: $ 1/3 - \gamma L E \geq 0 $, $ 1/3 - 16 L^2 E^2 \gamma^2 \delta_{E, \gamma, L} \geq 0 $, and $ 1/3 - (1 + \delta_{E, \gamma, L}) 8 J^2 M^2 \gamma^2 E^2 \geq 0 $. This imposes the following conditions on the learning rate:

\begin{equation}
    \begin{cases}
        \gamma \leq \frac{1}{3 L E} \\
        \gamma \leq \sqrt{\frac{-8 L^2 + \sqrt{64 L^4 + 4 (L^2 + M^2 J^2) L^2}}{384 (L^2 + M^2 J^2)}} \frac{1}{L E} \\
        \gamma \leq \frac{1}{\sqrt{24} J M E}.
    \end{cases}
\end{equation}

We thus have

\begin{equation}
    \begin{cases}
    \gamma < \frac{1}{4 L E} \\
    \gamma \leq \min \left\{ \sqrt{\frac{-8 L^2 + \sqrt{64 L^4 + 4 (L^2 + M^2 J^2) L^2}}{384 (L^2 + M^2 J^2)}} \frac{1}{L} ;\frac{1}{\sqrt{24} J M} \right\} \frac{1}{E}.
    \end{cases}
\end{equation}

\end{proof}

\begin{corollary}
    By taking $ \gamma = \gamma_0  \frac{1}{L E \sqrt{R}} $, we obtain
    
    \begin{dmath}
        \min_r \mathbb{E} \left[ \| \nabla_w F ( \bar{w}_{r-1} ; S^{r,0} ) \|^2 \right] \\
        \leq 
        \frac{1}{R} \sum_{r=1}^R \mathbb{E} \left[ \| \nabla_w F ( \bar{w}_{r-1} ; S^{r,0} ) \|^2 \right] \\
        \leq \mathcal{O} \left( \frac{L}{\sqrt{R}} \left[ F_0 - \underbar{F} \right] + \frac{1}{R^2} \Delta^2c^0 + (1 + \frac{1}{R} ) \sigma^2 \right).
    \end{dmath}
    
\end{corollary}

\begin{proof}
    Refer to the proof of Corollary \ref{corollary:fed_avg_final_conv}.
\end{proof}

\subsection{BN-SCAFFOLD - option \RNum{2}}

\begin{lemma}
\label{lemma:bn_scaffold_option_2_convergence}

The convergence of BN-SCAFFOLD with option \RNum{2} is given by

\begin{dmath}
\label{eq:bn_scaffold_option2_convergence}
\frac{1}{R} \sum_{r=1}^R \mathbb{E} \left[ \| \nabla_w F_i ( \bar{w}_{r-1}; S^{r,0} ) \|^2 \right] 
\leq \\
\frac{1}{1 - 2 \delta_{E, \gamma, L} \left( 1 +  \alpha^*_{E, \gamma, L, M, J} \right)} \left[ \\
\frac{2}{\gamma R E} \left[ F_0 - \underbar{F} \right]
%
+ \left[ 4 J^2 \delta_{E, \gamma, L} \alpha^*_{E, \gamma, L, M, J} + (1 + \delta_{E, \gamma, L}) \right] \frac{1}{R} \Delta^2 k^0 \\
+ 2 \delta_{E, \gamma, L} \left[ 3 \alpha^*_{E, \gamma, L, M, J} + 4 \right] \frac{1}{R} \Delta^2 c^0 \\
+ \left[ \frac{1}{E} \frac{193}{2} \delta_{E, \gamma, L} \alpha^*_{E, \gamma, L, M, J} + ( 1 + 2 \delta_{E, \gamma, L} ) \right] \sigma^2 
+ 2 J^2  (1 + \delta_{E, \gamma, L}) \sigma^2_s \right],
\end{dmath}

where $ \alpha^*_{E, \gamma, L, M, J} \coloneq \frac{1 + 2 \alpha_{E, \gamma, L, M, J}}{1 - \alpha_{E, \gamma, L, M, J}} $ with $\alpha_{E, \gamma, L, M, J} \coloneq \frac{M^2 J^2}{L^2} \frac{\delta_{E, \gamma, L} + 1}{\delta_{E, \gamma, L}}$. The following set of conditions need to be verified

\begin{equation}
    \begin{cases}        
    \gamma \leq \min \left\{ \frac{1}{24 L}; \sqrt{\frac{-8 L^2 + \sqrt{64 L^4 + 48 L^{*2} L^2}}{24 L^{*2} L^2}}; \frac{1}{\sqrt{48}J M} \right\} \frac{1}{E} \\[10pt]
    \gamma \geq \frac{M J}{2 \sqrt{L^2 + M^2 J^2} L E} \\[10pt]
    M^2 J^2 < L^2 / 3.
    \end{cases}
\end{equation}

where 

\begin{equation}
    L^{*2} \coloneq 16 L^2 + \left( 16 + \frac{24}{1 - \alpha_{E, \gamma, L, M, J}} \right) M^2 J^2 + \frac{3L^4+32 M^4 J^4 + 20 M^2 J^2 L^2}{ (1 - \alpha_{E, \gamma, L, M, J}) L^2}.
\end{equation}

\end{lemma}

\begin{proof}

In this case, it is convenient to analyze $ \mathcal{T}_3 $ and $ \mathcal{T}_4 $ from Theorem \ref{theorem:variance_reduc_algos} together. We define

\begin{equation}
    \mathcal{T}'_3 \coloneq \sum_{j=1}^N P_j \mathbb{E} \left[ \| \nabla_w F_j ( \bar{w}_{r-1} ; S^{r,0} ) - \nabla_w F_j ( \bar{w}_{r-1} ; \tilde{S}^{r,0}_j ) \|^2 \right],
\end{equation}

and

\begin{equation}
    \mathcal{T}'_4 \coloneq \sum_{j=1}^N P_j \mathbb{E} \left[ \| \nabla_w F_j ( \bar{w}_{r -1} ; \tilde{S}^{r, 0}_j ) - c^r_j \|^2 \right].
\end{equation}

By using Lemmas \ref{lemma:separating_mean_and_var} and \ref{lemma:client_drift_bound}, while assuming $ \gamma \leq \frac{1}{24 L E} $, $ \mathcal{T}'_4 $ can be re-written as

\begin{dmath}
    \label{eq:bn_scaffold_option_2_intermediary_eq_2}
    \mathcal{T}^{'}_4
    \leq
    \frac{193}{192} \frac{\sigma^2}{E} 
    + \frac{1}{48} \mathbb{E} \left[ \| \nabla_w F ( \bar{w}_{r-1} ; S^{r,0}) - c^{r-1} \|^2 \right] \\
    + \frac{1}{48} \sum_{j=1}^N P_j \ \mathbb{E} \left[ \| \nabla_w F_j ( \bar{w}_{r-1} ; \tilde{S}^{r,0}_j) - c^{r-1}_j \|^2 \right]
    + \frac{1}{48} \mathbb{E} \left[ \| \nabla_w F ( \bar{w}_{r-1} ; S^{r,0}) \|^2 \right].
\end{dmath}

By using Lemmas \ref{lemma:bn_scaffold_global_grad_global_control_variate} and \ref{lemma:grad_vs_control_var} we have, for $r \geq 2$,

\begin{dmath}
    \label{eq:bn_scaffold_option_2_intermediary_eq_t4}
    \mathcal{T}^{'}_4
    \leq
    \frac{1}{3} J^2 M^2 \mathbb{E} \left[ \| \bar{w}_{r-1} - \bar{w}_{r-2} \|^2 \right] + \frac{1}{3} J^2 \sum_{j=1}^N P_j \mathbb{E} \left[ \| S^{r-1,0}_j - k^{r-1}_j \|^2 \right]
    + \frac{1}{48} \mathbb{E} \left[ \| \nabla_w F ( \bar{w}_{r-1} ; S^{r,0}) \|^2 \right] + \frac{193}{192} \frac{\sigma^2}{E} \\ + \frac{1}{16} \sum_{j=1}^N P_j \ \mathbb{E} \left[ \| \nabla_w F_j ( \bar{w}_{r-1} ; \tilde{S}^{r,0}_D) - c_j^{r-1} \|^2 \right] \\
    \leq
    \frac{1}{3} J^2 M^2 \mathbb{E} \left[ \| \bar{w}_{r-1} - \bar{w}_{r-2} \|^2 \right] + \frac{1}{3} J^2 \sum_{j=1}^N P_j \mathbb{E} \left[ \| S^{r-1,0}_j - k^{r-1}_j \|^2 \right]
    + \frac{1}{48} \mathbb{E} \left[ \| \nabla_w F ( \bar{w}_{r-1} ; S^{r,0}) \|^2 \right] + \frac{193}{192} \frac{\sigma^2}{E} \\ + \frac{1}{8} L^2 \gamma^2 E \ \mathbb{E} \left[ \sum_{t=1}^E \Big\| 
    \sum_{i=1}^N P_i \ g_i ( w^{r-1,t-1}_i ; \tilde{s}^{r-1,t-1}_i ) \Big\|^2 \right] + \frac{1}{8} \sum_{j=1}^N P_j \ \mathbb{E} \left[ \| \nabla_w F_j ( \bar{w}_{r-2} ; \tilde{S}^{r-1,0}_j ) - c^{r-1}_j \|^2 \right].
\end{dmath}

By using Assumption \ref{assumption_app:l_lipschitz_continuity_statistics}, $ \mathcal{T}_3 $ ca be re-written as

\begin{dmath}
\mathcal{T}^{'}_3
= 
\sum_{j=1}^N P_j \mathbb{E} \left[ \| \nabla_w F_j ( \bar{w}_{r-1} ; S^{r,0} ) - \nabla_w F_j ( \bar{w}_{r-1} ; S_j^{r, 0} - k^{r-1}_j + k^{r-1} ) \|^2 \right] 
\\
\leq J^2 \sum_{j=1}^N P_j \mathbb{E} \left[ \| S^{r,0} - S_j^{r, 0} + k^{r-1}_j - k^{r-1} \|^2 \right] \\
\leq 4 J^2 \sum_{j=1}^N P_j \left( \mathbb{E} \left[ \| S^{r,0}_j - S^{r-1,0}_j \|^2 \right] + \mathbb{E} \left[ \| S^{r,0} - S^{r-1,0} \|^2 \right] + \mathbb{E} \left[ \| S^{r-1,0}_j - k^{r-1}_j \|^2 \right] + \mathbb{E} \left[ \| S^{r-1,0} - k^{r-1} \|^2 \right] \right) \\
\leq 4 J^2 \sum_{j=1}^N P_j \left( \mathbb{E} \left[ \| S^{r,0}_j - S^{r-1,0}_j \|^2 \right] + \sum_{i=1}^N P_i \mathbb{E} \left[ \| S^{r,0}_{D_i} - S^{r-1,0}_{D_i} \|^2 \right] + \mathbb{E} \left[ \| S^{r-1,0}_j - k^{r-1}_j \|^2 \right] + \sum_{i=1}^N P_i \mathbb{E} \left[ \| S^{r-1,0}_{D_i} - k^{r-1}_i \|^2 \right] \right) \\
\leq 8 J^2 \sum_{j=1}^N P_j \ \mathbb{E} \left[ \| S^{r,0}_j - S^{r-1,0}_j \|^2 \right] + 8 J^2 \sum_{j=1}^N P_j \ \mathbb{E} \left[ \| S^{r-1,0}_j - k^{r-1}_j \|^2 \right] \\
\leq 8 J^2 M^2 \sum_{j=1}^N P_j \ \mathbb{E} \left[ \| w^{r,0}_j - w^{r-1,0}_j \|^2 \right] + 8 J^2 \sum_{j=1}^N P_j \ \mathbb{E} \left[ \| S^{r-1,0}_j - k^{r-1}_j \|^2 \right] \\
\leq 8 J^2 M^2 \sum_{j=1}^N P_j \ \mathbb{E} \left[ \| \bar{w}_{r-1} - \bar{w}_{r-2} \|^2 \right] + 8 J^2 \sum_{j=1}^N P_j \ \mathbb{E} \left[ \| S^{r-1,0}_j - k^{r-1}_j \|^2 \right] \\
\leq 8 J^2 M^2 \mathbb{E} \left[\| \bar{w}_{r-1} - \bar{w}_{r-2} \|^2 \right] + 8 J^2 \sum_{j=1}^N P_j \ \mathbb{E} \left[ \| S^{r-1,0}_j - k^{r-1}_j \|^2 \right].
\end{dmath}

Thus, 

\begin{dmath}
\label{eq:bn_scaffold_option_2_intermediary_eq_t3}
\mathcal{T}^{'}_3
\leq \sum_{j=1}^N P_j \mathbb{E} \left[ \| \nabla_w F_j ( \bar{w}_0 ; S^{1,0} ) - \nabla_w F_j ( \bar{w}_0 ; \tilde{S}^{1,0}_j ) \|^2 \right] \\
+ 8 J^2 M^2 \sum_{r=2}^R  \mathbb{E} \left[\| \bar{w}_{r-1} - \bar{w}_{r-2} \|^2 \right] 
+ 8 J^2 \sum_{r=2}^R  \sum_{j=1}^N P_j \ \mathbb{E} \left[ \| S^{r-1,0}_j - k^{r-1}_j \|^2 \right] \\
\leq
\sum_{j=1}^N P_j \| S^{1,0}_j - k^{0}_j \|^2 + \| S^{1,0}_{D} - k^{0} \|^2
+ 8 J^2 M^2 \sum_{r=2}^R  \mathbb{E} \left[\| \bar{w}_{r-1} - \bar{w}_{r-2} \|^2 \right] 
+ 8 J^2 \sum_{r=2}^R  \sum_{j=1}^N P_j \ \mathbb{E} \left[ \| S^{r-1,0}_j - k^{r-1}_j \|^2 \right] \\
\leq
2 \sum_{j=1}^N P_j \| S^{1,0}_j - k^{0}_j \|^2
+ 8 J^2 M^2 \sum_{r=1}^R  \mathbb{E} \left[\| \bar{w}_r - \bar{w}_{r-1} \|^2 \right] 
+ 8 J^2 \sum_{r=1}^R  \sum_{j=1}^N P_j \ \mathbb{E} \left[ \| S^{r,0}_j - k^r_j \|^2 \right].
\end{dmath}

Putting Equations \eqref{eq:bn_scaffold_option_2_intermediary_eq_t4} and \eqref{eq:bn_scaffold_option_2_intermediary_eq_t3} together, we have

\begin{dmath}
    \label{eq:bn_scaffold_option_2_intermediary_eq_4}
    \mathcal{T}_3 + \mathcal{T}_4 = \frac{2}{R} (1 + \delta_{E, \gamma, L}) \sum_{r=1}^R \sum_{j=1}^N P_j \mathbb{E} \left[ \| \nabla_w F_j ( \bar{w}_{r-1} ; S^{r,0} ) - \nabla_w F_j ( \bar{w}_{r-1} ; \tilde{S}^{r,0}_j ) \|^2 \right] \\
    + 16 \ \delta_{E, \gamma, L} \sum_{j=1}^N P_j \mathbb{E} \left[ \| \nabla_w F_j ( \bar{w}_{r -1} ; \tilde{S}^{r, 0}_j ) - c^r_j \|^2 \right] \\
    \leq
    \frac{4}{R} (1 + \delta_{E, \gamma, L}) \sum_{j=1}^N P_j \| S^{1,0}_j - k^{0}_j \|^2
    + \frac{16}{R} J^2 M^2  (1 + \delta_{E, \gamma, L}) \sum_{r=1}^R  \mathbb{E} \left[\| \bar{w}_r - \bar{w}_{r-1} \|^2 \right] \\
    + \frac{16}{R} \sum_{r=1}^R \sum_{j=1}^N P_j  \left( J^2  (1 + \delta_{E, \gamma, L}) \mathbb{E} \left[ \| S^{r,0}_j - k^r_j \|^2 \right] + \delta_{E, \gamma, L} \mathbb{E} \left[ \| \nabla_w F_j ( \bar{w}_{r -1} ; \tilde{S}^{r, 0}_j ) - c^r_j \|^2 \right] \right).
\end{dmath}

We now analyze the last term of Equation \eqref{eq:bn_scaffold_option_2_intermediary_eq_4}

\begin{dmath}
J^2  (1 + \delta_{E, \gamma, L}) \sum_{j=1}^N P_j \mathbb{E} \left[ \| S^{r,0}_j - k^r_j \|^2 \right] + \delta_{E, \gamma, L} \sum_{j=1}^N P_j \mathbb{E} \left[ \| \nabla_w F_j ( \bar{w}_{r -1} ; \tilde{S}^{r, 0}_j ) - c^r_j \|^2 \right] \\
\leq
%
\frac{1}{24} \left[ \frac{1}{2} \delta_{E, \gamma, L} + \frac{M^2 J^2}{L^2}  (1 + \delta_{E, \gamma, L}) \right] \mathbb{E} \left[ \| \nabla_w F ( \bar{w}_{r-1} ; S^{r,0}) \|^2 \right]
+ \frac{1}{96} \left[ \frac{193}{2} \delta_{E, \gamma, L} + \frac{M^2 J^2}{L^2} (1 + \delta_{E, \gamma, L}) \right] \frac{\sigma^2}{E} \\
+ \gamma^2 E \left[ \frac{1}{8} L^2 \delta_{E, \gamma, L} + M^2 J^2  (1 + \delta_{E, \gamma, L}) \left( \frac{1}{4} + \frac{2}{3 L^2} J^2 M^2 \right) + \frac{J^2 M^2}{3} \delta_{E, \gamma, L} \right] \mathbb{E} \left[ \sum_{t=1}^E \Big\| 
\sum_{i=1}^N P_i \ g_i ( w^{r-1,t-1}_i ; \tilde{s}^{r-1,t-1}_i ) \Big\|^2 \right] \\
+ 2 J^2  (1 + \delta_{E, \gamma, L}) \sigma^2_s
+ \frac{1}{3} \left[ \frac{\delta_{E, \gamma, L} }{(1 + \delta_{E, \gamma, L})}
+ 2 \frac{M^2 J^2}{L^2} \right] (1 + \delta_{E, \gamma, L}) J^2 \sum_{j=1}^N P_j \mathbb{E} \left[ \| S^{r-1,0}_j - k^{r-1}_j \|^2 \right] \\
+ \frac{1}{8} \left[ 1 + 2 \frac{M^2 J^2}{L^2} \frac{1 + \delta_{E, \gamma, L} }{\delta_{E, \gamma, L}} \right] \delta_{E, \gamma, L} \sum_{j=1}^N P_j \ \mathbb{E} \left[ \| \nabla_w F_j ( \bar{w}_{r-2} ; \tilde{S}^{r-1,0}_j ) - c^{r-1}_j \|^2 \right].
\end{dmath}

If $   \delta_{E, \gamma, L} \leq 1 $, we have that $ \frac{\delta_{E, \gamma, L} }{1 + \delta_{E, \gamma, L}}
+ 2\frac{M^2 J^2}{L^2} \leq 1
+ 2 \frac{M^2 J^2}{L^2} \frac{ 1 + \delta_{E, \gamma, L}}{\delta_{E, \gamma, L}} $, and, as $ \frac{1}{8} < \frac{1}{3} $, we obtain the following recursion

\begin{dmath}
\underbrace{J^2  (1 + \delta_{E, \gamma, L}) \sum_{j=1}^N P_j \mathbb{E} \left[ \| S^{r,0}_j - k^r_j \|^2 \right] + \delta_{E, \gamma, L} \sum_{j=1}^N P_j \mathbb{E} \left[ \| \nabla_w F_j ( \bar{w}_{r -1} ; \tilde{S}^{r, 0}_j ) - c^r_j \|^2 \right]}_{u_r} \\
\leq
%
\frac{1}{24} \left[ \frac{1}{2} \delta_{E, \gamma, L} + \frac{M^2 J^2}{L^2}  (1 + \delta_{E, \gamma, L}) \right] \mathbb{E} \left[ \| \nabla_w F ( \bar{w}_{r-1} ; S^{r,0}) \|^2 \right]
+ \frac{1}{96} \left[ \frac{193}{2} \delta_{E, \gamma, L} + \frac{M^2 J^2}{L^2} (1 + \delta_{E, \gamma, L}) \right] \frac{\sigma^2}{E} + 2 J^2  (1 + \delta_{E, \gamma, L}) \sigma^2_s \\
+ \gamma^2 E \left[ \frac{1}{8} L^2 \delta_{E, \gamma, L} + M^2 J^2  (1 + \delta_{E, \gamma, L}) \left( \frac{1}{4} + \frac{2}{3 L^2} J^2 M^2 \right) + \frac{J^2 M^2}{3} \delta_{E, \gamma, L} \right] \mathbb{E} \left[ \sum_{t=1}^E \Big\| 
\sum_{i=1}^N P_i \ g_i ( w^{r-1,t-1}_i ; \tilde{s}^{r-1,t-1}_i ) \Big\|^2 \right] \\
+ \underbrace{\frac{1}{3} \left[1 + 2 \frac{M^2 J^2}{L^2} \frac{1 + \delta_{E, \gamma, L} }{\delta_{E, \gamma, L}} \right]}_{a} \\ \underbrace{\left[  (1 + \delta_{E, \gamma, L}) J^2 \sum_{j=1}^N P_j \mathbb{E} \left[ \| S^{r-1,0}_j - k^{r-1}_j \|^2 \right] + \delta_{E, \gamma, L} \sum_{j=1}^N P_j \ \mathbb{E} \left[ \| \nabla_w F_j ( \bar{w}_{r-2} ; \tilde{S}^{r-1,0}_j ) - c^{r-1}_j \|^2 \right] \right]}_{u_{r-1}}.
\end{dmath}

Assuming $ a = \frac{1}{3} \left[1 + 2 \frac{M^2 J^2}{L^2} \frac{1 + \delta_{E, \gamma, L} }{\delta_{E, \gamma, L}} \right] < 1 \leftrightarrow \frac{M^2 J^2}{L^2} \frac{1 + \delta_{E, \gamma, L} }{\delta_{E, \gamma, L}} < 1 $ and using Lemma \ref{lemma:recursion} yields

\begin{dmath}
\label{eq:bn_scaffold_option_2_intermediary_eq_5}
\frac{1}{R} \sum_{r=1}^R J^2  (1 + \delta_{E, \gamma, L}) \sum_{j=1}^N P_j \mathbb{E} \left[ \| S^{r,0}_j - k^r_j \|^2 \right] \\ + \frac{1}{R} \sum_{r=1}^R \delta_{E, \gamma, L} \sum_{j=1}^N P_j \mathbb{E} \left[ \| \nabla_w F_j ( \bar{w}_{r -1} ; \tilde{S}^{r, 0}_j ) - c^r_j \|^2 \right] \\
\leq \frac{3}{2} \frac{1}{1 - \frac{M^2 J^2}{L^2} \frac{1 + \delta_{E, \gamma, L} }{\delta_{E, \gamma, L}} } \left[ \frac{1}{R} J^2  (1 + \delta_{E, \gamma, L}) \sum_{j=1}^N P_j \mathbb{E} \left[ \| S^{1,0}_j - k^1_j \|^2 \right] + \frac{1}{R} \delta_{E, \gamma, L} \sum_{j=1}^N P_j \mathbb{E} \left[ \| \nabla_w F_j ( \bar{w}_{0} ; \tilde{S}^{1, 0}_j ) - c^1_j \|^2 \right] \\
+ \frac{1}{R} \frac{1}{24} \left[ \frac{1}{2} \delta_{E, \gamma, L} + \frac{M^2 J^2}{L^2}  (1 + \delta_{E, \gamma, L}) \right] \sum_{r=2}^R \mathbb{E} \left[ \| \nabla_w F ( \bar{w}_{r-1} ; S^{r,0}) \|^2 \right] \\
+ \frac{1}{96} \frac{R-1}{R} \left[ \frac{193}{2} \delta_{E, \gamma, L} + \frac{M^2 J^2}{L^2} (1 + \delta_{E, \gamma, L}) \right] \frac{\sigma^2}{E} + 2 \frac{R-1}{R} J^2  (1 + \delta_{E, \gamma, L}) \sigma^2_s \\
+ \frac{1}{R} \gamma^2 E \left[ \frac{1}{8} L^2 \delta_{E, \gamma, L} + M^2 J^2  (1 + \delta_{E, \gamma, L}) \left( \frac{1}{4} + \frac{2}{3 L^2} J^2 M^2 \right) + \frac{J^2 M^2}{3} \delta_{E, \gamma, L} \right] 
\sum_{r=1}^R \mathbb{E} \left[ \sum_{t=1}^E \Big\| 
\sum_{i=1}^N P_i \ g_i ( w^{r,t-1}_i ; \tilde{s}^{r,t-1}_i ) \Big\|^2 \right]
\right].
\end{dmath}

By using Lemma \ref{lemma:bn_scaffold_option_2_drift_in_estimated_stats} with $ r = 1 $

\begin{dmath}
\label{eq:bn_scaffold_option_2_intermediary_eq_6}
\frac{1}{R} \sum_{r=1}^R J^2  (1 + \delta_{E, \gamma, L}) \sum_{j=1}^N P_j \mathbb{E} \left[ \| S^{r,0}_j - k^r_j \|^2 \right] \\ + \frac{1}{R} \sum_{r=1}^R \delta_{E, \gamma, L} \sum_{j=1}^N P_j \mathbb{E} \left[ \| \nabla_w F_j ( \bar{w}_{r -1} ; \tilde{S}^{r, 0}_j ) - c^r_j \|^2 \right] \\
\leq \frac{3}{2} \frac{1}{1 - \frac{M^2 J^2}{L^2} \frac{1 + \delta_{E, \gamma, L} }{\delta_{E, \gamma, L}} } \left[
\frac{1}{R} J^2  (1 + \delta_{E, \gamma, L}) \frac{1}{3} \frac{M^2 J^2}{L^2} \sum_{j=1}^N P_j \ \mathbb{E} \left[ \| S^{1,0}_j - k^0_j  \|^2 \right] \\
\frac{3}{24} \frac{1}{R} (1 + \delta_{E, \gamma, L}) \frac{M^2 J^2}{L^2} \sum_{j=1}^N P_j \ \mathbb{E} \left[ \| c^0_j - \nabla_w F_j ( \bar{w}_0 ; \tilde{S}^{1,0}_j ) \|^2 \right] \\
+ \frac{1}{R} \delta_{E, \gamma, L} \sum_{j=1}^N P_j \mathbb{E} \left[ \| \nabla_w F_j ( \bar{w}_{0} ; \tilde{S}^{1, 0}_j ) - c^1_j \|^2 \right] \\
+ \frac{1}{R} \frac{1}{24} \frac{1}{2} \delta_{E, \gamma, L} \sum_{r=2}^R \mathbb{E} \left[ \| \nabla_w F ( \bar{w}_{r-1} ; S^{r,0}) \|^2 \right] \\
+ \frac{1}{R} \frac{1}{24} \frac{M^2 J^2}{L^2}  (1 + \delta_{E, \gamma, L}) \sum_{r=1}^R \mathbb{E} \left[ \| \nabla_w F ( \bar{w}_{r-1} ; S^{r,0}) \|^2 \right] \\
+ \frac{1}{96} \left[ \frac{R-1}{R} \frac{193}{2} \delta_{E, \gamma, L} + \frac{M^2 J^2}{L^2} (1 + \delta_{E, \gamma, L}) \right] \frac{\sigma^2}{E} + 2 J^2  (1 + \delta_{E, \gamma, L}) \sigma^2_s \\
+ \frac{1}{R} \gamma^2 E \left[ \frac{1}{8} L^2 \delta_{E, \gamma, L} + M^2 J^2  (1 + \delta_{E, \gamma, L}) \left( \frac{1}{4} + \frac{2}{3 L^2} J^2 M^2 \right) + \frac{J^2 M^2}{3} \delta_{E, \gamma, L} \right] 
\sum_{r=1}^R \mathbb{E} \left[ \sum_{t=1}^E \Big\| 
\sum_{i=1}^N P_i \ g_i ( w^{r,t-1}_i ; \tilde{s}^{r,t-1}_i ) \Big\|^2 \right]
\right].
\end{dmath}

Using Equation \eqref{eq:bn_scaffold_option_2_intermediary_eq_2} and Lemma \ref{lemma:bn_scaffold_global_grad_global_control_variate} with $r=1$ we have

\begin{dmath}
    \sum_{j=1}^N P_j \mathbb{E} \left[ \| \nabla_w F_j ( \bar{w}_{0} ; \tilde{S}^{1, 0}_j ) - c^1_j \|^2 \right] \\
    \leq
    \frac{193}{192} \frac{\sigma^2}{E} 
    + \frac{1}{6} J^2 \sum_{j=1}^N P_j \ \mathbb{E} \left[ \| S^{1,0}_j - k^0_j  \|^2 \right]
    + \frac{3}{48} \sum_{j=1}^N P_j \ \mathbb{E} \left[ \| \nabla_w F_j ( \bar{w}_{0} ; \tilde{S}^{1,0}_j) - c^{0}_j \|^2 \right]
    + \frac{1}{48} \mathbb{E} \left[ \| \nabla_w F ( \bar{w}_{0} ; S^{1,0}) \|^2 \right],
\end{dmath}

which can be plugged in Equation \eqref{eq:bn_scaffold_option_2_intermediary_eq_6} to yield

\begin{dmath}
\label{eq:bn_scaffold_option_2_intermediary_eq_7b}
\frac{1}{R} \sum_{r=1}^R J^2  (1 + \delta_{E, \gamma, L}) \sum_{j=1}^N P_j \mathbb{E} \left[ \| S^{r,0}_j - k^r_j \|^2 \right] \\ + \frac{1}{R} \sum_{r=1}^R \delta_{E, \gamma, L} \sum_{j=1}^N P_j \mathbb{E} \left[ \| \nabla_w F_j ( \bar{w}_{r -1} ; \tilde{S}^{r, 0}_j ) - c^r_j \|^2 \right] \\
\leq \frac{3}{2} \frac{1}{1 - \frac{M^2 J^2}{L^2} \frac{1 + \delta_{E, \gamma, L} }{\delta_{E, \gamma, L}} } \left[
\frac{1}{R} \frac{J^2}{3} \left[ \frac{1}{2} \delta_{E, \gamma, L} + (1 + \delta_{E, \gamma, L}) \frac{M^2 J^2}{L^2} \right] \sum_{j=1}^N P_j \ \mathbb{E} \left[ \| S^{1,0}_j - k^0_j  \|^2 \right] \\
\frac{1}{R} \frac{3}{24} \left[ \frac{1}{2} \delta_{E, \gamma, L} + (1 + \delta_{E, \gamma, L}) \frac{M^2 J^2}{L^2} \right] \sum_{j=1}^N P_j \ \mathbb{E} \left[ \| c^0_j - \nabla_w F_j ( \bar{w}_0 ; \tilde{S}^{1,0}_j ) \|^2 \right] \\
+ \frac{1}{R} \frac{1}{24} \left( \frac{1}{2} \delta_{E, \gamma, L} + (1 + \delta_{E, \gamma, L}) \frac{M^2 J^2}{L^2} \right) \sum_{r=1}^R \mathbb{E} \left[ \| \nabla_w F ( \bar{w}_{r-1} ; S^{r,0}) \|^2 \right] \\
+ \frac{193}{96} \left[ \frac{1}{2} \delta_{E, \gamma, L} + (1 + \delta_{E, \gamma, L}) \frac{M^2 J^2}{L^2} \right] \frac{\sigma^2}{E} + 2 J^2  (1 + \delta_{E, \gamma, L}) \sigma^2_s \\
+ \frac{1}{R} \gamma^2 E \left[ \frac{1}{8} L^2 \delta_{E, \gamma, L} + M^2 J^2  (1 + \delta_{E, \gamma, L}) \left( \frac{1}{4} + \frac{2}{3 L^2} J^2 M^2 \right) + \frac{J^2 M^2}{3} \delta_{E, \gamma, L} \right] 
\sum_{r=1}^R \mathbb{E} \left[ \sum_{t=1}^E \Big\| 
\sum_{i=1}^N P_i \ g_i ( w^{r,t-1}_i ; \tilde{s}^{r,t-1}_i ) \Big\|^2 \right]
\right].
\end{dmath}

By setting $ \alpha_{E, \gamma, L, M, J} = \frac{M^2 J^2}{L^2} \frac{1 + \delta_{E, \gamma, L} }{\delta_{E, \gamma, L}} $, we have

\begin{dmath}
\label{eq:bn_scaffold_option_2_intermediary_eq_7a}
\frac{1}{R} \sum_{r=1}^R J^2  (1 + \delta_{E, \gamma, L}) \sum_{j=1}^N P_j \mathbb{E} \left[ \| S^{r,0}_j - k^r_j \|^2 \right] \\ + \frac{1}{R} \sum_{r=1}^R \delta_{E, \gamma, L} \sum_{j=1}^N P_j \mathbb{E} \left[ \| \nabla_w F_j ( \bar{w}_{r -1} ; \tilde{S}^{r, 0}_j ) - c^r_j \|^2 \right] \\
\leq \frac{3}{2} \frac{1}{1 - \alpha_{E, \gamma, L, M, J} } \left[
\frac{1}{R} \frac{2}{3} J^2 \delta_{E, \gamma, L} \left( 1 + 2 \alpha_{E, \gamma, L, M, J} \right) \sum_{j=1}^N P_j \ \mathbb{E} \left[ \| S^{1,0}_j - k^0_j  \|^2 \right] \\
\frac{1}{R} \frac{1}{4} \delta_{E, \gamma, L} \left( 1 + 2 \alpha_{E, \gamma, L, M, J} \right)  \sum_{j=1}^N P_j \ \mathbb{E} \left[ \| c^0_j - \nabla_w F_j ( \bar{w}_0 ; \tilde{S}^{1,0}_j ) \|^2 \right] \\
+ \frac{1}{R} \frac{1}{12} \delta_{E, \gamma, L} \left( 1 + 2 \alpha_{E, \gamma, L, M, J} \right)  \sum_{r=1}^R \mathbb{E} \left[ \| \nabla_w F ( \bar{w}_{r-1} ; S^{r,0}) \|^2 \right] \\
+ \frac{193}{48} \delta_{E, \gamma, L} \left( 1 + 2 \alpha_{E, \gamma, L, M, J} \right) \frac{\sigma^2}{E} + 2 J^2  (1 + \delta_{E, \gamma, L}) \sigma^2_s \\
+ \frac{1}{R} \gamma^2 E \left[ \frac{1}{8} L^2 \delta_{E, \gamma, L} + M^2 J^2  (1 + \delta_{E, \gamma, L}) \left( \frac{1}{4} + \frac{2}{3 L^2} J^2 M^2 \right) + \frac{J^2 M^2}{3} \delta_{E, \gamma, L} \right] 
\sum_{r=1}^R \mathbb{E} \left[ \sum_{t=1}^E \Big\| 
\sum_{i=1}^N P_i \ g_i ( w^{r,t-1}_i ; \tilde{s}^{r,t-1}_i ) \Big\|^2 \right]
\right].
\end{dmath}

Combining Equations \eqref{eq:bn_scaffold_option_2_intermediary_eq_4} and \eqref{eq:bn_scaffold_option_2_intermediary_eq_7b} yields

\begin{dmath}
\label{eq:bn_scaffold_option_2_intermediary_eq_8}
\mathcal{T}_3 + \mathcal{T}_4 = \frac{2}{R} (1 + \delta_{E, \gamma, L}) \sum_{r=1}^R \sum_{j=1}^N P_j \mathbb{E} \left[ \| \nabla_w F_j ( \bar{w}_{r-1} ; S^{r,0} ) - \nabla_w F_j ( \bar{w}_{r-1} ; \tilde{S}^{r,0}_j ) \|^2 \right] \\
+ 16 \ \delta_{E, \gamma, L} \sum_{j=1}^N P_j \mathbb{E} \left[ \| \nabla_w F_j ( \bar{w}_{r -1} ; \tilde{S}^{r, 0}_j ) - c^r_j \|^2 \right] \\
\leq
\left[ 4 \frac{J^2 \delta_{E, \gamma, L} ( 1 + 2 \alpha_{E, \gamma, L, M, J}) }{1 - \alpha_{E, \gamma, L, M, J} } + (1 + \delta_{E, \gamma, L}) \right] \frac{1}{R} \sum_{j=1}^N P_j \ \mathbb{E} \left[ \| S^{1,0}_j - k^0_j  \|^2 \right] \\
6 \frac{\delta_{E, \gamma, L} ( 1 + 2 \alpha_{E, \gamma, L, M, J} ) }{1 - \alpha_{E, \gamma, L, M, J} } \frac{1}{R} \sum_{j=1}^N P_j \ \mathbb{E} \left[ \| c^0_j - \nabla_w F_j ( \bar{w}_0 ; \tilde{S}^{1,0}_j ) \|^2 \right] \\
+ 2 \frac{\delta_{E, \gamma, L} ( 1 + 2 \alpha_{E, \gamma, L, M, J} ) }{1 - \alpha_{E, \gamma, L, M, J} } \frac{1}{R} \sum_{r=1}^R \mathbb{E} \left[ \| \nabla_w F ( \bar{w}_{r-1} ; S^{r,0}) \|^2 \right] \\
+ \frac{193}{2} \frac{\delta_{E, \gamma, L} ( 1 + 2 \alpha_{E, \gamma, L, M, J} ) }{1 - \alpha_{E, \gamma, L, M, J} } \frac{\sigma^2}{E} + 2 J^2  (1 + \delta_{E, \gamma, L}) \sigma^2_s \\
+ \gamma^2 E^2 \left[ \frac{24}{1 - \alpha_{E, \gamma, L, M, J} } \left( \frac{1}{8} L^2 \delta_{E, \gamma, L} + M^2 J^2  (1 + \delta_{E, \gamma, L}) \left( \frac{1}{4} + \frac{2}{3 L^2} J^2 M^2 \right) + \frac{J^2 M^2}{3} \delta_{E, \gamma, L} \right) + 16 J^2 M^2  (1 + \delta_{E, \gamma, L}) \right] \\
\frac{1}{R E} \sum_{r=1}^R \mathbb{E} \left[ \sum_{t=1}^E \Big\| 
\sum_{i=1}^N P_i \ g_i ( w^{r,t-1}_i ; \tilde{s}^{r,t-1}_i ) \Big\|^2
\right].
\end{dmath}

Thus, using Equation \eqref{eq:bn_scaffold_option_2_intermediary_eq_8} in Theorem \ref{theorem:variance_reduc_algos}, gives 

\begin{dmath}
\frac{1}{R} ( 1 - 2 \delta_{E, \gamma, L} ) \sum_{r=1}^R \mathbb{E} \left[ \| \nabla_w F_i ( \bar{w}_{r-1}; S^{r,0} ) \|^2 \right] 
\leq \\
\frac{2}{\gamma R E} \left[ F( \bar{w}_0 ; S^{1,0} ) - \underbar{F} \right] \\
%
+ \left[ 4 \frac{J^2 \delta_{E, \gamma, L} ( 1 + 2 \alpha_{E, \gamma, L, M, J}) }{1 - \alpha_{E, \gamma, L, M, J} } + (1 + \delta_{E, \gamma, L}) \right] \frac{1}{R} \sum_{j=1}^N P_j \ \mathbb{E} \left[ \| S^{1,0}_j - k^0_j  \|^2 \right] \\
+ 2 \delta_{E, \gamma, L} \left[ 3 \frac{ ( 1 + 2 \alpha_{E, \gamma, L, M, J} ) }{1 - \alpha_{E, \gamma, L, M, J} } + 4 \right] \frac{1}{R} \sum_{j=1}^N P_j \ \mathbb{E} \left[ \| c^0_j - \nabla_w F_j ( \bar{w}_0 ; \tilde{S}^{1,0}_j ) \|^2 \right] \\
+ 2 \frac{\delta_{E, \gamma, L} ( 1 + 2 \alpha_{E, \gamma, L, M, J} ) }{1 - \alpha_{E, \gamma, L, M, J} } \frac{1}{R} \sum_{r=1}^R \mathbb{E} \left[ \| \nabla_w F ( \bar{w}_{r-1} ; S^{r,0}) \|^2 \right] \\
+ \left[ \frac{1}{E} \frac{193}{2} \frac{\delta_{E, \gamma, L} ( 1 + 2 \alpha_{E, \gamma, L, M, J} ) }{1 - \alpha_{E, \gamma, L, M, J} } + ( 1 + 2 \delta_{E, \gamma, L} ) \right] \sigma^2 + 2 J^2  (1 + \delta_{E, \gamma, L}) \sigma^2_s \\
- \left[ (1 - \gamma L E - 16 L^2 E^2 \gamma^2 \delta_{E, \gamma, L} - \gamma^2 E^2 \left[ \frac{24}{1 - \alpha_{E, \gamma, L, M, J} } \left( \frac{1}{8} L^2 \delta_{E, \gamma, L} + M^2 J^2  (1 + \delta_{E, \gamma, L}) \\ \left( \frac{1}{4} + \frac{2}{3 L^2} J^2 M^2 \right) + \frac{J^2 M^2}{3} \delta_{E, \gamma, L} \right) + 16 J^2 M^2  (1 + \delta_{E, \gamma, L}) \right] \right]
\frac{1}{R E} \sum_{r=1}^R \mathbb{E} \left[ \sum_{t=1}^E \Big\| 
\sum_{i=1}^N P_i \ g_i ( w^{r,t-1}_i ; \tilde{s}^{r,t-1}_i ) \Big\|^2 \right].
\end{dmath}

By re-organizing and setting $ \gamma $ such that

\begin{dmath}
1 - \gamma L E - 16 L^2 E^2 \gamma^2 \delta_{E, \gamma, L} - \gamma^2 E^2 \left[ \frac{24}{1 - \alpha_{E, \gamma, L, M, J} } \left( \frac{1}{8} L^2 \delta_{E, \gamma, L} + M^2 J^2  (1 + \delta_{E, \gamma, L}) \\ \left( \frac{1}{4} + \frac{2}{3 L^2} J^2 M^2 \right) + \frac{J^2 M^2}{3} \delta_{E, \gamma, L} \right) + 16 J^2 M^2  (1 + \delta_{E, \gamma, L}) \right] \geq 0,
\end{dmath}

we obtain

\begin{dmath}
\frac{1}{R} \left[ 1 - 2 \delta_{E, \gamma, L} \left( 1 +  \frac{ 1 + 2 \alpha_{E, \gamma, L, M, J} }{1 - \alpha_{E, \gamma, L, M, J} } \right) \right] \sum_{r=1}^R \mathbb{E} \left[ \| \nabla_w F_i ( \bar{w}_{r-1}; S^{r,0} ) \|^2 \right] 
\leq \\
\frac{2}{\gamma R E} \left[ F_0 - \underbar{F} \right] \\
%
+ \left[ 4 J^2 \delta_{E, \gamma, L} \frac{ 1 + 2 \alpha_{E, \gamma, L, M, J} }{1 - \alpha_{E, \gamma, L, M, J} } + (1 + \delta_{E, \gamma, L}) \right] \frac{1}{R} \Delta^2 k^0 \\
+ 2 \delta_{E, \gamma, L} \left[ 3 \frac{ 1 + 2 \alpha_{E, \gamma, L, M, J} }{1 - \alpha_{E, \gamma, L, M, J} } + 4 \right] \frac{1}{R} \Delta^2 c^0 \\
+ \left[ \frac{1}{E} \frac{193}{2} \delta_{E, \gamma, L} \frac{ 1 + 2 \alpha_{E, \gamma, L, M, J} }{1 - \alpha_{E, \gamma, L, M, J} } + ( 1 + 2 \delta_{E, \gamma, L} ) \right] \sigma^2 + 2 J^2  (1 + \delta_{E, \gamma, L}) \sigma^2_s,
\end{dmath}

where $ F( \bar{w}_0 ; S^{1,0}) = F_0 $, $ \Delta^2 c^0 = \sum_{j=1}^N P_j \| c^0_j - \nabla_w F_j ( \bar{w}_0 ; \tilde{S}^{1,0}_j ) \|^2  $, and 

$ \Delta^2 k^0 = \sum_{j=1}^N P_j \| S^{1,0}_j - k^0_j  \|^2 $. For the inequality to be valid, the three following conditions need to be met:

\begin{dmath}
    \begin{cases}
    \alpha_{E, \gamma, L, M, J} = \frac{M^2 J^2}{L^2} \frac{\delta_{E, \gamma, L} + 1}{\delta_{E, \gamma, L}} = \frac{M^2 J^2}{L^2} \frac{1 - 4 E^2 \gamma^2 L^2}{4
    E^2 \gamma^2 L^2} \leq 1 \\[10pt]
    1 - \gamma L E - 16 L^2 E^2 \gamma^2 \delta_{E, \gamma, L} 
    1 - \gamma L E - 16 L^2 E^2 \gamma^2 \delta_{E, \gamma, L} - \gamma^2 E^2 \Big[ \frac{24}{1 - \alpha_{E, \gamma, L, M, J} } \Big( \frac{1}{8} L^2 \delta_{E, \gamma, L} + M^2 J^2  (1 + \delta_{E, \gamma, L} ) \\ \left( \frac{1}{4} + \frac{2}{3 L^2} J^2 M^2 \right) + \frac{J^2 M^2}{3} \delta_{E, \gamma, L} \Big) + 16 J^2 M^2  (1 + \delta_{E, \gamma, L}) \Big] \geq 0.
\end{cases}
\end{dmath}

The first two conditions impose $ \frac{M J }{2 \sqrt{L^2 + M^2 J^2} L E} \leq \gamma \leq \frac{1}{2 L E}$ and $ \gamma < \frac{1}{4 L E}$, with $ L^2 > 3 M^2 J^2 $. The last condition can be re-written as follows

\begin{dmath}
    1 - \gamma L E - L^{*2} \gamma^2 E^2 \delta_{E, \gamma, L} - 16 J^2 M^2 \gamma^2 E^2 \geq 0,
\end{dmath}

with 

\begin{equation}
    L^{*2} \coloneq 16 L^2 + \left( 16 + \frac{24}{1 - \alpha_{E, \gamma, L, M, J}} \right) M^2 J^2 + \frac{3L^4+32 M^4 J^4 + 20 M^2 J^2 L^2}{ (1 - \alpha_{E, \gamma, L, M, J}) L^2}.
\end{equation}

This gives three conditions:

\begin{equation}
    \begin{cases}
        1/3 - \gamma L E \geq 0 \\
        1/3 - L^{*2} \gamma^2 E^2 \delta_{E, \gamma, L} \geq 0 \\
        1/3 - 16 J^2 M^2 \gamma^2 E^2 \geq 0,
    \end{cases}
\end{equation}

which impose the following conditions in the learning rate

\begin{equation}
    \begin{cases}
        \gamma \leq \frac{1}{3 L E} \\
        \gamma \leq \sqrt{\frac{-8 L^2 + \sqrt{64 L^4 + 48 L^{*2} L^2}}{24 L^{*2} L^2}} \frac{1}{E} \\
        \gamma \leq \frac{1}{\sqrt{48}J M E},
    \end{cases}
\end{equation}

with the additional implicit condition on $ L$

\begin{equation}
    48 L^6 \geq 35 L^{*2} M^4 J^4 + 48 M^2 J^2 L^2 (L^2 + J^2 M^2).
\end{equation}

We repeat the above methodology by imposing:

\begin{equation}
    \begin{cases}
        24 L^6 \geq 35 L^{*2} M^4 J^4 \\
        24 L^6 \geq 48 M^2 J^2 L^2 ( L^6 + J^2 M^2),
    \end{cases}
\end{equation}

where the second condition is verified if $L^2 \geq 3 M^2 J^2$. It can be shown that the first condition can be re-written as

\begin{dmath}
    \label{eq:temp_verifs_1}
    \left( \frac{M^2 J^2}{L^2} \right)^4 (1120 - 560 \frac{\delta_{E, \gamma, L}+1}{\delta_{E, \gamma, L}}) + \left( \frac{M^2 J^2}{L^2} \right)^3 (2100 - 560 \frac{\delta_{E, \gamma, L}+1}{\delta_{E, \gamma, L}}) + 665 \left( \frac{M^2 J^2}{L^2} \right)^2 \\ + 24 \frac{M^2 J^2}{L^2} \frac{\delta_{E, \gamma, L}+1}{\delta_{E, \gamma, L}} \\
    \leq \left( \frac{M^2 J^2}{L^2} \right)^4 1120 + \left( \frac{M^2 J^2}{L^2} \right)^3 2100 + 665 \left( \frac{M^2 J^2}{L^2} \right)^2 + 24 \frac{M^2 J^2}{L^2} \frac{\delta_{E, \gamma, L}+1}{\delta_{E, \gamma, L}} \\
    \leq 24.
\end{dmath}

We decompose Equation \eqref{eq:temp_verifs_1} into the following conditions

\begin{equation}
    \begin{cases}
        1120 \left( \frac{M^2 J^2}{L^2} \right)^4 \leq 96 \\
        2100 \left( \frac{M^2 J^2}{L^2} \right)^3 \leq 96 \\
        665 \left( \frac{M^2 J^2}{L^2} \right)^2 \leq 96 \\
        24 \frac{M^2 J^2}{L^2} \frac{\delta_{E, \gamma, L}+1}{\delta_{E, \gamma, L}}\leq 96
    \end{cases},
\end{equation}

which yields

\begin{equation}
    L^2 \geq \max \left\{ \underbrace{\sqrt[4]{\frac{1120}{96}}}_{\approx 1.84}; \underbrace{\sqrt[3]{\frac{2100}{96}}}_{\approx 2.79}; \underbrace{\sqrt{\frac{665}{96}}}_{\approx 2.63} \right\} M^2 J^2,
\end{equation}

which is already verified as we assumed $ L^2 > 3 M^2 J^2 $, and

\begin{equation}
    \gamma \geq \frac{M J}{2 \sqrt{(2L)^2 + M^2 J^2}} \frac{1}{LE},
\end{equation}

which is also already verified as we supposed $ \gamma \geq \frac{M J}{2 \sqrt{L^2 + M^2 J^2} L E} $. We finally have, by recalling the previous condition $ \gamma \leq \frac{1}{24 L E} $, the following set of conditions for convergence:

\begin{equation}
    \begin{cases}        
    \gamma \leq \min \left\{ \frac{1}{24 L}; \sqrt{\frac{-8 L^2 + \sqrt{64 L^4 + 48 L^{*2} L^2}}{24 L^{*2} L^2}}; \frac{1}{\sqrt{48}J M} \right\} \frac{1}{E} \\[10pt]
    \gamma \geq \frac{M J}{2 \sqrt{L^2 + M^2 J^2} L E} \\[10pt]
    M^2 J^2 \leq L^2 / 3.
    \end{cases}
\end{equation}

\end{proof}

\begin{corollary}

By taking the constant learning rate $ \gamma = \frac{\gamma_0}{L E } $, we obtain

\begin{dmath}
    \min_r \mathbb{E} \left[ \| \nabla_w F ( \bar{w}_{r-1} ; S^{r,0} ) \|^2 \right] \\
    \leq 
    \frac{1}{R} \sum_{r=1}^R \mathbb{E} \left[ \| \nabla_w F ( \bar{w}_{r-1} ; S^{r,0} ) \|^2 \right] \\
    \leq \mathcal{O} \left( \frac{L \ \Omega}{R} \left[ F_0 - \underbar{F} \right] + \frac{J^2 + \Omega}{R} \Delta^2 k^0 + \frac{1 + \Omega}{R} \Delta^2 c^0 + (\frac{1}{E} + \Omega ) \sigma^2 + J^2 \Omega \ \sigma^2_s \right)
\end{dmath},

where $ \Omega \coloneq \frac{L^2 - M^2 J^2}{L^2 + M^2 J^2}$ measures the difference between the Lipschitz constants with respect to the gradients and with respect to the BN statistics, in a relative manner.

\end{corollary}

\begin{proof}

We have

\begin{equation}
\begin{cases}
        \delta_{E, \gamma, L} = \frac{4 E^2 \gamma^2 L^2 }{1 - 8 E^2 \gamma^2 L^2} = \frac{4 \gamma^2_0}{1 - 8 \gamma^2_0} = \mathcal{O} (1) \\[10pt]
        \alpha_{E, \gamma, L, M, J} = \mathcal{O} ( \frac{M^2 J^2}{L^2} ) \\[10pt]
        \frac{1 + \alpha_{E, \gamma, L, M, J}}{1 - \alpha_{E, \gamma, L, M, J}} = \mathcal{O} \left( \frac{L^2 + M^2 J^2}{L^2 - M^2 J^2} \right).
\end{cases}
\end{equation}

Applying this to Equation \eqref{eq:bn_scaffold_option2_convergence} and defining $ \Omega \coloneq \frac{L^2 - M^2 J^2}{L^2 + M^2 J^2}$ proofs the corollary.

\end{proof}

\begin{lemma}[BN-SCAFFOLD - Global gradient and global control variate drift]
\label{lemma:bn_scaffold_global_grad_global_control_variate}

In BN-SCAFFOLD, where we have $ \tilde{S}^{r,0}_j = S^{r,0} - k^{r-1}_j + k^{r-1} $, the drift between the global gradient and the global control variate is given by:

\begin{dmath}
\mathbb{E} \left[ \| c^{r-1} - \nabla_w F ( \bar{w}_{r-1} ; S^{r,0}) \|^2 \right] \\
\leq 
2 \sum_{j=1}^N P_j \ \mathbb{E} \left[ \| c^{r-1}_j - \nabla_w F_j ( \bar{w}_{r-1} ; \tilde{S}^{r,0}_j ) \|^2 \right] + 16 J^2 M^2 \mathbb{E} \left[ \| \bar{w}_{r-1} - \bar{w}_{r-2} \|^2 \right] + 16 J^2 \sum_{j=1}^N P_j \ \mathbb{E} \left[ \| S^{r-1,0}_j - k^{r-1}_j  \|^2 \right],
\end{dmath}

for $ r > 1$, and by

\begin{dmath}
\mathbb{E} \left[ \| c^0 - \nabla_w F ( \bar{w}_0 ; S^{1,0}) \|^2 \right] \\
\leq 
2 \sum_{j=1}^N P_j \ \mathbb{E} \left[ \| c^0_j - \nabla_w F_j ( \bar{w}_0 ; \tilde{S}^{1,0}_j ) \|^2 \right] + 8 J^2 \sum_{j=1}^N P_j \ \mathbb{E} \left[ \| S^{1,0}_j - k^0_j  \|^2 \right],
\end{dmath}

for $ r = 1$.
    
\end{lemma}

\begin{proof}

\begin{dmath}
\mathbb{E} \left[ \| c^{r-1} - \nabla_w F ( \bar{w}_{r-1} ; S^{r,0}) \|^2 \right] \\
\leq \sum_{j=1}^N P_j \ \mathbb{E} \left[ \| c^{r-1}_j - \nabla_w F_j ( \bar{w}_{r-1} ; S^{r,0}) \|^2 \right] \\
\leq 2 \sum_{j=1}^N P_j \ \mathbb{E} \left[ \| c^{r-1}_j - \nabla_w F_j ( \bar{w}_{r-1} ; \tilde{S}^{r,0}_j ) \|^2 \right] + 2 \sum_{j=1}^N P_j \ \mathbb{E} \left[ \| \nabla_w F_j ( \bar{w}_{r-1} ; S^{r,0})  - \nabla_w F_j ( \bar{w}_{r-1} ; \tilde{S}^{r,0}_j ) \|^2 \right] \\
\leq 2 \sum_{j=1}^N P_j \ \mathbb{E} \left[ \| c^{r-1}_j - \nabla_w F_j ( \bar{w}_{r-1} ; \tilde{S}^{r,0}_j ) \|^2 \right] + 2 J^2 \sum_{j=1}^N P_j \ \mathbb{E} \left[ \| S^{r,0}- \tilde{S}^{r,0}_j \|^2 \right] \\
\leq 2 \sum_{j=1}^N P_j \ \mathbb{E} \left[ \| c^{r-1}_j - \nabla_w F_j ( \bar{w}_{r-1} ; \tilde{S}^{r,0}_j ) \|^2 \right] + 2 J^2 \sum_{j=1}^N P_j \ \mathbb{E} \left[ \| S^{r,0}- S^{r,0}_j + k^{r-1}_j - k^{r-1} \|^2 \right] \\
\leq 2 \sum_{j=1}^N P_j \ \mathbb{E} \left[ \| c^{r-1}_j - \nabla_w F_j ( \bar{w}_{r-1} ; \tilde{S}^{r,0}_j ) \|^2 \right] + 4 J^2 \mathbb{E} \left[ \| S^{r,0} - k^{r-1} \|^2 \right] + 4 J^2 \sum_{j=1}^N P_j \ \mathbb{E} \left[ \| S^{r,0}_j - k^{r-1}_j  \|^2 \right] \\
\leq 2 \sum_{j=1}^N P_j \ \mathbb{E} \left[ \| c^{r-1}_j - \nabla_w F_j ( \bar{w}_{r-1} ; \tilde{S}^{r,0}_j ) \|^2 \right] + 8 J^2 \sum_{j=1}^N P_j \ \mathbb{E} \left[ \| S^{r,0}_j - k^{r-1}_j  \|^2 \right].
\end{dmath}

For $ r > 1 $

\begin{dmath}
\mathbb{E} \left[ \| c^{r-1} - \nabla_w F ( \bar{w}_{r-1} ; S^{r,0}) \|^2 \right] \\
\leq 2 \sum_{j=1}^N P_j \ \mathbb{E} \left[ \| c^{r-1}_j - \nabla_w F_j ( \bar{w}_{r-1} ; \tilde{S}^{r,0}_j ) \|^2 \right] + 16 J^2 \sum_{j=1}^N P_j \ \mathbb{E} \left[ \| S^{r,0}_j - S^{r-1,0}_j \|^2 + \| S^{r-1,0}_j - k^{r-1}_j  \|^2 \right] \\
\leq 2 \sum_{j=1}^N P_j \ \mathbb{E} \left[ \| c^{r-1}_j - \nabla_w F_j ( \bar{w}_{r-1} ; \tilde{S}^{r,0}_j ) \|^2 \right] + 16 J^2 M^2 \sum_{j=1}^N P_j \ \mathbb{E} \left[ \| \bar{w}_{r-1} - \bar{w}_{r-2} \|^2 \right] + 16 J^2 \sum_{j=1}^N P_j \ \mathbb{E} \left[ \| S^{r-1,0}_j - k^{r-1}_j  \|^2 \right] \\
\leq 2 \sum_{j=1}^N P_j \ \mathbb{E} \left[ \| c^{r-1}_j - \nabla_w F_j ( \bar{w}_{r-1} ; \tilde{S}^{r,0}_j ) \|^2 \right] + 16 J^2 M^2 \mathbb{E} \left[ \| \bar{w}_{r-1} - \bar{w}_{r-2} \|^2 \right] + 16 J^2 \sum_{j=1}^N P_j \ \mathbb{E} \left[ \| S^{r-1,0}_j - k^{r-1}_j  \|^2 \right].
\end{dmath}
    
\end{proof}

\begin{lemma}[BN-SCAFFOLD option \RNum{2} - Drift in estimated statistics]
\label{lemma:bn_scaffold_option_2_drift_in_estimated_stats}

In BN-SCAFFOLD with option \RNum{2}, if $ \gamma \leq \frac{1}{24 L E} $ , the drift in the estimated statistics is given by:

\begin{dmath}
    \sum_{j=1}^N P_j \ \mathbb{E} \left[ \| S^{r,0}_j - k^{r}_j \|^2 \right] \\
    \leq
    2 \sigma^2_s + \frac{M^2 }{96 L^2 E} \sigma^2 
    + M^2 \gamma^2 E  \left[ \frac{1}{4} + \frac{2}{3 L^2} J^2 M^2 \right] \sum_{j=1}^N P_j \ \mathbb{E} \left[ \sum_{t=1}^E \Big\| \sum_{i=1}^N P_i \ g_i ( w^{r-1, t-1}_i ; \tilde{s}^{r-1,t-1}_i ) \Big\|^2 \right] \\
    + \frac{M^2}{4 L^2} \sum_{j=1}^N P_j \ \mathbb{E} \left[ \| \nabla_w F_j ( \bar{w}_{r -2} ; \tilde{S}^{r-1, 0}_j ) - c^{r-1}_j \|^2 \right] \\
    + \frac{2 M^2}{3 L^2} J^2 \sum_{j=1}^N P_j \ \mathbb{E} \left[ \| S^{r-1,0}_j - k^{r-1}_j  \|^2 \right]
    + \frac{M^2}{24 L^2} \sum_{j=1}^N P_j \ \mathbb{E} \left[ \| \nabla_w F ( \bar{w}_{r-1} ; S^{r,0}) \|^2 \right],
\end{dmath}

for $ r > 1 $, and by

\begin{dmath}
\sum_{j=1}^N P_j \ \mathbb{E} \left[ \| S^{1,0}_j - k^{1}_j \|^2 \right] \\
\leq
2 \sigma^2_s + \frac{1}{96} \frac{M^2}{L^2 E} \sigma^2 
+ \frac{3}{24} \frac{M^2}{L^2} \sum_{j=1}^N P_j \ \mathbb{E} \left[ \| c^0_j - \nabla_w F_j ( \bar{w}_0 ; \tilde{S}^{1,0}_j ) \|^2 \right] \\
+ \frac{1}{3} \frac{M^2 J^2}{L^2} \sum_{j=1}^N P_j \ \mathbb{E} \left[ \| S^{1,0}_j - k^0_j  \|^2 \right]
+ \frac{1}{24} \frac{M^2}{L^2} \sum_{j=1}^N P_j \ \mathbb{E} \left[ \| \nabla_w F ( \bar{w}_{0} ; S^{1,0}) \|^2 \right],
\end{dmath}

for $ r = 1$.

\end{lemma}

\begin{proof}

By using that $ k^r_j = \frac{1 - \rho}{1 - \rho^E} \sum_{t=1}^E \rho^{E-t} s^{r, t-1}_j $, and Assumptions \ref{assumption_app:l_lipschitz_continuity_statistics} and \ref{assumption_app:bounded_batch_statistics_variance} we have

\begin{dmath}
\label{eq:bn_scaffold_option2_int_eq_2}
\sum_{j=1}^N P_j \ \mathbb{E} \left[ \| S^{r,0}_j - k^{r}_j \|^2 \right] \\
= \sum_{j=1}^N P_j \ \mathbb{E} \left[ \Big\| S^{r,0}_j - \frac{1-\rho}{1 - \rho^E} \sum_{t=1}^E \rho^{E-t} s^{r, t-1}_j \Big\|^2 \right] \\
\leq 2 \sum_{j=1}^N P_j \ \mathbb{E} \left[ \Big\| \frac{1-\rho}{1 - \rho^E} \sum_{t=1}^E \rho^{E-t} ( s^{r, t-1}_j - S^{r, t-1}_j ) \Big\|^2 \right] \\ + 2 \sum_{j=1}^N P_j \ \mathbb{E} \left[ \Big\| \frac{1-\rho}{1 - \rho^E} \sum_{t=1}^E \rho^{E-t} ( S^{r, t-1}_j - S^{r, 0}_j) \Big\|^2 \right] \\
\leq 2 \frac{1-\rho}{1 - \rho^E} \sum_{t=1}^E \rho^{E-t} \sum_{j=1}^N P_j \ \mathbb{E} \left[ \|  s^{r, t-1}_j - S^{r, t-1}_j \|^2 \right] \\ + 2 \frac{1-\rho}{1 - \rho^E}  \sum_{t=1}^E \rho^{E-t} \sum_{j=1}^N P_j \ \mathbb{E} \left[ \| S^{r, t-1}_j - S^{r, 0}_j \|^2 \right] \\
\leq 2 \frac{1-\rho}{1 - \rho^E} \sum_{t=1}^E \rho^{E-t} \sum_{j=1}^N P_j \ \sigma^2_s + 2 \frac{1-\rho}{1 - \rho^E} M^2 \sum_{t=1}^E \rho^{E-t} \sum_{j=1}^N P_j \ \mathbb{E} \left[ \| w^{r, t-1}_j - w^{r, 0}_j \|^2 \right] \\
\leq 2 \sigma^2_s + 2 \frac{1-\rho}{1 - \rho^E}M^2 \sum_{t=1}^E \rho^{E-t} \sum_{j=1}^N P_j \ \underbrace{\mathbb{E} \left[ \| w^{r, t-1}_j - \bar{w}_{r-1} \|^2 \right]}_{\text{client drift}}.
\end{dmath}

If $ \gamma \leq \frac{1}{24 L E} $ , Lemma \ref{lemma:client_drift_bound} provides an upper bound for the client drift. We then have

\begin{dmath}
\label{eq:bn_scaffold_option2_int_eq_3}
\sum_{j=1}^N P_j \ \mathbb{E} \left[ \| S^{r,0}_j - k^{r}_j \|^2 \right] \\
\leq 2 \sigma^2_s + 2 \frac{1-\rho}{1 - \rho^E}  M^2 \sum_{t=1}^E \rho^{2(E-t)} \left[ \frac{1}{192 L^2 E} \sigma^2 
    + \frac{1}{48 L^2} \mathbb{E} \left[ \| c^{r-1} - \nabla_w F ( \bar{w}_{r-1} ; S^{r,0}) \|^2 \right] \\
    + \frac{1}{48 L^2} \sum_{j=1}^N P_j \ \mathbb{E} \left[ \| \nabla_w F_j ( \bar{w}_{r-1} ; \tilde{S}^{r,0}_j) - c^{r-1}_j \|^2 \right]
    + \frac{1}{48 L^2} \sum_{j=1}^N P_j \ \mathbb{E} \left[ \| \nabla_w F ( \bar{w}_{r-1} ; S^{r,0}) \|^2 \right] \right] \\
\leq
2 \sigma^2_s + 2 M^2 \left[  \frac{1}{192 L^2 E} \sigma^2 
    + \frac{1}{48 L^2} \mathbb{E} \left[ \| c^{r-1} - \nabla_w F ( \bar{w}_{r-1} ; S^{r,0}) \|^2 \right] \\
    + \frac{1}{48 L^2} \sum_{j=1}^N P_j \ \mathbb{E} \left[ \| \nabla_w F_j ( \bar{w}_{r-1} ; \tilde{S}^{r,0}_j) - c^{r-1}_j \|^2 \right]
    + \frac{1}{48 L^2} \sum_{j=1}^N P_j \ \mathbb{E} \left[ \| \nabla_w F ( \bar{w}_{r-1} ; S^{r,0}) \|^2 \right] \right].
\end{dmath}

By using Lemma \ref{lemma:bn_scaffold_global_grad_global_control_variate} with $ r > 1 $, we can bound the second term in the parenthesis 

\begin{dmath}
\label{eq:bn_scaffold_option2_int_eq_4}
\sum_{j=1}^N P_j \ \mathbb{E} \left[ \| S^{r,0}_j - k^r_j \|^2 \right] \\
\leq
2 \sigma^2_s + \frac{M^2 }{96 L^2 E} \sigma^2 
    + \frac{M^2}{12 L^2} \sum_{j=1}^N P_j \ \mathbb{E} \left[ \| c^{r-1}_j - \nabla_w F_j ( \bar{w}_{r-1} ; \tilde{S}^{r,0}_j ) \|^2 \right] \\
    + \frac{2 M^2}{3 L^2} J^2 M^2 \mathbb{E} \left[ \| \bar{w}_{r-1} - \bar{w}_{r-2} \|^2 \right] + \frac{2 M^2}{3 L^2} J^2 \sum_{j=1}^N P_j \ \mathbb{E} \left[ \| S^{r-1,0}_{D_j} - k^{r-1}_j  \|^2 \right] \\
    + \frac{M^2}{24 L^2} \sum_{j=1}^N P_j \ \mathbb{E} \left[ \| \nabla_w F_j ( \bar{w}_{r-1} ; \tilde{S}^{r,0}_j) - c^{r-1}_j \|^2 \right]
    + \frac{M^2}{24 L^2} \sum_{j=1}^N P_j \ \mathbb{E} \left[ \| \nabla_w F ( \bar{w}_{r-1} ; S^{r,0}) \|^2 \right] \\
    \leq
    2 \sigma^2_s + \frac{M^2 }{96 L^2 E} \sigma^2 
    + \frac{M^2}{8 L^2} \sum_{j=1}^N P_j \ \mathbb{E} \left[ \| c^{r-1}_j - \nabla_w F_j ( \bar{w}_{r-1} ; \tilde{S}^{r,0}_j ) \|^2 \right] \\
    + \frac{2 M^2}{3 L^2} J^2 M^2 \mathbb{E} \left[ \| \bar{w}_{r-1} - \bar{w}_{r-2} \|^2 \right] + \frac{2 M^2}{3 L^2} J^2 \sum_{j=1}^N P_j \ \mathbb{E} \left[ \| S^{r-1,0}_{D_j} - k^{r-1}_j  \|^2 \right]
    + \frac{M^2}{24 L^2} \sum_{j=1}^N P_j \ \mathbb{E} \left[ \| \nabla_w F ( \bar{w}_{r-1} ; S^{r,0}) \|^2 \right].
\end{dmath}

And by using Lemmas \ref{lemma:grad_vs_control_var} and \ref{lemma2} with $r > 1$

\begin{dmath}
\label{eq:bn_scaffold_option2_int_eq_5}
    \sum_{j=1}^N P_j \ \mathbb{E} \left[ \| S^{r,0}_j - k^r_j \|^2 \right] \\
    \leq
    2 \sigma^2_s + \frac{M^2 }{96 L^2 E} \sigma^2 
    + M^2 \gamma^2 E  \left[ \frac{1}{4} + \frac{2}{3 L^2} J^2 M^2 \right] \sum_{j=1}^N P_j \ \mathbb{E} \left[ \sum_{t=1}^E \Big\| \sum_{i=1}^N P_i \ g_i ( w^{r-1, t-1}_i ; \tilde{s}^{r-1,t-1}_{D_i} ) \Big\|^2 \right] \\
    + \frac{M^2}{4 L^2} \sum_{j=1}^N P_j \ \mathbb{E} \left[ \| \nabla_w F_j ( \bar{w}_{r -2} ; \tilde{S}^{r-1, 0}_{D_j} ) - c^{r-1}_j \|^2 \right] \\
    + \frac{2 M^2}{3 L^2} J^2 \sum_{j=1}^N P_j \ \mathbb{E} \left[ \| S^{r-1,0}_{D_j} - k^{r-1}_j  \|^2 \right]
    + \frac{M^2}{24 L^2} \sum_{j=1}^N P_j \ \mathbb{E} \left[ \| \nabla_w F ( \bar{w}_{r-1} ; S^{r,0}) \|^2 \right].
\end{dmath}

By using Lemma \ref{lemma:bn_scaffold_global_grad_global_control_variate} with $ r = 1 $ we have,

\begin{dmath}
\sum_{j=1}^N P_j \ \mathbb{E} \left[ \| S^{1,0}_j - k^{1}_j \|^2 \right] \\
\leq
2 \sigma^2_s + \frac{1}{96} \frac{M^2}{L^2 E} \sigma^2 
+ \frac{3}{24} \frac{M^2}{L^2} \sum_{j=1}^N P_j \ \mathbb{E} \left[ \| c^0_j - \nabla_w F_j ( \bar{w}_0 ; \tilde{S}^{1,0}_j ) \|^2 \right] \\
+ \frac{1}{3} \frac{M^2 J^2}{L^2} \sum_{j=1}^N P_j \ \mathbb{E} \left[ \| S^{1,0}_j - k^0_j  \|^2 \right]
+ \frac{1}{24} \frac{M^2}{L^2} \sum_{j=1}^N P_j \ \mathbb{E} \left[ \| \nabla_w F ( \bar{w}_{0} ; S^{1,0}) \|^2 \right].
\end{dmath}

\end{proof}

\section{FedTAN}
\label{app:fed_tan_convergence_proof}

\begin{lemma}
    \begin{dmath}
\frac{1}{R} \sum_{r=1}^R \mathbb{E} \left[ \| \nabla_w F ( \bar{w}_{r-1} ; S^{r,0} ) \|^2 \right]  \\
\leq \frac{1}{1 - 34 \delta_{E, \gamma, L}} \left[ \frac{2}{\gamma R E} \left[ F( \bar{w}_0 ; S^{1,0}) - \underbar{F} \right]
+ 16 \ (2 + \frac{1}{R} ) \delta_{E, \gamma, L} V^2 \\
+ 16 \ \delta_{E, \gamma, L} \frac{1}{R}  \| \nabla_w F ( \bar{w}_{0} ; S^{1, 0} ) \|^2 
+ ( 1 + 2 \delta_{E, \gamma, L} ) \sigma^2 \right],
\end{dmath}

with $ \gamma < \frac{1}{12 L E}$.

\end{lemma}

\begin{proof}

In the FedTAN algorithm, the batch statistics and their gradients are completely matched for the first local iteration at each client, that is,

\begin{equation}
\tilde{S}^{r, 0}_j = 
    \begin{cases}
        \sum_{i=1}^N P_i S^{r, 0}_i  = S^{r, 0}  & \quad \text{if} \quad r=0 \\
        S^{r, 0}_j &\quad \text{otherwise}.
    \end{cases}
\end{equation}

Thus, $\mathcal{T}_3$ yields

\begin{equation}
\mathcal{T}_3
=  \frac{2}{R} (1 + \delta_{E, \gamma, L}) \sum_{r=1}^R \mathbb{E} \left[ \sum_{j=1}^N P_j \| \nabla_w F_j ( \bar{w}_{r-1} ; S^{r,0} ) - \nabla_w F_j ( \bar{w}_{r-1} ; S^{r,0} ) \|^2 \right] = 0,
\end{equation}

and $ \mathcal{T}_2 $ yields

\begin{dmath}
\mathcal{T}_2 = \frac{8}{R} \delta_{E, \gamma, L} \mathbb{E} \left[ \sum_{j=1}^N P_j \| \nabla_w F_j ( \bar{w}_{0} ; \tilde{S}^{1, 0}_j )- \underbrace{c^{0}_j}_{=0} \|^2 \right] = \frac{8}{R} \delta_{E, \gamma, L} \mathbb{E} \left[ \sum_{j=1}^N P_j \| \nabla_w F_j ( \bar{w}_{0} ; S^{1, 0} ) \|^2 \right] \\
\leq \frac{16}{R} \delta_{E, \gamma, L} \mathbb{E} \left[\sum_{j=1}^N P_j \| \nabla_w F_j ( \bar{w}_{0} ; S^{1, 0} ) - \nabla_w F ( \bar{w}_{0} ; S^{1, 0} ) \|^2  \right] + \frac{16}{R} \delta_{E, \gamma, L} \| \nabla_w F ( \bar{w}_{0} ; S^{1, 0} ) \|^2 \\
\leq \frac{16}{R} \delta_{E, \gamma, L} V^2 + \frac{16}{R} \delta_{E, \gamma, L} \| \nabla_w F ( \bar{w}_{0} ; S^{1, 0} ) \|^2,
\end{dmath}

and $ \mathcal{T}_4 $,

\begin{dmath}
\label{eq:fed_tan_conv_eq_int_1}
    \mathcal{T}_4 = \frac{16}{R} \delta_{E, \gamma, L} \sum_{r=1}^R \sum_{j=1}^N P_j \mathbb{E} \left[ \| \nabla_w F_j (\bar{w}_{r-1}; \tilde{S}^{r, 0}_j) - c_j^r \|^2 \right] \\
    = \frac{16}{R} \delta_{E, \gamma, L} \sum_{r=1}^R \sum_{j=1}^N P_j \mathbb{E} \left[ \| \nabla_w F_j (\bar{w}_{r-1}; S^{r, 0}) \|^2 \right] \\
    = \frac{16}{R} \delta_{E, \gamma, L} \sum_{r=1}^R \sum_{j=1}^N P_j \mathbb{E} \left[  \| \nabla_w F_j (\bar{w}_{r-1}; S^{r,0}) - \nabla_w F (\bar{w}_{r-1}; S^{r,0}) + \nabla_w F (\bar{w}_{r-1}; S^{r,0})  \|^2 \right] \\
    \leq
    \frac{32}{R} \delta_{E, \gamma, L} \left[ \sum_{r=1}^R \mathbb{E} \left[ \sum_{j=1}^N P_j \| \nabla_w F_j (\bar{w}_{r-1}; S^{r,0}) - \nabla_w F (\bar{w}_{r-1}; S^{r,0}) \|^2 \right]
    + \sum_{r=1}^R \sum_{j=1}^N P_j \mathbb{E} \left[ \| \nabla_w F (\bar{w}_{r-1}; S^{r,0})  \|^2 \right] \right] \\
    \leq 32 \delta_{E, \gamma, L} V^2 + \frac{32}{R} \delta_{E, \gamma, L} \sum_{r=1}^R \mathbb{E} \left[ \| \nabla_w F (\bar{w}_{r-1}; S^{r,0})  \|^2 \right],
\end{dmath}

where Assumption \ref{assumption_app:bounded_local_grad_dev} was used. In addition, by setting $ \gamma $ sufficiently small such that $ 1 - \gamma L E - 16 L^2 E^2 \gamma^2 \delta_{E, \gamma, L} \geq 0 $, the last term of Equation \eqref{eq:convergence_var_reduction_algorithms_async} is negative and can be removed from the bound. We then get,

\begin{dmath}
\frac{1}{R} \left( 1 - 2 \delta_{E, \gamma, L} \right) \sum_{r=1}^R \mathbb{E} \left[ \| \nabla_w F ( \bar{w}_{r-1} ; S^{r,0} ) \|^2 \right]  \\
\leq \frac{2}{\gamma R E} \left[ F( \bar{w}_0 ; S^{1,0}) - \underbar{F} \right] \\
+ 16 \delta_{E, \gamma, L} \frac{1}{R} V^2 + 16 \delta_{E, \gamma, L} \frac{1}{R}  \| \nabla_w F ( \bar{w}_{0} ; S^{1, 0} ) \|^2 \\
+ 32 \delta_{E, \gamma, L} V^2 + 32 \delta_{E, \gamma, L} \frac{1}{R} \sum_{r=1}^R \| \nabla_w F (\bar{w}_{r-1}; S^{r,0})  \|^2
+ ( 1 + 2 \delta_{E, \gamma, L} ) \sigma^2.
\end{dmath}

And thus,

\begin{dmath}
\frac{1}{R} \left( 1 - 34 \delta_{E, \gamma, L} \right) \sum_{r=1}^R \mathbb{E} \left[ \| \nabla_w F ( \bar{w}_{r-1} ; S^{r,0} ) \|^2 \right]  \\
\leq \frac{2}{\gamma R E} \left[ F( \bar{w}_0 ; S^{1,0}) - \underbar{F} \right]
+ (32 + \frac{16}{R} ) \delta_{E, \gamma, L} V^2 \\
+ 16 \delta_{E, \gamma, L} \frac{1}{R}  \| \nabla_w F ( \bar{w}_{0} ; S^{1, 0} ) \|^2 
+ ( 1 + 2 \delta_{E, \gamma, L} ) \sigma^2.
\end{dmath}

The following conditions on the learning rate need to be verified:

\begin{equation}
    \begin{cases}
        1 - 34 \delta_{E, \gamma, L} > 0 \\
        1 - \gamma L E - 16 L^2 E^2 \gamma^2 \delta_{E, \gamma, L} \geq 0,
    \end{cases}
\end{equation}

which impose the following conditions on the learning rate

\begin{equation}
    \begin{cases}
    \gamma < \frac{1}{12 L E} \\
    \gamma \leq \min \left\{\frac{1}{2}; \frac{1}{4} \right\} \frac{1}{L E},
    \end{cases}
\end{equation}

and thus $ \gamma < \frac{1}{12 L E} $.

\end{proof}

\begin{corollary}

By taking $ \gamma = \gamma_0  \frac{1}{L E \sqrt{R}} $, we obtain

\begin{dmath}
    \min_r \mathbb{E} \left[ \| \nabla_w F ( \bar{w}_{r-1} ; S^{r,0} ) \|^2 \right] \\
    \leq 
    \frac{1}{R} \sum_{r=1}^R \mathbb{E} \left[ \| \nabla_w F ( \bar{w}_{r-1} ; S^{r,0} ) \|^2 \right] \\
    \leq \mathcal{O} \left( \frac{L}{\sqrt{R}} \left[ F_0 - \underbar{F} \right] + (1 + \frac{1}{R} ) \frac{1}{R} V^2 + \frac{1}{R^2} \nabla F_0^2 + (1 + \frac{1}{R} ) \sigma^2 \right)
\end{dmath}
    
\end{corollary}

\begin{proof}

Refer to the proof of Corollary \ref{corollary:fed_avg_final_conv}.

\end{proof}

\section{Useful Lemmas}
\label{app:useful_lemmas}

\begin{lemma}
\label{lemma:grad_vs_control_var}

For $ r > 1$, we have

\begin{dmath}
\| \nabla_w F_j ( \bar{w}_{r -1} ; \tilde{S}^{r, 0}_j ) - c^{r-1}_j \|^2 \\
  \leq 2 L^2 \gamma^2 E \sum_{t=1}^E \| \sum_{i=1}^N P_i \ g_i ( w^{r-1, t-1}_i ; \tilde{s}^{r-1,t-1}_i ) \|^2  + 2 \| \nabla_w F_j ( \bar{w}_{r -2} ; \tilde{S}^{r-1, 0}_j ) - c^{r-1}_j \|^2.
\end{dmath}

\end{lemma}

\begin{proof}

For $r > 1$, we can write

\begin{dmath}
\| \nabla_w F_j ( \bar{w}_{r -1} ; \tilde{S}^{r, 0}_j ) - c^{r-1}_j \|^2 \\
 = \| \nabla_w F_j ( \bar{w}_{r -1} ; \tilde{S}^{r, 0}_j ) - \nabla_w F_j ( \bar{w}_{r -2} ; \tilde{S}^{r-1, 0}_j ) + \nabla_w F_j ( \bar{w}_{r -2} ; \tilde{S}^{r-1, 0}_j ) - c^{r-1}_j \|^2   \\
 \leq  2 \| \nabla_w F_j ( \bar{w}_{r -1} ; \tilde{S}^{r, 0}_j ) - \nabla_w F_j ( \bar{w}_{r -2} ; \tilde{S}^{r-1, 0}_j ) \|^2 + 2 \| \nabla_w F_j ( \bar{w}_{r -2} ; \tilde{S}^{r-1, 0}_j ) - c^{r-1}_j \|^2 \\
  \leq 2 L^2 \| \bar{w}_{r-1} - \bar{w}_{r-2} \|^2  + 2 \| \nabla_w F_j ( \bar{w}_{r -2} ; \tilde{S}^{r-1, 0}_j ) - c^{r-1}_j \|^2 \\
  \leq 2 L^2 \gamma^2 E \sum_{t=1}^E \| \sum_{i=1}^N P_j \ g_i ( w^{r-1, t-1}_i ; \tilde{s}^{r-1,t-1}_i ) \|^2  + 2 \| \nabla_w F_j ( \bar{w}_{r -2} ; \tilde{S}^{r-1, 0}_j ) - c^{r-1}_j \|^2,
\end{dmath}

where Assumption \ref{assumption_app:l_lipschitz_continuity} and Lemma \ref{lemma2} were used.
    
\end{proof}

The following Lemma has been proven in the SCAFFOLD original paper \cite{SCAFFOLD} (see Lemma 18 in Appendix E.1), but the proof is reproduced here to keep the notation of this article.

\begin{lemma}[Client drift bound in SCAFFOLD and BN-SCAFFOLD]
\label{lemma:client_drift_bound}

By taking $ \gamma \leq \frac{1}{24 L E} $, the client drift in SCAFFOLD and BN-SCAFFOLD can be bounded by the following expression

\begin{dmath}
    \sum_{j=1}^N P_j \ \mathbb{E} \left[ \| w^{r, t}_j - \bar{w}_{r-1} \|^2 \right] \\
    \leq
    \frac{1}{192 L^2 E} \sigma^2 
    + \frac{1}{48 L^2} \mathbb{E} \left[ \| c^{r-1} - \nabla_w F ( \bar{w}_{r-1} ; S^{r,0}) \|^2 \right] \\
    + \frac{1}{48 L^2} \sum_{j=1}^N P_j \ \mathbb{E} \left[ \| \nabla_w F_j ( \bar{w}_{r-1} ; \tilde{S}^{r,0}_j) - c^{r-1}_j \|^2 \right]
    + \frac{1}{48 L^2} \sum_{j=1}^N P_j \ \mathbb{E} \left[ \| \nabla_w F ( \bar{w}_{r-1} ; S^{r,0}) \|^2 \right]. 
\end{dmath}

\end{lemma}

\begin{proof}

\begin{dmath}
    \mathbb{E} \left[ \| w^{r, t}_j - \bar{w}_{r-1} \|^2 \right] \\
    = \mathbb{E} \left[ \| w^{r, t-1}_j - \gamma \left( g_j ( w^{r, t-1}_j ; \tilde{s}^{r,t-1}_{D_j}) + c^{r-1} - c^{r-1}_j \right) - \bar{w}_{r-1} \|^2 \right] \\
    \leq \mathbb{E} \left[ \| w^{r, t-1}_j - \gamma \left( \nabla_w F_j ( w^{r, t-1}_j ; \tilde{S}^{r,t-1}_{D_j}) + c^{r-1} - c^{r-1}_j \right) - \bar{w}_{r-1} \|^2 \right] + \gamma^2 \sigma^2 \\
    \leq (1 + \frac{1}{E-1} ) \mathbb{E} \left[ \| w^{r, t-1}_j - \bar{w}_{r-1} \|^2 \right] + E \gamma^2 \mathbb{E} \left[ \| \nabla_w F_j ( w^{r,t-1}_j ; \tilde{S}^{r,t-1}_j ) + c^{r-1} - c^{r-1}_j \|^2 \right] + \gamma^2 \sigma^2 \\
    \leq (1 + \frac{1}{E-1} ) \mathbb{E} \left[ \| w^{r, t-1}_j - \bar{w}_{r-1} \|^2 \right] + \gamma^2 \sigma^2 + E \gamma^2 \mathbb{E} \left[ \| \nabla_w F_j ( w^{r,t-1}_j ; \tilde{S}^{r,t-1}_{D_j}) - \nabla_w F_j ( \bar{w}_{r-1} ; \tilde{S}^{r,0}_j) \\
    + c^{r-1} - \nabla_w F ( \bar{w}_{r-1} ; S^{r,0}) + \nabla_w F ( \bar{w}_{r-1} ; \tilde{S}^{r,0}_j)- c^{r-1}_j + \nabla_w F ( \bar{w}_{r-1} ; S^{r,0}) \|^2 \right] \\
    \leq (1 + \frac{1}{E-1} ) \mathbb{E} \left[ \| w^{r, t-1}_j - \bar{w}_{r-1} \|^2 \right] + \gamma^2 \sigma^2 + 4 E \gamma^2 \mathbb{E} \left[ \| \nabla_w F_j ( w^{r,t-1}_j ; \tilde{S}^{r,t-1}_{D_j}) - \nabla_w F_j ( \bar{w}_{r-1} ; \tilde{S}^{r,0}_j) \|^2 \right] \\
    + 4 E \gamma^2 \mathbb{E} \left[ \| c^{r-1} - \nabla_w F ( \bar{w}_{r-1} ; S^{r,0}) \|^2 \right] + 4 E \gamma^2 \mathbb{E} \left[ \| \nabla_w F_j ( \bar{w}_{r-1} ; \tilde{S}^{r,0}_j) - c^{r-1}_j \|^2 \right] + 4 E \gamma^2 \mathbb{E} \left[ \| \nabla_w F ( \bar{w}_{r-1} ; S^{r,0}) \|^2 \right] \\
    \leq (1 + \frac{1}{E-1} + 4 E \gamma^2 L^2 ) \mathbb{E} \left[ \| w^{r, t-1}_j - \bar{w}_{r-1} \|^2 \right] + \gamma^2 \sigma^2 + 4 E \gamma^2 \mathbb{E} \left[ \| c^{r-1} - \nabla_w F ( \bar{w}_{r-1} ; S^{r,0}) \|^2 \right] \\ + 4 E \gamma^2 \mathbb{E} \left[ \| \nabla_w F_j ( \bar{w}_{r-1} ; \tilde{S}^{r,0}_j) - c^{r-1}_j \|^2 \right] + 4 E \gamma^2 \mathbb{E} \left[ \| \nabla_w F ( \bar{w}_{r-1} ; S^{r,0}) \|^2 \right],
\end{dmath}

where $L$-Lipschitz continuity and Lemma \ref{lemma:relaxed_triangle_inequality} were used. By taking the mean across different clients, and unrolling the recursion relationship, we have

\begin{dmath}
    \sum_{j=1}^N P_j \mathbb{E} \left[ \| w^{r, t}_j - \bar{w}_{r-1} \|^2 \right] \\ 
    \leq (1 + \frac{1}{E-1} + 4 E \gamma^2 L^2 ) \sum_{j=1}^N P_j \ \mathbb{E} \left[ \| w^{r, t-1}_j - \bar{w}_{r-1} \|^2 \right] + \gamma^2 \sigma^2 
    + 4 E \gamma^2 \mathbb{E} \left[ \| c^{r-1} - \nabla_w F ( \bar{w}_{r-1} ; S^{r,0}) \|^2 \right] \\
    + 4 E \gamma^2 \sum_{j=1}^N P_j \ \mathbb{E} \left[ \| \nabla_w F_j ( \bar{w}_{r-1} ; \tilde{S}^{r,0}_D) - c^{r-1}_{D_j} \|^2 \right]
    + 4 E \gamma^2 \sum_{j=1}^N P_j \ \mathbb{E} \left[ \| \nabla_w F ( \bar{w}_{r-1} ; S^{r,0}) \|^2 \right] \\
    \leq \left[ 
    \gamma^2 \sigma^2 
    + 4 E \gamma^2 \mathbb{E} \left[ \| c^{r-1} - \nabla_w F ( \bar{w}_{r-1} ; S^{r,0}) \|^2 \right]
    + 4 E \gamma^2 \sum_{j=1}^N P_j \ \mathbb{E} \left[ \| \nabla_w F_j ( \bar{w}_{r-1} ; \tilde{S}^{r,0}_j) - c^{r-1}_j \|^2 \right] \\
    + 4 E \gamma^2 \sum_{j=1}^N P_j \ \mathbb{E} \left[ \| \nabla_w F ( \bar{w}_{r-1} ; S^{r,0}) \|^2 \right] 
    \right] \sum_{e=0}^{t-1} (1 + \frac{1}{E-1} + 4 E \gamma^2 L^2 )^e
    \\
    \leq \left[ 
    \frac{1}{576 L^2 E^2} \sigma^2 
    + \frac{1}{144 L^2 E} \mathbb{E} \left[ \| c^{r-1} - \nabla_w F ( \bar{w}_{r-1} ; S^{r,0}) \|^2 \right]
    + \frac{1}{144 L^2 E} \sum_{j=1}^N P_j \ \mathbb{E} \left[ \| \nabla_w F_j ( \bar{w}_{r-1} ; \tilde{S}^{r,0}_j) - c^{r-1}_j \|^2 \right] \\
    + \frac{1}{144 L^2 E} \sum_{j=1}^N P_j \ \mathbb{E} \left[ \| \nabla_w F ( \bar{w}_{r-1} ; S^{r,0}) \|^2 \right] 
    \right] \sum_{e=0}^{t-1} (1 + \frac{1}{E-1} + \frac{1}{144 E} )^e
    \\
    \leq \left[ 
    \frac{1}{576 L^2 E^2} \sigma^2 
    + \frac{1}{144 L^2 E} \mathbb{E} \left[ \| c^{r-1} - \nabla_w F ( \bar{w}_{r-1} ; S^{r,0}) \|^2 \right]
    + \frac{1}{144 L^2 E} \sum_{j=1}^N P_j \ \mathbb{E} \left[ \| \nabla_w F_j ( \bar{w}_{r-1} ; \tilde{S}^{r,0}_j) - c^{r-1}_j \|^2 \right] \\
    + \frac{1}{144 L^2 E} \sum_{j=1}^N P_j \ \mathbb{E} \left[ \| \nabla_w F ( \bar{w}_{r-1} ; S^{r,0}) \|^2 \right] 
    \right] 3 E \\
    \leq
    \frac{1}{192 L^2 E} \sigma^2 
    + \frac{1}{48 L^2} \mathbb{E} \left[ \| c^{r-1} - \nabla_w F ( \bar{w}_{r-1} ; S^{r,0}) \|^2 \right] \\
    + \frac{1}{48 L^2} \sum_{j=1}^N P_j \ \mathbb{E} \left[ \| \nabla_w F_j ( \bar{w}_{r-1} ; \tilde{S}^{r,0}_j) - c^{r-1}_j \|^2 \right]
    + \frac{1}{48 L^2} \sum_{j=1}^N P_j \ \mathbb{E} \left[ \| \nabla_w F ( \bar{w}_{r-1} ; S^{r,0}) \|^2 \right],
\end{dmath}

where it was used that $ \gamma \leq \frac{1}{24 L E}$.

\end{proof}

The following Lemma is the equivalent of Lemma 2 in \cite{FedTAN}.

\begin{lemma}[Difference between subsequent global models]
\label{lemma2}
When using the variance reduction algorithms described in Definition \ref{def:variance_reduc_algos}, the difference between two subsequent global models is given by,

\begin{dmath}
\bar{w}_r - \bar{w}_{r-1} = - \gamma \sum_{t=1}^E \sum_{i=1}^{N} P_i \ g_i( w_i^{r, t-1} ; \tilde{s}_i^{r, t-1}),
\end{dmath}

and can be bounded by, 

\begin{dmath}
\| \bar{w}_r - \bar{w}_{r-1} \|^2 \leq \gamma^2 E \sum_{t=1}^E \Big\| \sum_{i=1}^{N} P_i  \ g_i( w_i^{r, t-1} ; \tilde{s}_i^{r, t-1}) \Big\|^2.
\end{dmath}

\begin{proof}

\begin{dmath}
\bar{w}_r = \sum_{i=1}^{N} P_i w_i^{r, E}
 = \sum_{i=1}^{N} P_i  \left[ \bar{w}_{r-1} - \gamma \sum_{t=1}^E  \left[ \ g_i( w_i^{r, t-1} ; \tilde{s}_i^{r, t-1}) + c^{r-1} - c^{r-1}_i  \right] \right] \\
 = \bar{w}_{r-1} - \gamma \sum_{i=1}^{N} \sum_{t=1}^E P_i g_i( w_i^{r, t-1} ; \tilde{s}_i^{r, t-1}) + \gamma \sum_{t=1}^E  \underbrace{\left( \sum_{i=1}^{N} P_i c^{r-1}_i - c^{r-1} \right)}_{ = 0} \\
= \bar{w}_{r-1} - \gamma \sum_{i=1}^{N} \sum_{t=1}^E P_i  \ g_i( w_i^{r, t-1} ; \tilde{s}_i^{r, t-1})
\end{dmath}

And thus, by using Lemma \ref{lemma:relaxed_triangle_inequality},

\begin{dmath}
\| \bar{w}_r - \bar{w}_{r-1} \|^2 \\
= \gamma^2 \Big\| \sum_{t=1}^E \sum_{i=1}^{N} P_i  \ g_i( w_i^{r, t-1} ; \tilde{s}_i^{r, t-1}) \Big\|^2 \\
\leq \gamma^2 E \sum_{t=1}^E \Big\| \sum_{i=1}^{N} P_i  \ g_i( w_i^{r, t-1} ; \tilde{s}_i^{r, t-1}) \Big\|^2.
\end{dmath}

\end{proof}

\end{lemma}

The following Lemma is the equivalent of Lemma 3 in \cite{FedTAN}.

\begin{lemma}[Accumulated difference between local models]
\label{lemma3}

When using the variance reduction algorithms described in Definition \ref{def:variance_reduc_algos}, and with Assumptions \ref{assumption_app:l_lipschitz_continuity} and \ref{assumption_app:bounded_local_grad_dev}, the accumulated difference between the local models and the global initialization at global iteration $r$ is given by,

\begin{dmath}
\sum_{t=1}^E \mathbb{E} \left[ \| w_i^{r, t-1} - \bar{w}_{r-1} \|^2 \right] \\ 
\leq \frac{E}{L^2} \delta_{E, \gamma, L} \ \sigma^2 + \frac{2 E}{L^2} \delta_{E, \gamma, L} \mathbb{E} \left[ \| \nabla_w F_i ( \bar{w}_{r-1}; \tilde{S}^{r, 0}_i ) - c_i^{r-1} \|^2 \right] \\ + \frac{2 E}{L^2} \delta_{E, \gamma, L} \mathbb{E} \left[ \Big\| \sum_{j=1}^N P_j \left[ \nabla_w F_j ( \bar{w}_{r-1} ; \tilde{S}^{r,0}_j ) - c_j^{r-1} \right] \Big\|^2 \right] \\
+ \frac{2 E}{L^2} \delta_{E, \gamma, L} \sum_{j=1}^N P_j \mathbb{E} \left[ \| \nabla_w F_j ( \bar{w}_{r-1} ; \tilde{S}^{r,0}_j ) - \nabla_w F_j ( \bar{w}_{r-1} ; S^{r,0} ) \|^2 \right]  + \frac{E}{L^2} \delta_{E, \gamma, L} \mathbb{E} \left[ \| \nabla_w F ( \bar{w}_{r-1} ; S^{r,0} ) \|^2 \right].
\end{dmath}

with $ \delta_{E, \gamma, L} = \frac{4 E^2 \gamma^2 L^2}{1 - 8 E^2 \gamma^2 L^2} $ and $ \gamma < \frac{1}{\sqrt{8} L E} $.

\begin{proof}

The local model of client $i$ at the $r$-th global iteration and $t-1$-th local iteration is given by,
\begin{dmath}
w_i^{r, t-1} = \bar{w}_{r-1} - \gamma \sum_{e=1}^{t-1} ( \ g_i( w_i^{r, e-1} ; \tilde{s}_i^{r, e-1}) + c^{r-1} - c_i^{r-1} )
\end{dmath}

Thus, 

\begin{dmath}
\| w_i^{r, t-1} - \bar{w}_{r-1} \|^2 \\
=  \gamma^2 \Biggl\| \sum_{e=1}^{t-1} \left[ \ g_i( w_i^{r, e-1} ; \tilde{s}_i^{r, e-1}) + \sum_{j=1}^N P_j c_j^{r-1} - c_i^{r-1} \right] \Biggl\|^2 \\
= \gamma^2 \Biggl\| \sum_{e=1}^{t-1} \left[ \ g_i( w_i^{r, e-1} ; \tilde{s}_i^{r, e-1}) - \nabla_w F_i ( w_i^{r, e-1} ; \tilde{S}_i^{r, e-1} ) \right] + \sum_{e=1}^{t-1} \left[ \nabla_w F_i ( w_i^{r, e-1} ; \tilde{S}_i^{r, e-1} ) - c_i^{r-1} \right] \\ + \sum_{e=1}^{t-1} \left[ \sum_{j=1}^N P_j c_j^{r-1} - \nabla_w F ( \bar{w}_{r-1} ; S^{r,0} ) \right] + \sum_{e=1}^{t-1} \nabla_w F ( \bar{w}_{r-1} ; S^{r,0} ) \Biggl\|^2,
\end{dmath}

where it was used that $c^{r-1} = \sum_{j=1}^N P_j c_j^{r-1} $. By using Lemma \ref{lemma:relaxed_triangle_inequality} and the fact that $ t-1 \leq E$

\begin{dmath}
\| w_i^{r, t-1} - \bar{w}_{r-1} \|^2 \leq \\
 \leq 
 4 E \gamma^2 \left[ \sum_{e=1}^{t-1} \| g_i( w_i^{r, e-1} ; \tilde{s}_i^{r, e-1}) - \nabla_w F_i ( w_i^{r, e-1} ; \tilde{S}_i^{r, e-1} ) \|^2 + \sum_{e=1}^{t-1} \| \nabla_w F_i ( w_i^{r, e-1} ; \tilde{S}_i^{r, e-1} ) - c_i^{r-1} \|^2 \\ + E \Big\| \sum_{j=1}^N P_j \left[ c_j^{r-1} - \nabla_w F_j ( \bar{w}_{r-1} ; S^{r,0} ) \right] \Big\|^2 + E \| \nabla_w F ( \bar{w}_{r-1} ; S^{r,0} ) \|^2 \right].
\end{dmath}

We then have:

\begin{dmath}
\| w_i^{r, t-1} - \bar{w}_r \|^2 \leq \\
\leq  4 E \gamma^2 \left[  \sum_{e=1}^{t-1} \| g_i( w_i^{r, e-1} ; \tilde{s}_i^{r, e-1}) - \nabla_w F_i ( w_i^{r, e-1} ; \tilde{S}_i^{r, e-1} ) \|^2 + \sum_{e=1}^{t-1} \| \nabla_w F_i ( w_i^{r, e-1} ; \tilde{S}_i^{r, e-1} ) - \nabla_w F_i ( \bar{w}_{r-1}; \tilde{S}^{r, 0}_i ) \\ + \nabla_w F_i ( \bar{w}_{r-1}; \tilde{S}^{r, 0}_i ) - c_i^{r-1} \|^2 + E \Big\| \sum_{j=1}^N P_j \left[ c_j^{r-1} - \nabla_w F_j ( \bar{w}_{r-1} ; S^{r,0} ) \right] \Big\|^2 + E \| \nabla_w F ( \bar{w}_{r-1} ; S^{r,0} ) \|^2 \right] \\
\leq  4 E \gamma^2 \left[ \sum_{e=1}^{t-1} \| g_i( w_i^{r, e-1} ; \tilde{s}_i^{r, e-1}) - \nabla_w F_i ( w_i^{r, e-1} ; \tilde{S}_i^{r, e-1} ) \|^2 + 2 \sum_{e=1}^{t-1} \| \nabla_w F_i ( w_i^{r, e-1} ; \tilde{S}_i^{r, e-1} ) - \nabla_w F_i ( \bar{w}_{r-1}; \tilde{S}^{r, 0}_i ) \|^2 \\ + 2 \sum_{e=1}^{t-1} \| \nabla_w F_i ( \bar{w}_{r-1}; \tilde{S}^{r, 0}_i ) - c_i^{r-1} \|^2 + E \Big\| \sum_{j=1}^N P_j \left[ c_j^{r-1} - \nabla_w F_j ( \bar{w}_{r-1} ; S^{r,0} ) \right] \Big\|^2 + E \| \nabla_w F ( \bar{w}_{r-1} ; S^{r,0} ) \|^2 \right] \\
\leq 4 E \gamma^2 \left[ \sum_{e=1}^{t-1} \| g_i( w_i^{r, e-1} ; \tilde{s}_i^{r, e-1}) - \nabla_w F_i ( w_i^{r, e-1} ; \tilde{S}_i^{r, e-1} ) \|^2 + 2 \sum_{e=1}^{t-1} \| \nabla_w F_i ( w_i^{r, e-1} ; \tilde{S}_i^{r, e-1} ) - \nabla_w F_i ( \bar{w}_{r-1}; \tilde{S}^{r, 0}_i ) \|^2 \\ + 2 E \| \nabla_w F_i ( \bar{w}_{r-1}; \tilde{S}^{r, 0}_i ) - c_i^{r-1} \|^2 + E \Big\| \sum_{j=1}^N P_j \left[ c_j^{r-1} - \nabla_w F_j ( \bar{w}_{r-1} ; S^{r,0} ) \right] \Big\|^2 + E \| \nabla_w F ( \bar{w}_{r-1} ; S^{r,0} ) \|^2 \right].
\end{dmath}

From Assumption \ref{assumption_app:l_lipschitz_continuity} we have that,

\begin{dmath}
\| \nabla_w F_i ( w_i^{r, e-1} ; \tilde{S}_i^{r, e-1} ) - \nabla_w F_i ( \bar{w}_{r-1}; \tilde{S}^{r, 0}_i ) \|^2 \leq L^2 \| w^{r, e-1}_i - \bar{w}_{r-1} \|^2.
\end{dmath}

Thus,

\begin{dmath}
\| w_i^{r, t-1} - \bar{w}_{r-1} \|^2 \leq \\
 \leq 4 E \gamma^2 \sum_{e=1}^{t-1} \| g_i( w_i^{r, e-1} ; \tilde{s}_i^{r, e-1}) - \nabla_w F_i ( w_i^{r, e-1} ; \tilde{S}_i^{r, e-1} ) \|^2 + 8 E \gamma^2 L^2 \sum_{e=1}^{t-1} \| w^{r, e-1}_i - \bar{w}_{r-1} \|^2 \\ + 8 E^2 \gamma^2 \| \nabla_w F_i ( \bar{w}_{r-1}; \tilde{S}^{r, 0}_i ) - c_i^{r-1} \|^2 + 4 \gamma^2 E^2 \Big\| \sum_{j=1}^N P_j \left[ c^{r-1}_j - \nabla_w F_j ( \bar{w}_{r-1} ; S^{r,0} ) \right] \Big\|^2 + 4 E^2 \gamma^2 \| \nabla_w F ( \bar{w}_{r-1} ; S^{r,0} ) \|^2.
\end{dmath}

By summing from $t=1$ to $E$,

\begin{dmath}
\sum_{t=1}^E \| w_i^{r, t-1} - \bar{w}_{r-1} \|^2 \\ \leq 
4 E \gamma^2 \sum_{t=1}^E \sum_{e=1}^{t-1} \| g_i( w_i^{r, e-1} ; \tilde{s}_i^{r, e-1}) - \nabla_w F_i ( w_i^{r, e-1} ; \tilde{S}_i^{r, e-1} ) \|^2 + 8 E \gamma^2 L^2 \sum_{t=1}^E \sum_{e=1}^{t-1} \| w^{r, e-1}_i - \bar{w}_{r-1} \|^2 \\
+ 8 E^3 \gamma^2 \| \nabla_w F_i ( \bar{w}_{r-1}; \tilde{S}^{r, 0}_i ) - c_i^{r-1} \|^2 + 4 E^3 \gamma^2 \Big\| \sum_{j=1}^N P_j \left[ c_j - \nabla_w F_j ( \bar{w}_{r-1} ; S^{r,0} ) \right] \Big\|^2 + 4 E^3 \gamma^2 \| \nabla_w F ( \bar{w}_{r-1} ; S^{r,0} ) \|^2 \\
\leq 4 E^2 \gamma^2 \sum_{t=1}^E \| g_i( w_i^{r, t-1} ; \tilde{s}_i^{r, t-1}) - \nabla_w F_i ( w_i^{r, e-1} ; \tilde{S}_i^{r, e-1} ) \|^2 + 8 E^2 \gamma^2 L^2 \sum_{t=1}^E  \| w^{r, t-1}_i - \bar{w}_{r-1} \|^2 \\
+ 8 E^3 \gamma^2 \| \nabla_w F_i ( \bar{w}_{r-1}; \tilde{S}^{r, 0}_i ) - c_i^{r-1} \|^2 + 4 E^3 \gamma^2 \Big\| \sum_{j=1}^N P_j \left[ c_j - \nabla_w F_j ( \bar{w}_{r-1} ; S^{r,0} ) \right] \|^2 + 4 E^3 \gamma^2 \| \nabla_w F ( \bar{w}_{r-1} ; S^{r,0} ) \Big\|^2,
\end{dmath}

where it was used that, for any two sequences $\alpha_n$, $\beta_n$,

\begin{dmath}
\sum_{t=1}^E \sum_{e=1}^{t-1} \| \alpha_{e-1} - \beta_{e-1} \|^2 \leq E \sum_{t=1}^E \| \alpha_{t-1} - \beta_{t-1} \|^2.
\end{dmath}

We thus obtain,

\begin{dmath}
(1 - 8 E^2 \gamma^2 L^2) \sum_{t=1}^E \| w_i^{r, t-1} - \bar{w}_{r-1} \|^2 \\ \leq  4 E^2 \gamma^2 \sum_{t=1}^E \| g_i( w_i^{r, t-1} ; \tilde{s}_i^{r, t-1}) - \nabla_w F_i ( w_i^{r, e-1} ; \tilde{S}_i^{r, e-1} ) \|^2 + 8 E^3 \gamma^2 \| \nabla_w F_i ( \bar{w}_{r-1}; \tilde{S}^{r, 0}_i ) - c_i^{r-1} \|^2 \\
+ 4 E^3 \gamma^2 \Big\| \sum_{j=1}^N P_j \left[ c_j^{r-1} - \nabla_w F_j ( \bar{w}_{r-1} ; S^{r,0} ) \right] \Big\|^2 + 4 E^3 \gamma^2 \| \nabla_w F ( \bar{w}_{r-1} ; S^{r,0} ) \|^2.
\end{dmath}

And, by assuming $ 1 -8 E^2 \gamma^2 L^2 > 0 $:

\begin{dmath}
\sum_{t=1}^E \| w_i^{r, t-1} - \bar{w}_{r-1} \|^2 \\ 
\leq \frac{4 E^2 \gamma^2}{1 - 8 E^2 \gamma^2 L^2} \sum_{t=1}^E \| g_i( w_i^{r, t-1} ; \tilde{s}_i^{r, t-1}) - \nabla_w F_i ( w_i^{r, e-1} ; \tilde{S}_i^{r, e-1} ) \|^2 + \frac{8 E^3 \gamma^2}{1 - 8 E^2 \gamma^2 L^2} \| \nabla_w F_i ( \bar{w}_{r-1}; \tilde{S}^{r, 0}_i ) - c_i^{r-1} \|^2 \\ 
+ \frac{4 E^3 \gamma^2}{1 - 8 E^2 \gamma^2 L^2} \Big\| \sum_{j=1}^N P_j \left[ c_j^{r-1} - \nabla_w F_j ( \bar{w}_{r-1} ; S^{r,0} ) \right] \Big\|^2 + \frac{4 E^3 \gamma^2}{1 - 8 E^2 \gamma^2 L^2} \| \nabla_w F ( \bar{w}_{r-1} ; S^{r,0} ) \|^2.
\end{dmath}

In addition, by using that 

\begin{dmath}
    \| \sum_{j=1}^N P_j \left[ c_j^{r-1} - \nabla_w F_j ( \bar{w}_{r-1} ; S^{r,0} ) \right] \|^2 \\
    = \Big\| \sum_{j=1}^N P_j \left[ c_j^{r-1} - \nabla_w F_j ( \bar{w}_{r-1} ; \tilde{S}^{r,0}_j ) + \nabla_w F_j ( \bar{w}_{r-1} ; \tilde{S}^{r,0}_j ) - \nabla_w F_j ( \bar{w}_{r-1} ; S^{r,0} ) \right] \Big\|^2 \\
    \leq 2 \| \sum_{j=1}^N P_j \left[ c_j^{r-1} - \nabla_w F_j ( \bar{w}_{r-1} ; \tilde{S}^{r,0}_j ) \right] \|^2 + 2 \Big\| \sum_{j=1}^N P_j \left[ \nabla_w F_j ( \bar{w}_{r-1} ; \tilde{S}^{r,0}_j ) - \nabla_w F_j ( \bar{w}_{r-1} ; S^{r,0} ) \right] \Big\|^2 \\
    \leq 2 \Big\| \sum_{j=1}^N P_j \left[ c_j^{r-1} - \nabla_w F_j ( \bar{w}_{r-1} ; \tilde{S}^{r,0}_j ) \right] \Big\|^2 + 2  \sum_{j=1}^N P_j \Big\| \nabla_w F_j ( \bar{w}_{r-1} ; \tilde{S}^{r,0}_j ) - \nabla_w F_j ( \bar{w}_{r-1} ; S^{r,0} ) \Big\|^2,
\end{dmath}

we have that

\begin{dmath}
\label{eq:lemma3-intermediary_eq_1}
\sum_{t=1}^E \| w_i^{r, t-1} - \bar{w}_{r-1} \|^2 \\ \leq \frac{4 E^2 \gamma^2}{1 - 8 E^2 \gamma^2 L^2} \sum_{t=1}^E \| g_i( w_i^{r, t-1} ; \tilde{s}_i^{r, t-1}) - \nabla_w F_i ( w_i^{r, e-1} ; \tilde{S}_i^{r, e-1} ) \|^2 + \frac{8 E^3 \gamma^2}{1 - 8 E^2 \gamma^2 L^2} \| \nabla_w F_i ( \bar{w}_{r-1}; \tilde{S}^{r, 0}_i ) - c_i^{r-1} \|^2  \\
+ \frac{8 E^3 \gamma^2}{1 - 8 E^2 \gamma^2 L^2} \Big\| \sum_{j=1}^N P_j \left[ \nabla_w F_j ( \bar{w}_{r-1} ; \tilde{S}^{r,0}_j ) - c_j^{r-1} \right] \Big\|^2 + \frac{8 E^3 \gamma^2}{1 - 8 E^2 \gamma^2 L^2} \sum_{j=1}^N P_j \| \nabla_w F_j ( \bar{w}_{r-1} ; \tilde{S}^{r,0}_j ) - \nabla_w F_j ( \bar{w}_{r-1} ; S^{r,0} ) \|^2 
\\ + \frac{4 E^3 \gamma^2}{1 - 8 E^2 \gamma^2 L^2} \| \nabla_w F ( \bar{w}_{r-1} ; S^{r,0} ) \|^2.
\end{dmath}

By defining 

\begin{equation}
    \delta_{E, \gamma, L} = \frac{4 E^2 \gamma^2 L^2}{1 - 8 E^2 \gamma^2 L^2},
\end{equation}

Equation \eqref{eq:lemma3-intermediary_eq_1} can be re-written as,

\begin{dmath}
\label{eq:lemma3-intermediary_eq_2}
\sum_{t=1}^E \| w_i^{r, t-1} - \bar{w}_{r-1} \|^2 \\ 
\leq \frac{1}{L^2} \delta_{E, \gamma, L} \sum_{t=1}^E \| g_i( w_i^{r, t-1} ; \tilde{s}_i^{r, t-1}) - \nabla_w F_i ( w_i^{r, e-1} ; \tilde{S}_i^{r, e-1} ) \|^2 + \frac{2 E}{L^2} \delta_{E, \gamma, L} \| \nabla_w F_i ( \bar{w}_{r-1}; \tilde{S}^{r, 0}_i ) - c_i^{r-1} \|^2  \\ + \frac{2 E}{L^2} \delta_{E, \gamma, L} \Big\| \sum_{j=1}^N P_j \left[ \nabla_w F_j ( \bar{w}_{r-1} ; \tilde{S}^{r,0}_j ) - c_j^{r-1} \right] \Big\|^2 + \frac{2 E}{L^2} \delta_{E, \gamma, L} \sum_{j=1}^N P_j \| \nabla_w F_j ( \bar{w}_{r-1} ; \tilde{S}^{r,0}_j ) - \nabla_w F_j ( \bar{w}_{r-1} ; S^{r,0} ) \|^2 \\
+ \frac{E}{L^2} \delta_{E, \gamma, L} \| \nabla_w F ( \bar{w}_{r-1} ; S^{r,0} ) \|^2.
\end{dmath}

By taking the expectation and using Assumption \ref{assumption_app:bounded_stochastic_grad_var}, we get

\begin{dmath}
\sum_{t=1}^E \mathbb{E} \left[ \| w_i^{r, t-1} - \bar{w}_{r-1} \|^2 \right] \\ 
\leq \frac{E}{L^2} \delta_{E, \gamma, L} \ \sigma^2 + \frac{2 E}{L^2} \delta_{E, \gamma, L} \mathbb{E} \left[ \| \nabla_w F_i ( \bar{w}_{r-1}; \tilde{S}^{r, 0}_i ) - c_i^{r-1} \|^2 \right] \\ + \frac{2 E}{L^2} \delta_{E, \gamma, L} \mathbb{E} \left[ \Big\| \sum_{j=1}^N P_j \left[ \nabla_w F_j ( \bar{w}_{r-1} ; \tilde{S}^{r,0}_j ) - c_j^{r-1} \right] \Big\|^2 \right] \\
+ \frac{2 E}{L^2} \delta_{E, \gamma, L} \sum_{j=1}^N P_j \mathbb{E} \left[ \| \nabla_w F_j ( \bar{w}_{r-1} ; \tilde{S}^{r,0}_j ) - \nabla_w F_j ( \bar{w}_{r-1} ; S^{r,0} ) \|^2 \right]  + \frac{E}{L^2} \delta_{E, \gamma, L} \mathbb{E} \left[ \| \nabla_w F ( \bar{w}_{r-1} ; S^{r,0} ) \|^2 \right].
\end{dmath}

\end{proof}

\end{lemma}

The following Lemma is the equivalent of Lemma 1 in \cite{FedTAN}.

\begin{lemma}
\label{lemma1}

When using the variance reduction algorithms as defined in Definition \ref{def:variance_reduc_algos}, and with Assumptions \ref{assumption_app:bounded_local_grad_dev} and \eqref{assumption_app:l_lipschitz_continuity}, the similarity between the global gradient and the difference between two subsequent global models can be bounded by:

\begin{dmath}
\label{eq:lemma1}
\mathbb{E} \left[ \langle \nabla_w F ( \bar{w}_{r-1} ; S^{r,0} ), \bar{w}_r - \bar{w}_{r-1} \rangle \right] \\
\leq
- \frac{ \gamma E}{2} \| \nabla_w F ( \bar{w}_{r-1} ; S^{r,0} )  \|^2 - \frac{ \gamma}{2} \sum_{t=1}^E \mathbb{E} \left[ \| \sum_{j=1}^N P_j \ g_j( w_j^{r, t-1} ; \tilde{s}_j^{r, t-1})  \|^2 \right] 
+ \gamma E \sum_{j=1}^N  P_j \| \nabla_w F_j ( \bar{w}_{r-1} ; S^{r,0} ) - \nabla_w F_j ( \bar{w}_{r-1} ; \tilde{S}^{r,0}_j ) \|^2 \\ + \gamma L^2 \sum_{j=1}^N  P_j \sum_{t=1}^E \| \bar{w}_{r-1} - w_j^{r, t-1} \|^2 + \frac{\gamma}{2} E \sigma^2.
\end{dmath}

\end{lemma}

\begin{proof}

By using Lemma \ref{lemma2}

\begin{dmath}
\mathbb{E} \left[ \langle \nabla_w F ( \bar{w}_{r-1} ; S^{r,0} ), \bar{w}_r - \bar{w}_{r-1} \rangle \right]
= \mathbb{E} \left[ \langle \nabla_w F ( \bar{w}_{r-1} ; S^{r,0} ), - \gamma 
\sum_{t=1}^E \sum_{j=1}^N P_j \ g_j( w_j^{r, t-1} ; \tilde{s}_j^{r, t-1}) \rangle \right]
= - \gamma  \sum_{t=1}^E \mathbb{E} \left[ \langle \nabla_w F ( \bar{w}_{r-1} ; S^{r,0} ), \sum_{j=1}^N P_j \ g_j( w_j^{r, t-1} ; \tilde{s}_j^{r, t-1}) \rangle \right].
\end{dmath}

By using Lemma \ref{lemma:identity_v1_cdot_v2},

\begin{dmath}
\label{eq:lemma_1_inter_eq_1}
\mathbb{E} \left[  \langle \nabla_w F ( \bar{w}_{r-1} ; S^{r,0} ), \bar{w}_r - \bar{w}_{r-1} \rangle \right] \\
= - \frac{ \gamma}{2} \sum_{t=1}^E \| \nabla_w F ( \bar{w}_{r-1} ; S^{r,0} ) \|^2 - \frac{ \gamma}{2} \sum_{t=1}^E \mathbb{E} \left[ \Big\| \sum_{j=1}^N P_j \ g_j( w_j^{r, t-1} ; \tilde{s}_j^{r, t-1})  \Big\|^2 \right] + \frac{ \gamma}{2} \sum_{t=1}^E \mathbb{E} \left[  \Big\| \nabla_w F ( \bar{w}_{r-1} ; S^{r,0} ) - \sum_{j=1}^N P_j \ g_j( w_j^{r, t-1} ; \tilde{s}_j^{r, t-1})  \Big\|^2 \right] \\
= - \frac{ \gamma E}{2} \| \nabla_w F ( \bar{w}_{r-1} ; S^{r,0} )  \|^2 - \frac{ \gamma}{2} \sum_{t=1}^E \mathbb{E} \left[ \Big\| \sum_{j=1}^N P_j \ g_j( w_j^{r, t-1} ; \tilde{s}_j^{r, t-1})  \Big\|^2 \right] \\ + \frac{ \gamma}{2} \sum_{t=1}^E \mathbb{E} \left[  \Big\| \nabla_w F ( \bar{w}_{r-1} ; S^{r,0} ) - \sum_{j=1}^N P_j \ g_j( w_j^{r, t-1} ; \tilde{s}_j^{r, t-1})  \Big\|^2 \right]
\end{dmath}

Using Jensen's inequality and Lemma \ref{lemma:separating_mean_and_var}, the third term of Equation \ref{eq:lemma_1_inter_eq_1} can be bounded by:

\begin{dmath}
\label{eq:lemma_1_inter_eq_2}
\sum_{t=1}^E \mathbb{E} \left[ \Big\| \nabla_w F ( \bar{w}_{r-1} ; S^{r,0} ) - \sum_{j=1}^N P_j \ g_j( w_j^{r, t-1} ; \tilde{s}_j^{r, t-1}) \Big\|^2 \right] \\
= \sum_{t=1}^E \mathbb{E} \left[ \Big\| \sum_{j=1}^N  P_j \left[ \nabla_w F_j ( \bar{w}_{r-1} ; S^{r,0} ) - \ g_j( w_j^{r, t-1} ; \tilde{s}_j^{r, t-1}) \right] \Big\|^2 \right] \\
\leq \sum_{t=1}^E \sum_{j=1}^N  P_j \mathbb{E} \left[ \| \nabla_w F_j ( \bar{w}_{r-1} ; S^{r,0} ) - \ g_j( w_j^{r, t-1} ; \tilde{s}_j^{r, t-1}) \|^2 \right] \\
\leq \sum_{t=1}^E \sum_{j=1}^N  P_j \left[ \| \nabla_w F_j ( \bar{w}_{r-1} ; S^{r,0} ) - \ \nabla_w F_j( w_j^{r, t-1} ; \tilde{S}_{D_j}^{r, t-1}) \|^2 + \sigma^2 \right] \\
\leq \sum_{t=1}^E \sum_{j=1}^N  P_j \| \nabla_w F_j ( \bar{w}_{r-1} ; S^{r,0} ) - \nabla_w F_j ( \bar{w}_{r-1} ; \tilde{S}^{r,0}_j ) + \nabla_w F_j ( \bar{w}_{r-1} ; \tilde{S}^{r,0}_j ) - \nabla_w F_j ( w^{r, t -1}_j ; \tilde{S}^{r,0}_j ) \|^2 + E \sigma^2 \\
\leq 2 \sum_{t=1}^E \sum_{j=1}^N  P_j \| \nabla_w F_j ( \bar{w}_{r-1} ; S^{r,0} ) - \nabla_w F_j ( \bar{w}_{r-1} ; \tilde{S}^{r,0}_j ) \|^2 + 2 \sum_{t=1}^E \sum_{j=1}^N  P_j \| \nabla_w F_j ( \bar{w}_{r-1} ; \tilde{S}^{r,0}_j ) - \nabla_w F_j ( w^{r, t -1}_j ; \tilde{S}^{r,0}_j ) \|^2 + E \sigma^2
\\ 
\\
\leq 2 E \sum_{j=1}^N  P_j \| \nabla_w F_j ( \bar{w}_{r-1} ; S^{r,0} ) - \nabla_w F_j ( \bar{w}_{r-1} ; \tilde{S}^{r,0}_j ) \|^2 + 2 L^2 \sum_{j=1}^N  P_j \sum_{t=1}^E \| \bar{w}_{r-1} - w_j^{r, t-1} \|^2 + E \sigma^2,
\end{dmath}

where in the last inequality, Assumptions \ref{assumption_app:bounded_local_grad_dev} and \ref{assumption_app:l_lipschitz_continuity} were used.

By plugging Equation \eqref{eq:lemma_1_inter_eq_2} into Equation \eqref{eq:lemma_1_inter_eq_1}:

\begin{dmath}
\mathbb{E} \left[ \langle \nabla_w F ( \bar{w}_{r-1} ; S^{r,0} ), \bar{w}_r - \bar{w}_{r-1} \rangle \right] \\
\leq
- \frac{ \gamma E}{2} \| \nabla_w F ( \bar{w}_{r-1} ; S^{r,0} )  \|^2 - \frac{ \gamma}{2} \sum_{t=1}^E \mathbb{E} \left[ \| \sum_{j=1}^N P_j \ g_j( w_j^{r, t-1} ; \tilde{s}_j^{r, t-1}) \|^2 \right]  
+ \gamma E \sum_{j=1}^N  P_j \| \nabla_w F_j ( \bar{w}_{r-1} ; S^{r,0} ) - \nabla_w F_j ( \bar{w}_{r-1} ; \tilde{S}^{r,0}_j ) \|^2 \\ + \gamma L^2 \sum_{j=1}^N  P_j \sum_{t=1}^E \| \bar{w}_{r-1} - w_j^{r, t-1} \|^2 + \frac{\gamma}{2} E \sigma^2.
\end{dmath}

\end{proof}